\documentclass[10pt,twocolumn,letterpaper]{article}

\usepackage{iccv}
\usepackage{times}
\usepackage{epsfig}
\usepackage{graphicx}
\usepackage{amsmath}
\usepackage{amssymb}
\usepackage{ifthen}
\usepackage{epsfig}
\usepackage{graphicx}
\usepackage{color}
\usepackage{microtype}
\usepackage{algorithm}
\usepackage{algorithmic}
\usepackage{ifthen}
\usepackage{booktabs}
\usepackage{mwe}
\usepackage{subcaption}

\newcommand{\final}{0}

\definecolor{SithColor}{rgb}{0.7,0,0} 
\definecolor{SimingColor}{rgb}{0,0.7,0}
\definecolor{HaitaoColor}{rgb}{0,0,0.7}
\newcommand{\chongyang}[1]{{\color{SithColor} Chongyang: #1}}
\newcommand{\siming}[1]{{\color{SimingColor} Siming: #1}}
\newcommand{\haitao}[1]{{\color{HaitaoColor} Haitao: #1}}

\newcommand{\warning}[1]{{\it\color{red} #1}}
\newcommand{\note}[1]{{\it\color{blue} #1}}
\newcommand{\nothing}[1]{}

\usepackage{array}
\newcolumntype{P}[1]{>{\centering\arraybackslash}p{#1}}
\usepackage{multirow}

\ifthenelse{\equal{\final}{1}}
{
\renewcommand{\chongyang}[1]{}
\renewcommand{\siming}[1]{}
\renewcommand{\haitao}[1]{}
\renewcommand{\warning}[1]{}
\renewcommand{\note}[1]{}
}
{}

\newcommand{\filename}[1]{\url{#1}}
\newcommand{\foldername}[1]{\url{#1}}

\let \bs = \boldsymbol
\let \set = \mathcal

\newcommand{\pred}{0}
\newcommand{\GMM}{M}

\newcommand{\gt}{\textup{gt}}
\newcommand{\abar}{\overline{\bs{a}}}

\newcommand{\R}{\mathcal{R}}

\usepackage[pagebackref=true,breaklinks=true,letterpaper=true,colorlinks,bookmarks=false]{hyperref}

\iccvfinalcopy 

\def\iccvPaperID{2877} 

\ificcvfinal\fi

\begin{document}

\title{Scene Synthesis via Uncertainty-Driven Attribute Synchronization}

\author{Haitao Yang\textsuperscript{1} \hspace{0.2in}
Zaiwei Zhang\textsuperscript{1} \hspace{0.2in}
Siming Yan\textsuperscript{1} \hspace{0.2in}
Haibin Huang\textsuperscript{2} \hspace{0.2in}
Chongyang Ma\textsuperscript{2}
\\
Yi Zheng\textsuperscript{2} \hspace{0.3in}
Chandrajit Bajaj\textsuperscript{1} \hspace{0.3in}
Qixing Huang\textsuperscript{1}
\vspace{4pt}
\\
\textsuperscript{1}{The University of Texas at Austin}\hspace{0.3in} \textsuperscript{2}{Kuaishou Technology}
}

\maketitle
\ificcvfinal\thispagestyle{empty}\fi

\begin{abstract}
Developing deep neural networks to generate 3D scenes is a fundamental problem in neural synthesis with immediate applications in architectural CAD, computer graphics, as well as in generating virtual robot training environments. This task is challenging because 3D scenes exhibit diverse patterns, ranging from continuous ones, such as object sizes and the relative poses between pairs of shapes, to discrete patterns, such as occurrence and co-occurrence of objects with symmetrical relationships. This paper introduces a novel neural scene synthesis approach that can capture diverse feature patterns of 3D scenes. Our method combines the strength of both neural network-based and conventional scene synthesis approaches. We use the parametric prior distributions learned from training data, which provide uncertainties of object attributes and relative attributes, to regularize the outputs of feed-forward neural models. Moreover, instead of merely predicting a scene layout, our approach predicts an over-complete set of attributes. This methodology allows us to utilize the underlying consistency constraints among the predicted attributes to prune infeasible predictions. Experimental results show that our approach outperforms existing methods considerably. The generated 3D scenes interpolate the training data faithfully while preserving both continuous and discrete feature patterns.
\end{abstract}

\section{Introduction}
\label{sec:introduction}

3D scene synthesis is a fundamental problem in deep generative modeling. This task is challenging because 3D scenes exhibit diverse patterns, ranging from continuous ones, such as the size of each object and the relative poses between pairs of shapes, to discrete patterns, such as occurrence and co-occurrence of objects and symmetric relations. Moreover, there are also generic geometric constraints, e.g., synthesized objects in a 3D scene should not inter-penetrate. Developing neural networks to capture all feature patterns while enforcing geometric constraints remains an open problem. Due to the diversity of feature patterns and constraints, the popular approach of developing a single data representation and training approach proves insufficient.  

\begin{figure}
\centering
\setlength\tabcolsep{0.1pt}
\begin{tabular}{cc}
\includegraphics[width=0.23\textwidth, trim=80px 130px 80px 20px, clip]{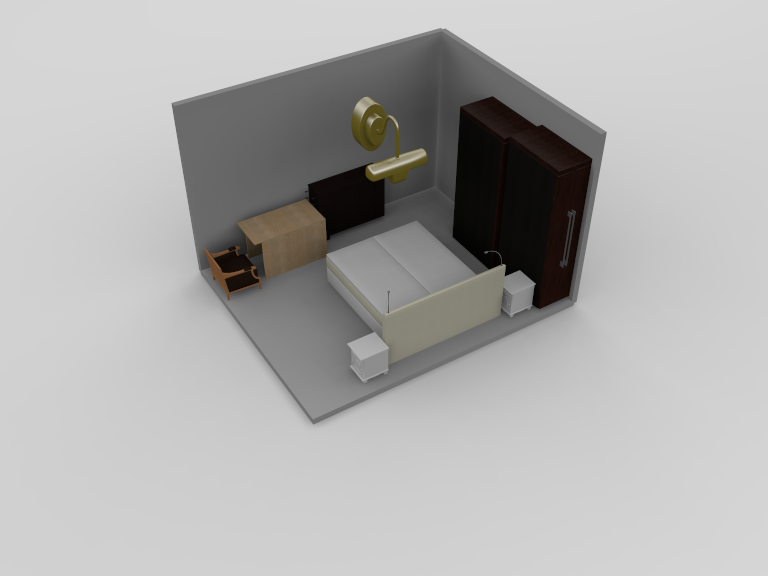} &
\includegraphics[width=0.23\textwidth, trim=80px 130px 80px 20px, clip]{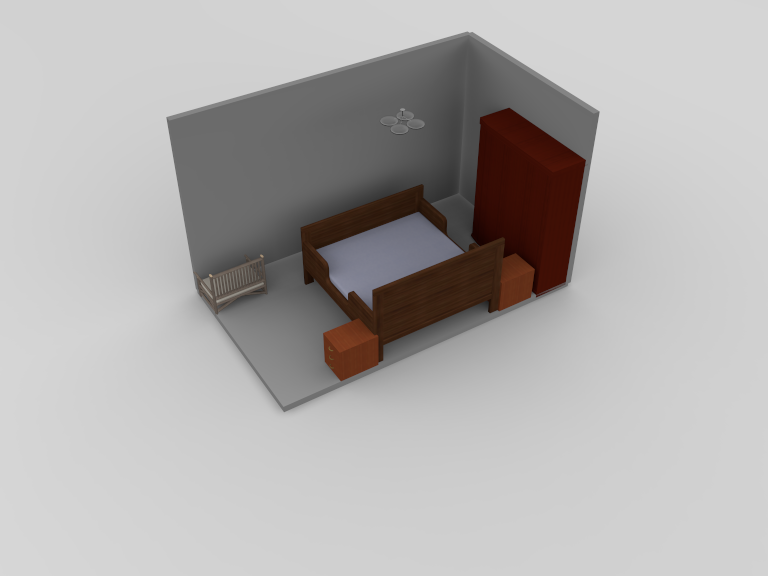}
\\
\includegraphics[width=0.23\textwidth, trim=80px 80px 80px 0px, clip]{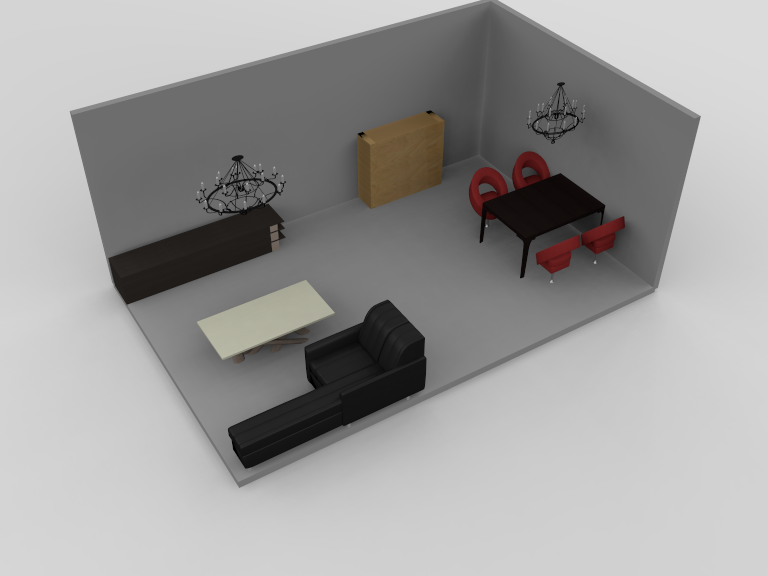} &
\includegraphics[width=0.23\textwidth, trim=80px 80px 80px 0px, clip]{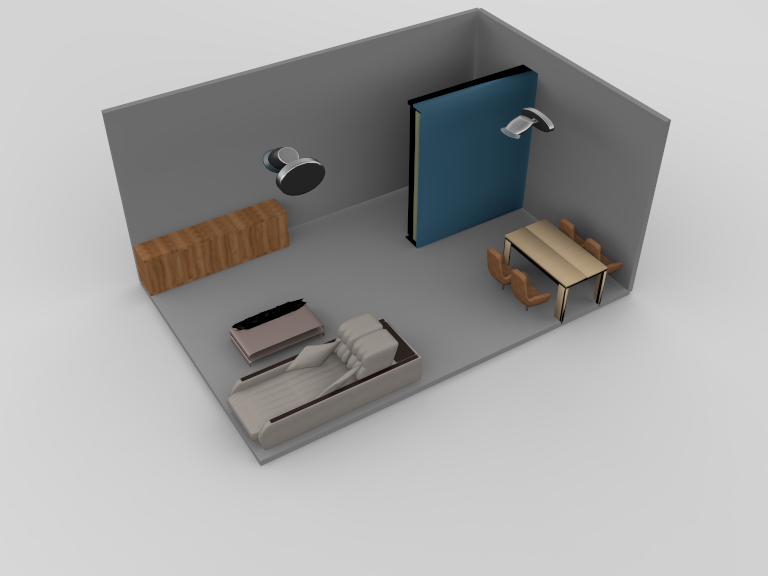}
\end{tabular}
\vspace{-10pt}
\caption{Randomly generated scenes (left) and their nearest neighbours (right) in the training set in 3D-FRONT.}
\label{Figure:Teaser}
\end{figure}

This paper introduces a novel approach to synthesizing 3D scenes represented as a collection of objects. Each object is encoded by its attributes such as size, pose, existence indicator, and geometric codes (c.f.~\cite{TulsianiSGEM17,ZhangYMLHVH20}). The theme of our approach is to look at 3D scene synthesis from hybrid viewpoints. Our goal is to combine the strengths of different approaches and representations that can capture diverse feature patterns and enforce different constraints. We execute this hybrid methodology at two levels.

First, instead of merely synthesizing the absolute attributes of each individual object, our approach predicts an \emph{over-complete} set of attributes which also include relative attributes (e.g., relative poses) between object pairs. Such relative attributes better capture spatial correlations among objects compared to only synthesizing absolute attributes. From a robust optimization point of view, over-complete attributes possess generic consistency constraints, e.g., the relative attributes should be consistent with object attributes. These constraints allow us to prune infeasible attributes in synthesis output (c.f.~\cite{Huang:2013:CSM,chen2014near,NIPS2017_6744,BajajGHHL18,Huang_2019_CVPR,Zhang_2019_CVPR,GuibasHL19,0001SRCPY16,SongSH20,Zhang:2020:H3DNet,YangYH20}). This approach is particularly suitable for neural outputs that exhibit weak correlations due to random initialization~\cite{Zhang:2020:H3DNet,DBLP:journals/corr/abs-1901-01608,DBLP:journals/corr/abs-1911-09189,collegial-ensembles}. We can therefore suppress output errors effectively by enforcing the consistency constraints among absolute and relative attributes.

Second, our approach combines the strengths of neural scene synthesis models and conventional scene generation methods. Neural models possess unbounded expressibility and can encode both continuous and discrete patterns. However, they typically produce single outputs that do not possess useful signals of uncertainties for synchronizing object attributes and relative attributes. For example, suppose we know the uncertainty of object attribute is high. In such cases, we can replace it with another one based on the attributes of other objects and the corresponding relative attributes. Similarly, we can discard a relative attribute if its uncertainty is high. Our approach addresses this issue by learning parametric prior distributions of absolute and relative attributes. Such distributions provide uncertainties of generated object attributes and relative attributes, 
offering rich signals to regularize them and prune outliers. Moreover, they also help enforce the penetration-free constraints. We introduce a Bayesian framework to integrate neural outputs and parametric prior distributions seamlessly. The hyperparameters of this Bayesian framework are optimized to maximize the performance of the final output.

We evaluate our approach on 3D-FRONT~\cite{Fu:2020:3dfront}. We also provide results on SUNCG~\cite{song2017semantic} to provide sufficient comparisons with baseline techniques. Experimental results show that our approach can generate 3D scenes different from the training examples while preserving discrete and continuous feature patterns. Our method outperforms baseline approaches both qualitatively and quantitatively. An ablation study justifies the design choices of our approach.
Our code is available at  \href{https://github.com/yanghtr/Sync2Gen}{https://github.com/yanghtr/Sync2Gen}.

\section{Related Work}
\label{sec:prior}

\begin{figure*}
\centering
\includegraphics[width=0.95\textwidth]{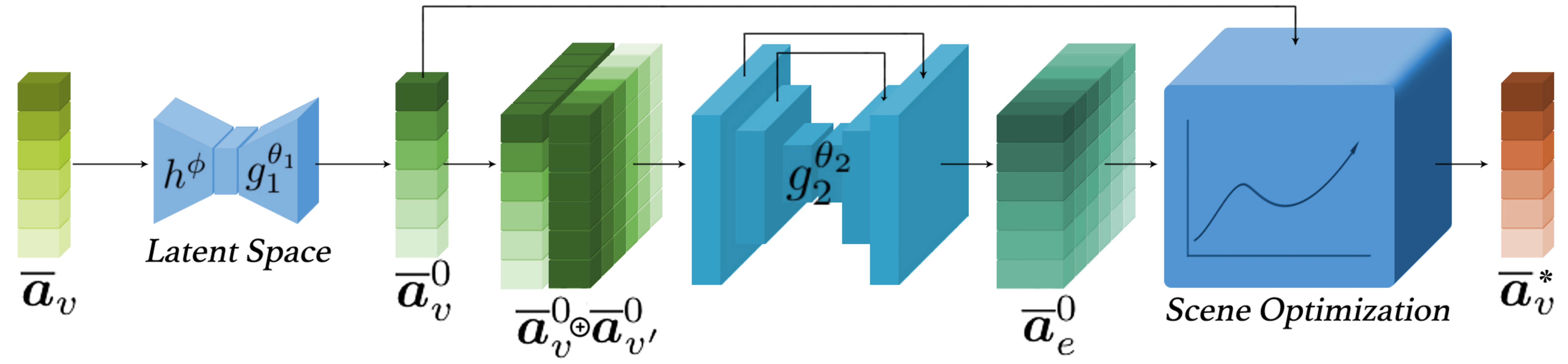}
\vspace{-0.14in}
\caption{Our network has three modules. The first module is a VAE model on object attributes. During testing, our approach takes a latent code as input and outputs synthesized object attributes. The induced relative attributes are fed into the second VAE module, which outputs synthesized relative attributes. The third module performs Bayesian scene optimization, combining synthesized object and relative attributes and parametric prior distributions to output the final object attributes.}
\vspace{-0.1in}
\label{Fig:Shape:Geometry}
\end{figure*}

3D scene synthesis has been studied considerably in the past. We refer to~\cite{zhang2019survey} for a recent survey and to \cite{chaudhuri2020learning} for a recent tutorial on this topic.

\vspace{-12pt}
\paragraph{Conventional scene synthesis approaches.}
Non-deep learning scene synthesis approaches fall into two categories. The first category applies data-driven and non-parametric approaches~\cite{Funkhouser:2004:ME,Kreavoy:2007:MIC,Chaudhuri:2010:DSC,Shen:2012:SRP,Xu:2012:FDS,Huang:2015:SRV}. The advantages of these methods are that they can handle datasets with significant structural variability. The downside is that these methods require complicated systems and careful parameter tuning. Another category employs probabilistic graphical models (e.g., Bayesian networks) for assembly-based modeling and synthesis ~\cite{Merrell:2010:CRB,Chaudhuri:2011:PRA,Kalogerakis:2012:PMC,Fisher:2012:ESO,Chaudhuri:2013:ACC,Chen:2014:ASM,Liu:2014:CCS,Kermani:2016:LSS}, these methods show improved performance but reply on hand-crafted modeling of structural patterns. Moreover, they typically only capture low-order correlations among the objects. Similar to these approaches, our approach learns distributions of object attributes and relative attributes. However, unlike using them to synthesize 3D scenes directly, we use the distributions to regularize and prune neural outputs. Most signals of the final output still come from the neural outputs.

\vspace{-12pt}
\paragraph{Deep scene synthesis approaches.}
Recent scene synthesis approaches use deep neural networks. Many of them focus on recurrent formulations~\cite{Li:2017:GGR,Ritchie:2016:NPM,DBLP:journals/corr/HaE17,DBLP:journals/corr/abs-1712-08290,DBLP:conf/iccv/ZouYYCH17,wang2018deep,ritchie2019fast,Li:2019:GRAINS,wang2019planit} that model relations between objects. These approaches either explicitly or implicitly utilize a hierarchical structure among the objects. Another category consists of feed-forward models ~\cite{tulsiani2017learning,Nash:2017:SVA,ZhangYMLHVH20} that synthesize a scene layout directly. Our approach advances the state-of-the-art on neural scene models in two ways. First, we propose synthesizing both object attributes and relative attributes and leveraging the underlying consistency constraints to prune wrong synthesis results. Second, we utilize parametric prior distributions to regularize synthesized object attributes and relative attributes further.

\vspace{-12pt}\paragraph{Over-complete predictions.}
Several recent works~\cite{0001SRCPY16,SongSH20,Zhang:2020:H3DNet,YangYH20} studied the methodology of first predicting over-complete intermediate geometric representations and then aggregating them into the final output. 
While our approach also predicts over-complete constraints, i.e., object attributes and relative attributes. We introduce how to integrate prior distributions that provide uncertainties of the predictions. Moreover, the hyperparameters of the synchronization procedure are jointly optimized.

\vspace{-12pt}\paragraph{Synchronization.}
Our approach is also relevant to recent works on transformation synchronization~\cite{singer2009angular,Huang:2013:CSM,Wang:2013:IMA,DBLP:conf/iccv/ChatterjeeG13,chen2014near,NIPS2017_6744,BajajGHHL18,Huang_2019_CVPR,Zhang_2019_CVPR,GuibasHL19,10.1145/3386569.3392402}, which takes relative transformations among a collection of geometric objects as input and outputs absolute transformations across the entire object collection. Our setting's major difference is to model prior distributions on the relative attributes (including relative poses), which provide uncertainties for inputs to arrangement optimization. Such uncertainties enable us to detect outliers more robustly. 

\section{Overview}
\label{Section:Problem:Statement:Approach:Overview}

\paragraph{Problem statement.}
Our goal is to train a variational auto-encoder that takes a 3D scene as input, maps it to a latent code, and finally decodes the latent code into the original input. In this paper, we assume that each scene is an arrangement of objects. Each object is given by its attributes, i.e., category label, size, pose, and a vector (i.e., a shape code) that encodes its geometry. We assume each instance of the training dataset is pre-segmented into these objects. 

\vspace{-10pt}\paragraph{Approach overview.}
Similar to~\cite{tulsiani2017multi,ZhangYMLHVH20}, we model each scene as selecting and deforming objects from a pre-defined over-complete set of objects. Each object is encoded by a vector that concatenates its attributes and a binary indicator that specifies whether an object is selected or not. A scene is then represented as a matrix, whose columns encode objects. 

Unlike standard approaches of designing a feed-forward network that directly outputs the object attributes (c.f.~\cite{tulsiani2017multi,ZhangYMLHVH20}), our method combines two new ideas for scene synthesis. First, our generative model outputs object attributes and relative attributes (e.g., the relative pose between a pair of objects). As shown in Figure~\ref{Fig:Shape:Geometry}, this is done by learning a VAE to synthesize object attributes. We will also use the latent space of this VAE to synthesize new scenes. The induced relative attributes from the synthesized object attributes are then fed into another VAE to produce synthesized relative attributes. 

The neural outputs provide an over-complete set of constraints on the final object attributes. We can recover accurate object attributes by exploring the underlying consistency constraints among this over-complete set even though some predictions are incorrect. Second, we learn prior distributions of object attributes and relative attributes from the training data. These prior distributions provide uncertainties of the predictions, which further regularize the final output. Note that the prior distributions consider both the continuous variables such as relative poses and discrete variables such as object counts and co-occurrences.  

We introduce a Bayesian scene optimization framework that seamlessly integrates neural predictions and prior distributions (See Figure~\ref{Fig:Shape:Geometry}). This framework exhibits two novel properties. First, it relaxes object indicators as real variables and employs continuous optimization to refine the object attributes. We show how to solve the induced optimization problem effectively via alternating minimization. Second, we introduce a simple and practical approach to optimize hyperparameters of the induced objective function. 

Note that our approach decouples training of the neural synthesis modules and learning of hyperparameters for the Bayesian scene optimization framework. Specifically, the neural synthesis modules are trained on a large-scale training set $\set{T}$. In contrast, the hyperparameters are trained on a separate small-scale validation dataset $\set{T}_{val}$. This approach allows us to alleviate the gap between the distribution of neural synthesis outputs on the training set and that on the testing set. We have found that this approach offers better results than naive end-to-end learning. 

\section{Bayesian Scene Optimization}
\label{Section:Bayesian:Opt}

This section presents the Bayesian scene optimization framework, which is the main contribution of this paper. It integrates the output of the variational auto-encoders which predict object attributes and relative attributes (see Section~\ref{Section:Implementation} for the details) and parametric prior distributions learned from the training data. 
In the following, we first introduce the problem statement of scene optimization in Section~\ref{Subsec:Statement}. We then describe the Bayesian formulation in Section~\ref{Subsec:Formulation}. Section~\ref{Subsec:Optimization} presents the strategy for solving the induced optimization problem. Finally, we describe hyperparameter optimization in Section~\ref{Subsec:Hyper:Parameter:Opt}.

\subsection{Problem Setup}
\label{Subsec:Statement}

We model a scene as a pre-defined graph $\set{G} = (\set{V},\set{E})$, where graph vertices encode objects and where edges connect pairs of objects. Each vertex $v\in \set{V}$ is associated with a pre-defined class label $c_v\in \set{C}$, where $\set{C}$ denote all the classes. Scene optimization amounts to recover each object's attributes encoded as a vector $\bs{a}_v$ and the indicator $z_v \in \{0,1\}$ that specifies whether $v$ is active or not. In other words, $\{z_v\}$ characterize which objects appear in a scene, and $\{\bs{a}_v\}$ specify the scene layout. Note that the precise parameterization of $\bs{a}_v$ will be described in Section~\ref{Subsec:Encoding:Parameterization}. 

Denote $\abar_v = (\bs{a}_v^T,z_v)^T$. The input to scene optimization consists of a prediction $\abar_v^{\pred}$ associated with each vertex $v\in \set{V}$ and a prediction $\abar_{e}^{\pred}$ associated with each edge $e = (v,v')\in \set{E}$, where the ground-truth $\abar_{e} = \phi(\abar_{v},\abar_{v'})$ encodes the relative attributes between $\bs{a}_v$ and $\bs{a}_{v'}$ (e.g., relations between beds and nightstands). The explicit expression of $\abar_{e}$ and $\phi$ are introduced in Section~\ref{Subsec:Encoding:Relative:Parameterization}. Note that both $\abar_v^{\pred}$ and $\abar_{e}^{\pred}$ are outputs of neural networks. 

Besides the vertex predictions $\{\abar_v^{\pred}\}$ and edge predictions $\{\abar_e^{\pred}\}$, our approach also utilizes prior distributions learned from the input data. The predictions and prior distributions are combined in a Bayesian framework.


\subsection{Formulation}
\label{Subsec:Formulation}

We formulate scene optimization as maximizing the following posterior distribution
\begin{align}
& \ P\big(\{\abar_v\}|\{\abar_v^{\pred}\}\cup\{\abar_{e}^{\pred}\}\big) \nonumber \\
\sim & \ P\big(\{\abar_v^{\pred}\}\cup\{\abar_{e}^{\pred}\}|\{\abar_v\}\big)\cdot P\big(\{\abar_v\}\big).
\label{Eq:Posterior:Distri}
\end{align}
where $P(\{\abar_v^{\pred}\}\cup\{\abar_{e}^{\pred}\}|\{\abar_v\})$ and $P(\{\abar_v\})$ are total likelihood and prior terms, respectively and $\sim$ denotes equal up to a scaling constant.

\vspace{-10pt}
\paragraph{Likelihood modeling.}
We model the total likelihood term by multiplying unary terms and pairwise terms associated with vertices and edges:
\begin{align*}
& \ P\big(\{\abar_v^{\pred}\}\cup\{\abar_{e}^{\pred}\}|\{\abar_{v}\}\big) \\
\sim &\  \prod_{v\in \set{V}}P\big(\abar_v^{\pred}|\abar_v\big) \cdot \prod_{e = (v,v')\in \set{E}}P\big(\abar_{e}^{\pred}|\abar_v,\abar_{v'}\big).
\end{align*}
Each unary term $P\big(\abar_v^{\pred}|\abar_v\big)$ measures the closeness between the prediction $\abar_v^{\pred}$ and the corresponding recovery $\abar_v$. We model the variance and employ a robust norm to handle outliers:
\begin{equation}
P\big(\abar_v^{\pred}|\abar_v\big) \sim \exp(-\frac{1}{2}\rho(\|\abar_v^{\pred}-\abar_v\|_{\Sigma_{c_v}^{-1}}, \alpha_{c_v}))
\label{Eq:Unary:Term}
\end{equation}
where $\rho(x,\alpha) = x^2/(x^2 + \alpha)$ is the Geman-McClure robust function~\cite{Barron19}; $\|\bs{x}\|_{A} = \bs{x}^TA\bs{x}$; $\alpha_{c_v}$ and the covariance matrix $\Sigma_{c_v}\succ 0$ are hyperparameters of class $c_v$. 

We use a similar formulation to model the pariwise term associated with each edge $e = (v,v') \in \set{E}$:
\begin{align}
& \ P\big(\abar_{e}^{\pred}|\abar_v,\abar_{v'}\big) \nonumber \\
\sim & \ \exp(-\frac{1}{2}\rho(\|\abar_{e}^{\pred}-\phi(\abar_{v},\abar_{v'})\|_{\Sigma_{c_e}^{-1}}, \alpha_{c_e}))
\label{Eq:Pairwise:Term}    
\end{align}
where $c_e = (c_v,c_{v'})$ denotes the class label of edge $e$; $\Sigma_{c_e}\succ 0$ and $\alpha_{c_e}$ are hyperparameters of class $c_e$. 

Combing (\ref{Eq:Unary:Term}) and (\ref{Eq:Pairwise:Term}), we arrive at the following formulation for the total likelihood term $P\big(\{\abar_v^{\pred}\}\cup\{\abar_{e}^{\pred}\}|\{\abar_{v}\}\big) $
\begin{align}
\sim &\exp\Big(-\frac{1}{2}\sum\limits_{v\in \set{V}}\rho(\|\abar_v^{\pred}-\abar_v\|_{\Sigma_{c_v}^{-1}},\alpha_{c_v}) \nonumber \\
& -\frac{1}{2}\sum\limits_{e = (v,v')\in \set{E}}\rho(\|\abar_{e}^{\pred}-\phi(\abar_v,\abar_{v'})\|_{\Sigma_{c_e}^{-1}},\alpha_{c_e})\Big)
\label{Eq:Likelihood:Term}
\end{align}

\vspace{-10pt}
\paragraph{Prior modeling.}
We model the total prior term by decoupling attributes and indicators and by multiplying unary terms and pairwise terms: $P(\{\abar_v\}) $
\begin{equation}
\sim  \prod_{v\in \set{V}}P_{c_v}(\bs{a}_v)\prod_{e = (v,v')\in \set{E}}P_{c_e}(\phi(\bs{a}_v,\bs{a}_{v'})) P(\{z_v\})
\label{Eq:Prior:Data}
\end{equation}
where $P_{c_v}(\bs{a}_v)$ models the attribute prior of the vertex class $c_v$;$P_{c_e}(\phi(\bs{a}_v,\bs{a}_{v'}))$ models the relative attribute prior of the edge class $c_e$; $P(\{z_v\})$ denotes the object count prior. 

We use generalized Gaussian mixture models (or GGMMs) to model $P_{c}$ and $P_{(c,c')}$:
\begin{align}
P_{c}(\bs{a}_v) & =  \GMM_{\mu_{c}}(\bs{a}_v),\label{Eq:GMM:G} \\
P_{(c,c')}(\phi(\bs{a}_v,\bs{a}_{v'})) &= \GMM_{\mu_{(c,c')}}(\phi(\bs{a}_v,\bs{a}_{v'})) \label{Eq:GMM:H}
\end{align}
where $\mu_c$ and $\mu_{(c,c')}$ denote hyperparameters of the mixture models. By GGMMs, we mean each mixture component is associated with an optimal mask to model the self-penetration free constraint between pairs of objects. Due to space constraints, we defer details of GGMMs and visualizations of the resulting GGMMs to the supplementary material. Note that our approach learns $\mu_c$ and $\mu_{(c,c')}$ from the training data and refines all the hyperparameters jointly to maximize the output of our approach. The joint optimization procedure is explained in Section~\ref{Subsec:Hyper:Parameter:Opt}. 

The prior term $P(\{z_v\})$ models object counts and object co-occurrences. Similar to the likelihood term, we model $P(\{z_v\})$ as a combination of unary and pairwise terms:
\begin{equation}
P(\{z_v\})\sim \prod_{c\in \set{C}} P_c(\bs{z}_{\set{V}_c}) \prod_{c,c'\in \set{C}} P_{(c,c')}(\bs{z}_{\set{V}_c},\bs{z}_{\set{V}_{c'}}). 
\label{Eq:Prior:Object:Count}
\end{equation}
where $\bs{z}_{\set{V}_c}$ collects indicators of vertices that belong to the vertex class $c$.
We again model both $P_c$ and $P_{(c,c')}$ using 1D and 2D GGMMs:
\begin{align}
P_c(\bs{z}_{\set{V}_c}) & = \GMM_{\gamma_c}(\bs{1}^T\bs{z}_{\set{V}_c}) \label{Eq:Gamma:Object:Indicator} \\
P_{(c,c')}(\bs{z}_{\set{V}_c},\bs{z}_{\set{V}_{c'}}) &= \GMM_{\gamma_{(c,c')}}\big((\bs{1}^T\bs{z}_{\set{V}_c},\bs{1}^T\bs{z}_{\set{V}_{c'}})\big) \label{Eq:Gamma:Object:Pair:Indicator}
\end{align}
Note that we again initialize $\gamma_c$ and $\gamma_{(c,c')}$ from data and refine them and other hyperparameters jointly. Please refer to the supplementary material for visualizations of the resulting GGMMs. 

\begin{figure}
\centering
\includegraphics[width=0.32\linewidth, trim=80px 100px 80px 0px, clip]{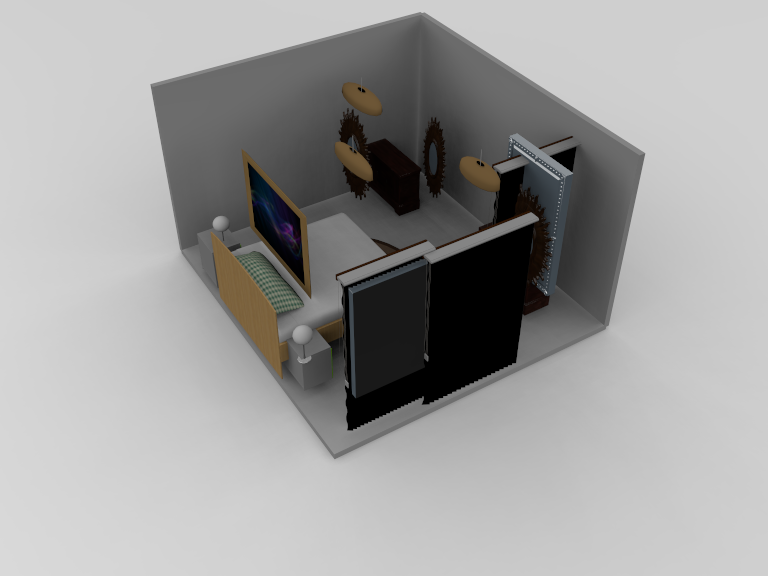}
\includegraphics[width=0.32\linewidth, trim=80px 100px 80px 0px, clip]{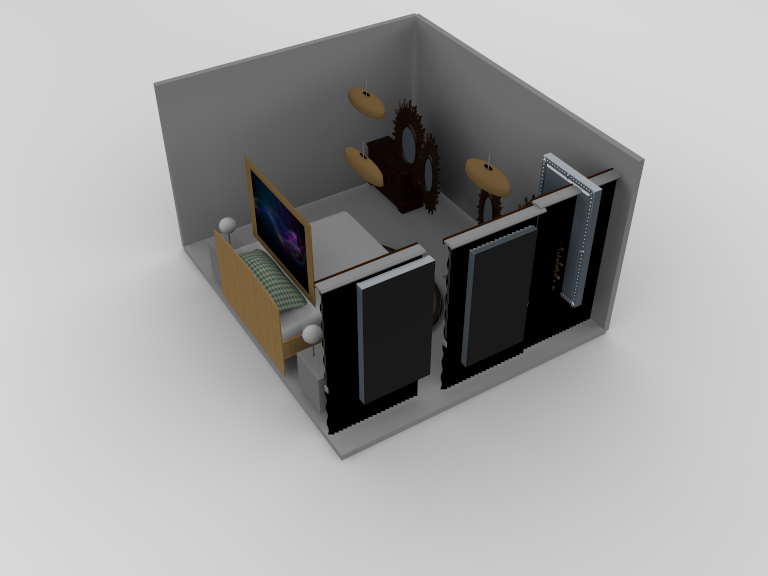} 
\includegraphics[width=0.32\linewidth, trim=80px 100px 80px 0px, clip]{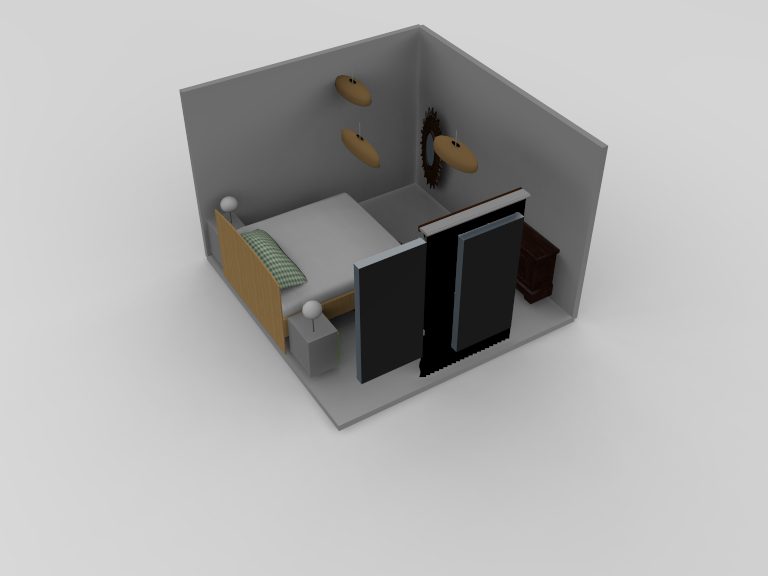}

\includegraphics[width=0.32\linewidth, trim=80px 100px 80px 0px, clip]{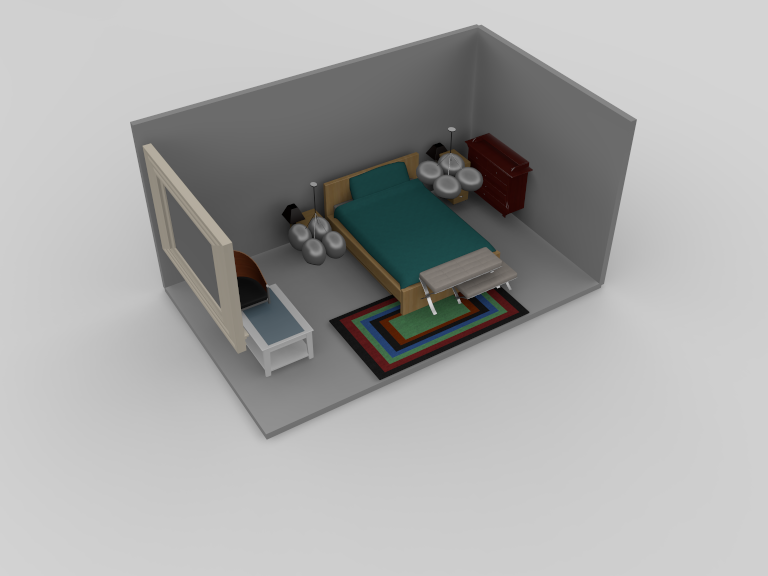}
\includegraphics[width=0.32\linewidth, trim=80px 100px 80px 0px, clip]{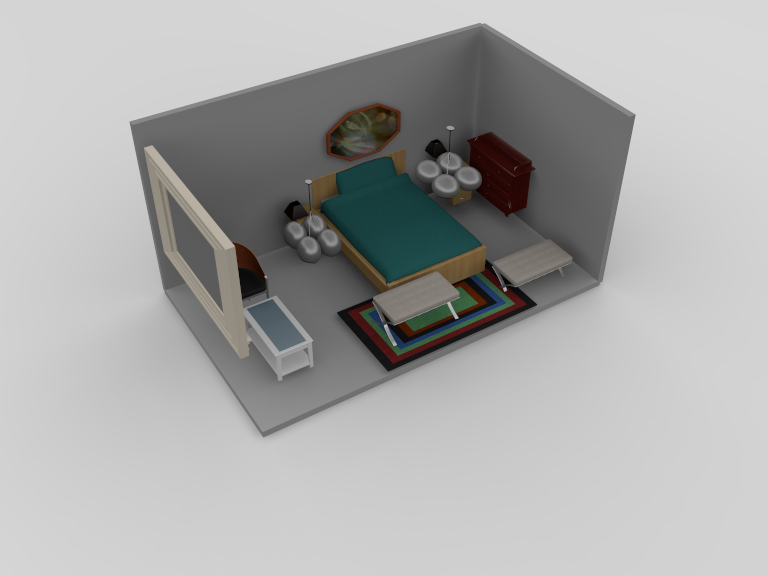}
\includegraphics[width=0.32\linewidth, trim=80px 100px 80px 0px, clip]{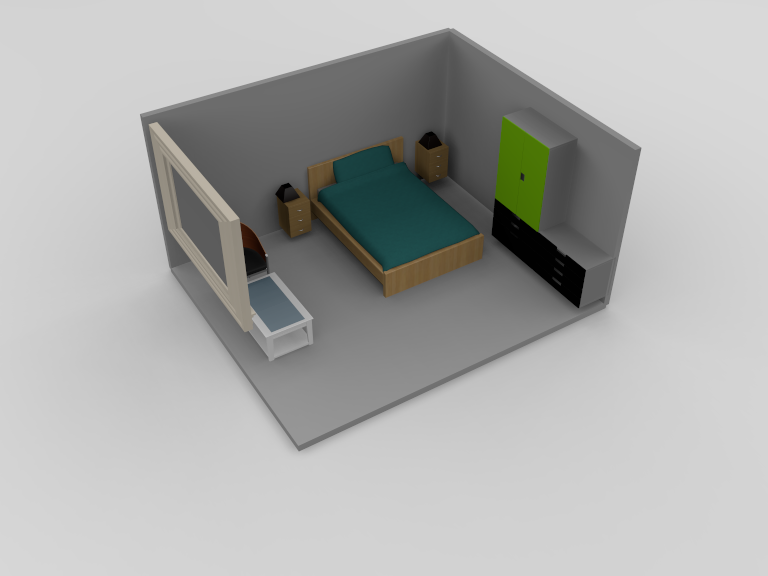}
\caption{Left: scene layout from predicted object attributes. Middle: scene layout by synchronizing predicted object attributes and relative attributes, i.e., only likelihood terms are used. Right: the output of scene optimization that combines both likelihood terms and prior terms.}
\label{Figure:Scene:Opt:Effects}
\vspace{-7pt}
\end{figure}

%
%
%

Substituting (\ref{Eq:Gamma:Object:Indicator}) and (\ref{Eq:Gamma:Object:Pair:Indicator}) into (\ref{Eq:Prior:Object:Count}) and combing (\ref{Eq:GMM:G}) and (\ref{Eq:GMM:H}), we arrive at the following prior model:
\begin{align}
& P(\{\abar_v\}) \sim  \prod_{v\in \set{V}}\GMM_{\mu_{c_v}}(\bs{a}_v)\prod_{e=(v,v')\in \set{E}}\GMM_{\mu_{c_e}}(\phi(\bs{a}_v,\bs{a}_{v'})) \nonumber \\
& \prod_{c\in \set{C}}\GMM_{\gamma_c}(\bs{1}^T\bs{z}_{\set{V}_c}) \prod_{c,c'\in \set{C}}\GMM_{\gamma_{(c,c')}}\big((\bs{1}^T\bs{z}_{\set{V}_c},\bs{1}^T\bs{z}_{\set{V}_{c'}})\big)
\label{Eq:Prior:Data:2}
\end{align}


\subsection{Scene Optimization}
\label{Subsec:Optimization}

Our goal is to find $\abar_v,v\in \set{V}$ that maximize the posterior distribution defined in (\ref{Eq:Posterior:Distri}). The variables consist of primitive parameters and primitive indicators. Our optimization strategy relaxes the indicator variables as real variables $z_v\in \R$. It then performs alternating optimization to refine these two categories of variables.
This relaxation strategy not only makes the optimization problem easy to solve but also facilitates hyperparameter learning (See Section~\ref{Subsec:Hyper:Parameter:Opt}).  

Specifically, when the indicator variables $z_v, v\in \set{V}$ are fixed, the optimization problem reduces to (we minimize the negation of the log-posterior)
\begin{align}
\underset{\{\bs{a}_v\}}{\min} & \sum\limits_{v\in \set{V}}\Big(\frac{1}{2}\rho\big(\|\abar_v^{\pred}-\abar_v\|_{\Sigma_{c_v}^{-1}},\alpha_{c_v}\big)-\log\big( M_{\mu_c}(\bs{a}_v)\big)\Big) \nonumber \\
& + \sum\limits_{e = (v,v')\in \set{E}}\Big(\frac{1}{2}\rho\big(\|\abar_{e}^{\pred}-\phi(\abar_v,\abar_{v'})\|_{\Sigma_{c_e}^{-1}},\alpha_{c_e}\big)\nonumber \\
& \qquad \qquad -\log\big(\GMM_{\mu_{(c,c')}}(\phi(\bs{a}_v,\bs{a}_{v'}))\big)\Big)
\label{Eq:Parameter:Opt}
\end{align}
The objective function in (\ref{Eq:Parameter:Opt}) is continuous in $\bs{a}_v$. We employ the limited-memory Broyden-Fletcher-Goldfarb-Shanno (or L-BFGS) algorithm for optimization. The initial solution is first set as $\bs{a}_{v}^{\pred}$ and then uses the output of the previous iteration. 

When $\bs{a}_v, v\in \set{V}$ are fixed, we solve
\begin{align}
\underset{\{z_v\}}{\min} & \frac{1}{2}\sum\limits_{v\in \set{V}}\rho\big(\|\abar_v^{\pred}-\abar_v\|_{\Sigma_{c_v}^{-1}}, \alpha_{c_v}\big)-\sum\limits_{c\in \set{C}}\log\big(\GMM_{\gamma_c}(\bs{1}^T\bs{z}_{\set{V}_c})\big) \nonumber \\
& + \frac{1}{2}\sum\limits_{e = (v,v')\in \set{E}}\rho\big(\|\abar_{e}^{\pred}-\bs{h}(\abar_v,\abar_{v'})\|_{\Sigma_{c_{e}}^{-1}}, \alpha_{c_{e}}\big)\nonumber \\
& - \sum\limits_{c,c'\in \set{C}}\log\big(\GMM_{\gamma_{(c,c')}}(\bs{1}^T\bs{z}_{\set{V}_c},\bs{1}^T\bs{z}_{\set{V}_{c'}})\big)
\label{Eq:Indicator:Opt}
\end{align}
We employ the same strategy as (\ref{Eq:Parameter:Opt}) for optimization. Our experiments suggest that 20-30 alternating iterations are sufficient. Figure~\ref{Figure:Scene:Opt:Effects} shows typical examples, where relative attributes and prior terms provide effective regularizations for object attributes. 


\subsection{Joint Hyperparameter Learning}
\label{Subsec:Hyper:Parameter:Opt}

In this section, we present an approach that learns hyperparameters of the scene optimization formulation described above. Specifically, to make the notations uncluttered, let 
$$
\Phi = \{\Sigma_c, \alpha_c, \mu_c,\gamma_c\}\cup\{\Sigma_{(c,c')}, \alpha_{(c,c')}, \mu_{(c,c')}, \gamma_{(c,c')}\}
$$
collect all the hyperparameters. Let $\bs{x}$ and $\bs{y}$ denote the inputs $\{\abar_v^{\pred}\}\cup \{\abar_e^{\pred}\}$ and the optimal solution to $\{\abar_v\}$ to (\ref{Eq:Posterior:Distri}). Finally, we denote the objective function in (\ref{Eq:Posterior:Distri}) as $f(\Phi,\bs{x},\bs{y})$.

Our goal is to train $\Phi$ using a validation set $\set{T}_{val}:=\{(\bs{x}_i,\bs{y}_i^{\gt})\}$ and a regularization loss $l(\Phi)$. Each $(\bs{x}_i,\bs{y}_i^{\gt})$ is computed by feeding $\bs{y}_i^{\gt}$ as the input to the encoder modules and setting $\bs{x}_i$ as outputs of the decoder modules. The regularization term $l(\Phi)$ combines all the loss terms that learn hyperparameters of the prior distributions, i.e., (\ref{Eq:GMM:G}), (\ref{Eq:GMM:H}), (\ref{Eq:Gamma:Object:Indicator}), and (\ref{Eq:Gamma:Object:Pair:Indicator}). We defer the explicit expression of $l(\Phi)$ to the supplementary material.

The performance of scene optimization depends on whether the ground-truth solution is a local minimum and whether the prediction modules' initial solution reaches this local minimum through optimization. We introduce a novel formulation that only involves function values of $f$ to enforce these two constraints:
\begin{align}
&\min\limits_{\Phi}  \quad l(\Phi) +  \sum\limits_{(\bs{x}_i,\bs{y}_i^{\gt})\in \set{T}_{val}}\underset{\bs{y}_i\sim \set{N}(\bs{y}_i^{\gt},r_m I)}{E} \nonumber   \\
& \qquad \Big( \max\big(f(\Phi, \bs{x}_i,\bs{y}_i^{\gt})-f(\Phi, \bs{x}_i,\bs{y}_i)+\delta,0\big)\label{Eq:Form2}\\
& \ + \lambda_s\underset{\bs{y}_i'\sim \set{N}(\bs{y}_i,r_s I)}{E} \big(f(\Phi, \bs{x}_i,\bs{y}_i)-f(\Phi, \bs{x}_i,\bs{y}_i')\big)^2\Big) 
\label{Eq:Form}  
\end{align}
where $\set{N}(\bs{y},rI)$ is the normal distribution with mean $\bs{y}$ and variance $rI$. Specifically, (\ref{Eq:Form2}) forces the ground-truth solution to be a local minimum. (\ref{Eq:Form}) prioritizes that the loss surface of $f$ is smooth, and therefore the local minimum has a large convergence radius. We determine the hyperparameters $r_m$, $r_s$, $\delta$, and $\lambda_s$ via cross-validation to minimize the L2 distances between the optimized object attributes and the ground-truth object attributes on the validation set $\set{T}_{val}$.

\section{Attribute Encoding and Synthesis}
\label{Section:Implementation}

This section describes the details of predicting initial object attributes and relative attributes. In Section~\ref{Subsec:Encoding:Parameterization}, we present the encoding of object attributes and the corresponding network architecture. Section~\ref{Subsec:Encoding:Relative:Parameterization} then presents the encoding of relative attributes and the corresponding network architecture. Finally, we present the training procedure for the neural models described above in Section~\ref{Subsec:Training}.

\subsection{Object Attributes} 
\label{Subsec:Encoding:Parameterization}

We use a similar approach as~\cite{TulsianiSGEM17,ZhangYMLHVH20} to encode a 3D scene as a collection of object attributes $\bs{a}_v$ and object indicators $z_v$. Each object attribute $\bs{a}_v$ is encoded as a vector in $\R^{12}$. The elements of $\bs{a}_v$ include size parameters $\bs{s}_v\in \R^3$, orientation parameters $\bs{r}_v\in \R^3$, location parameters $\bs{t}_v\in \R^3$, and shape codes $\bs{d}_v \in \R^3$. The size parameters $\bs{s}_v = (s_v^x,s_v^y,s_v^z)^T$ encode the scalings of $v$ aligned with the axis of the coordinate system associated with each object $v$. $\bs{r}_v = (\theta_v^x,\theta_v^y,\theta_v^z)$ collects the Euler angles that specify the orientation (i.e., a rotation) of $v$ in the world coordinate system. $\bs{t}_v$ specifies the location of $v$ in the world coordinate system. Finally, $\bs{d}_v$ is obtained in two steps. The first step uses the pre-trained model~\cite{0001ZXFT16} to obtain a latent code for each object's shape. The second step then performs PCA among latent codes of all the objects in training set to obtain the coordinates of the top-3 principal vectors. During testing, we use the synthesized code to search for the closest object in the training set.

Let $N_c$ be the maximum number of objects of each object class $c\in \set{C}$. We parameterize a 3D scene using a matrix $\overline{A}_{\set{C}} \in \R^{13\times (\sum_{c\in \set{C}} N_c)}$, where the columns of $\overline{A}_{\set{C}}$ are $\abar_v = (\bs{a}_v,z_v)^T$ of the corresponding objects.  

We adopt the variational auto-encoder (or VAE) architecture in~\cite{ZhangYMLHVH20}. The network design utilizes sparsely connected layers to alleviate the overfitting issue. As shown in Figure~\ref{Fig:Shape:Geometry}, we will sample the latent space of this VAE to synthesis 3D scenes. Specifically, the decoder of this VAE synthesizes object attributes. As we will discuss shortly, the output is then fed into another network to output relative attributes. The object attributes are optimized by feeding the the neural outputs into the scene optimization framework described in Section~\ref{Section:Bayesian:Opt}. 

\begin{figure*}
\centering
    \begin{subfigure}[b]{0.49\textwidth}
        \centering
        \includegraphics[width=0.32\textwidth, trim=80px 80px 80px 0px, clip]{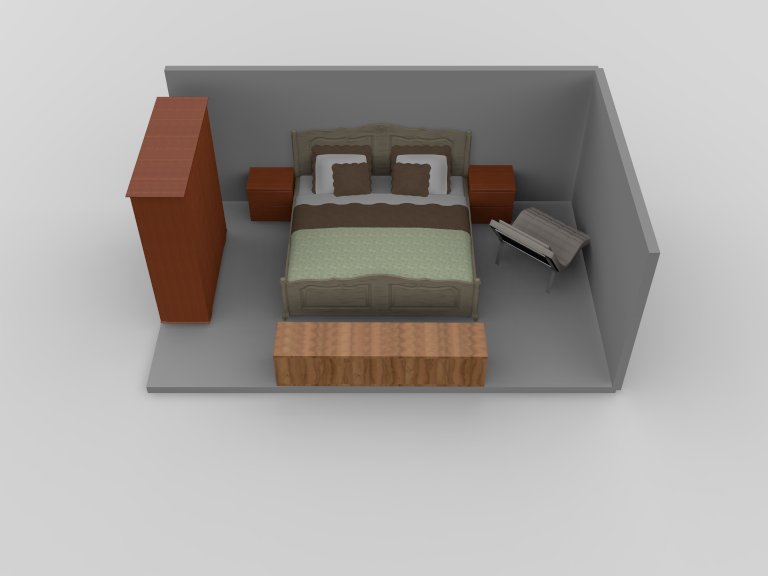}
        \includegraphics[width=0.32\textwidth, trim=80px 80px 80px 0px, clip]{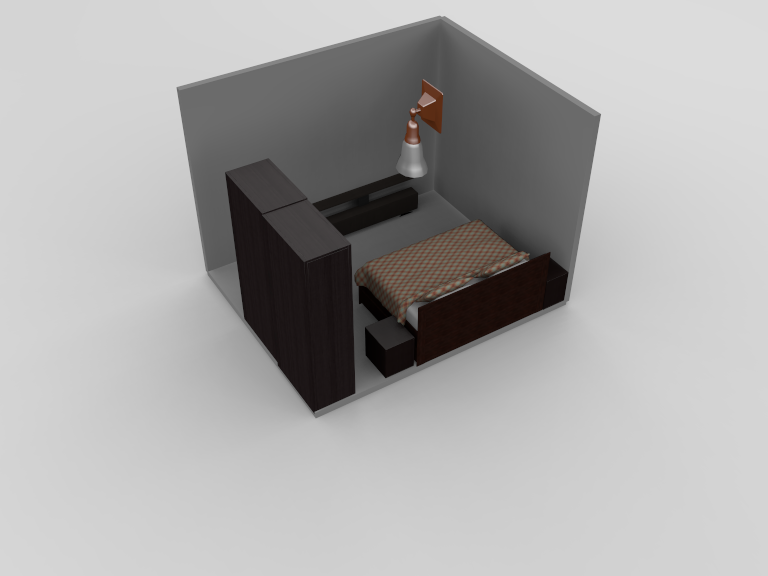}
        \includegraphics[width=0.32\textwidth, trim=80px 80px 80px 0px, clip]{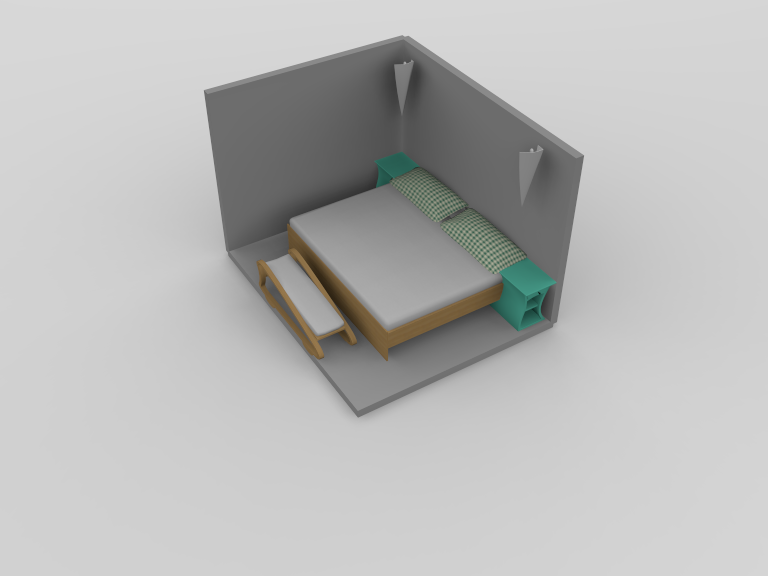}
        
        \includegraphics[width=0.32\textwidth, trim=80px 80px 80px 0px, clip]{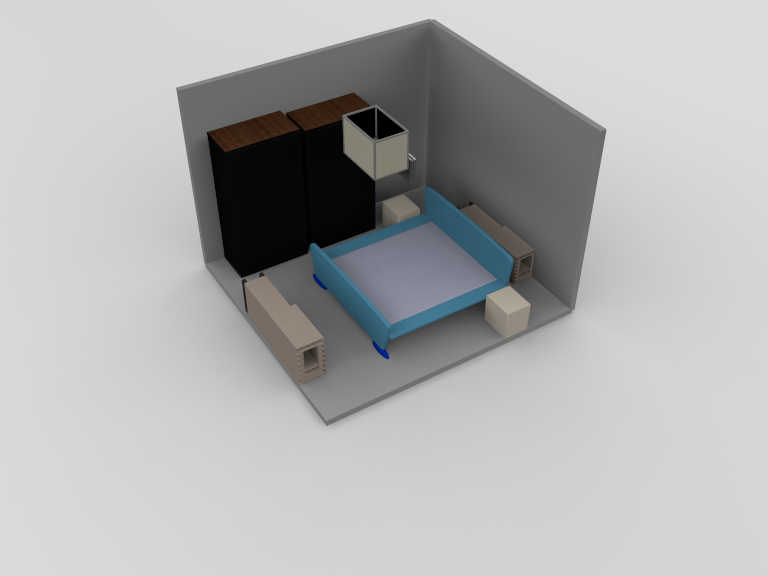}
        \includegraphics[width=0.32\textwidth, trim=80px 80px 80px 0px, clip]{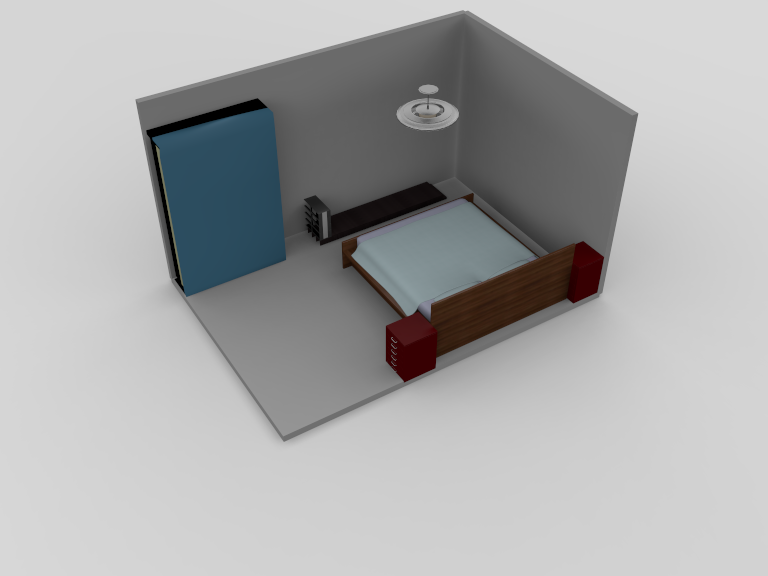}
        \includegraphics[width=0.32\textwidth, trim=80px 80px 80px 0px, clip]{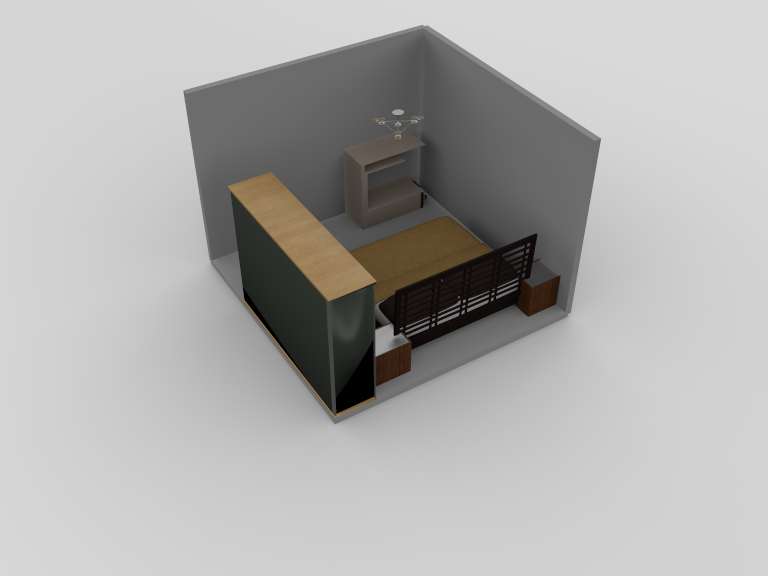}
        
        \includegraphics[width=0.32\textwidth, trim=80px 80px 80px 0px, clip]{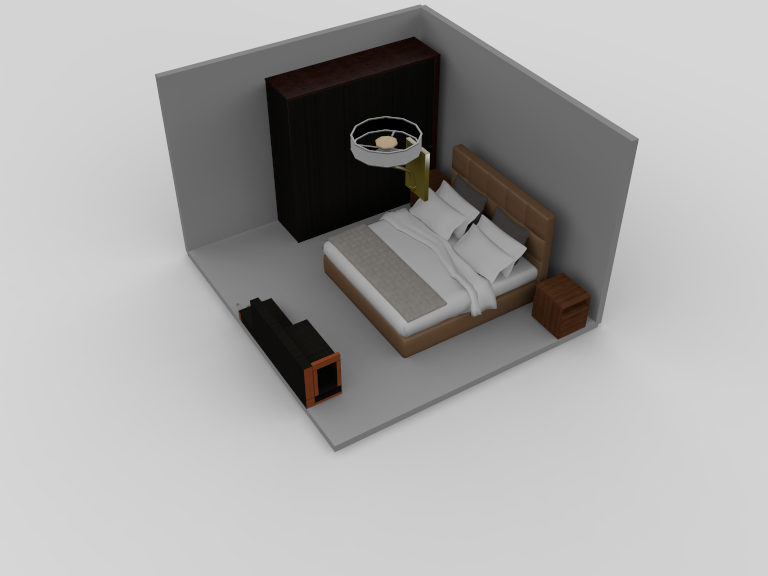}
        \includegraphics[width=0.32\textwidth, trim=80px 80px 80px 0px, clip]{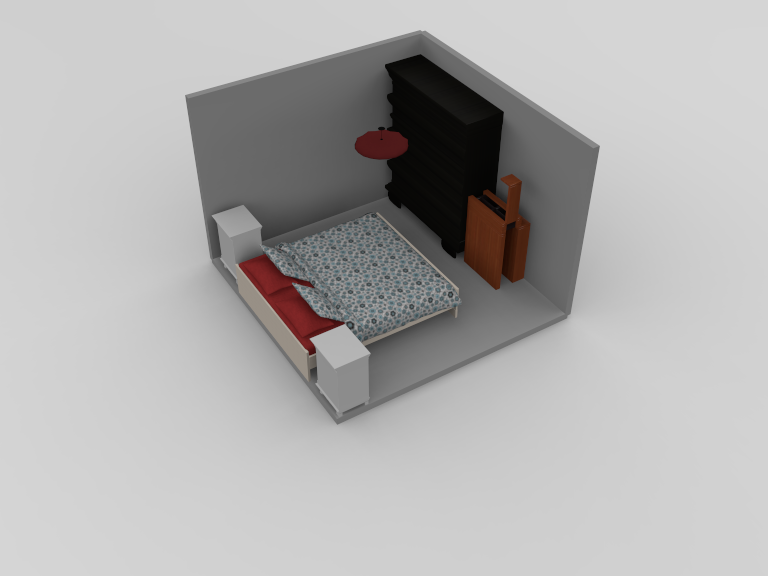}
        \includegraphics[width=0.32\textwidth, trim=80px 80px 80px 0px, clip]{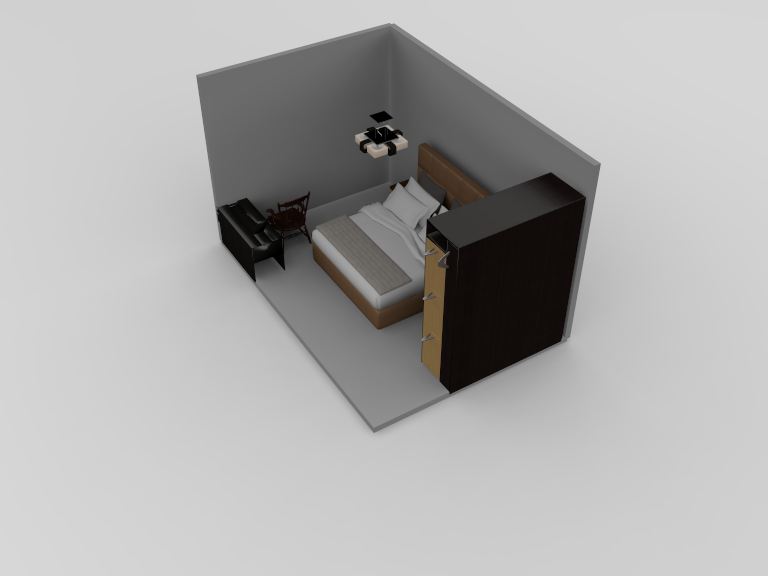}
        \caption[]{{\small 3D-FRONT Bedroom}}    
        \label{fig:gen:front:bed}
    \end{subfigure}
    \hfill
    \begin{subfigure}[b]{0.49\textwidth}
        \centering
        \includegraphics[width=0.32\textwidth, trim=80px 80px 80px 0px, clip]{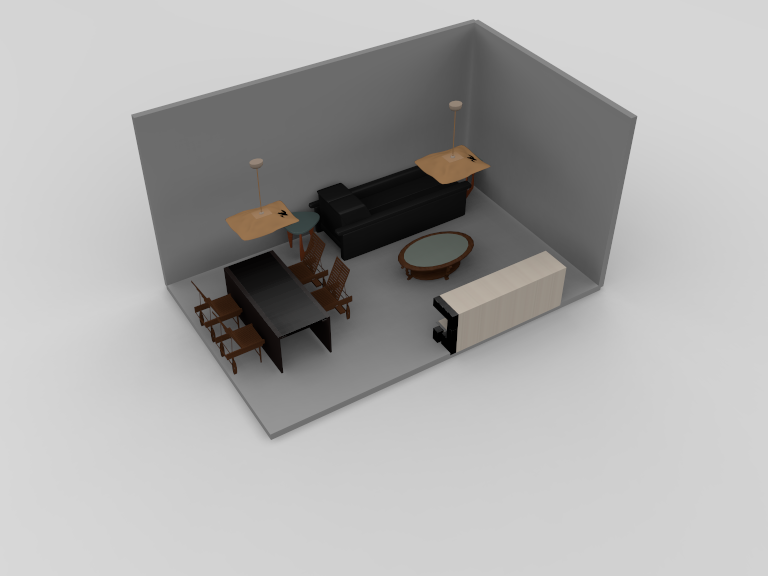}
        \includegraphics[width=0.32\textwidth, trim=80px 80px 80px 0px, clip]{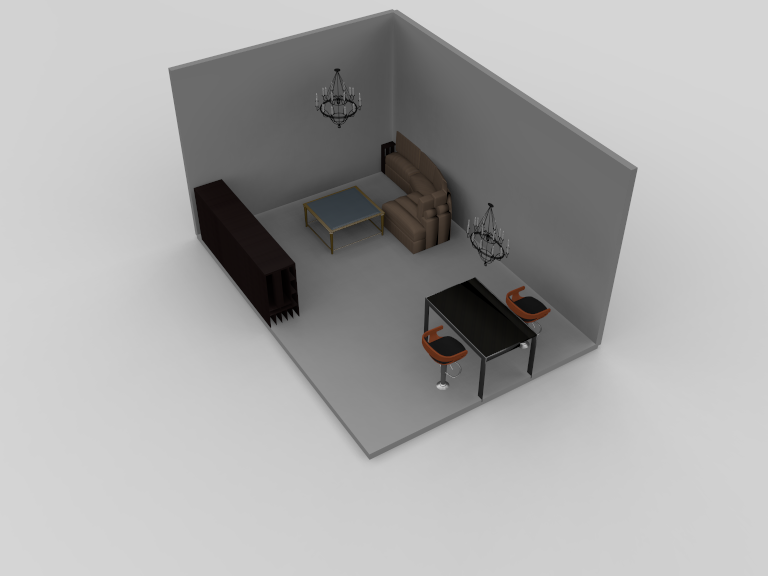}
        \includegraphics[width=0.32\textwidth, trim=80px 80px 80px 0px, clip]{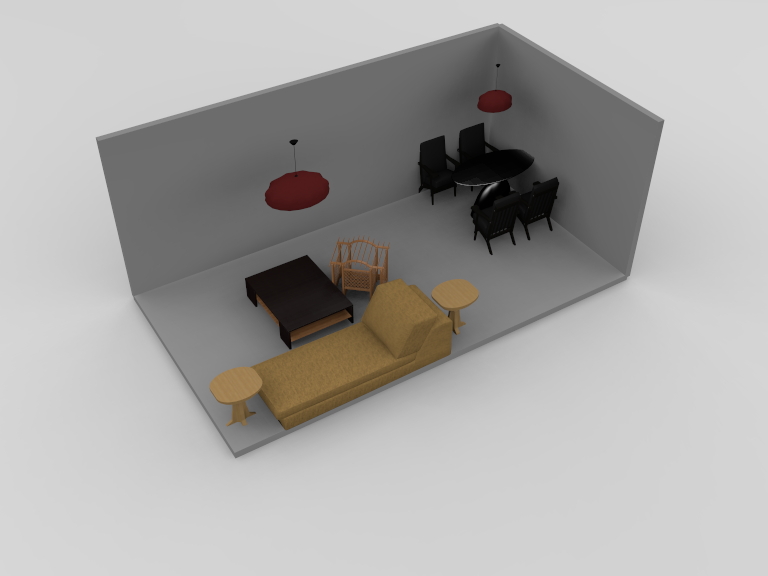}
        
        \includegraphics[width=0.32\textwidth, trim=80px 80px 80px 0px, clip]{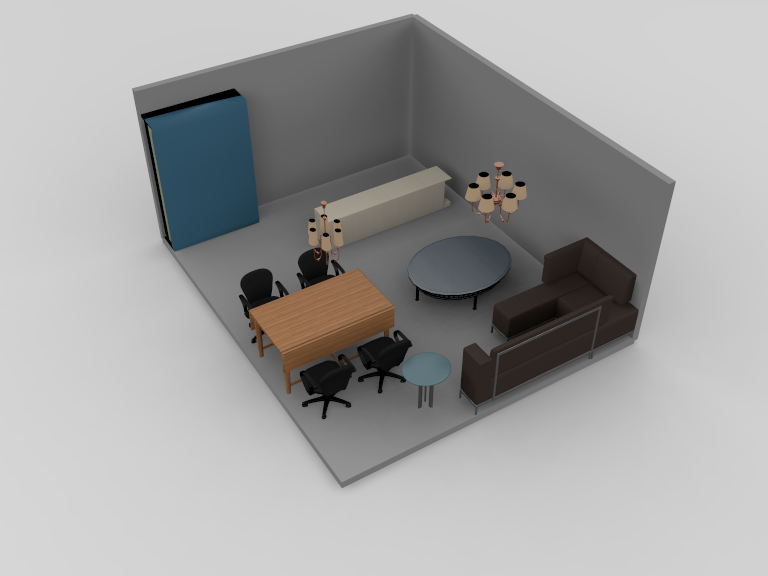}
        \includegraphics[width=0.32\textwidth, trim=80px 80px 80px 0px, clip]{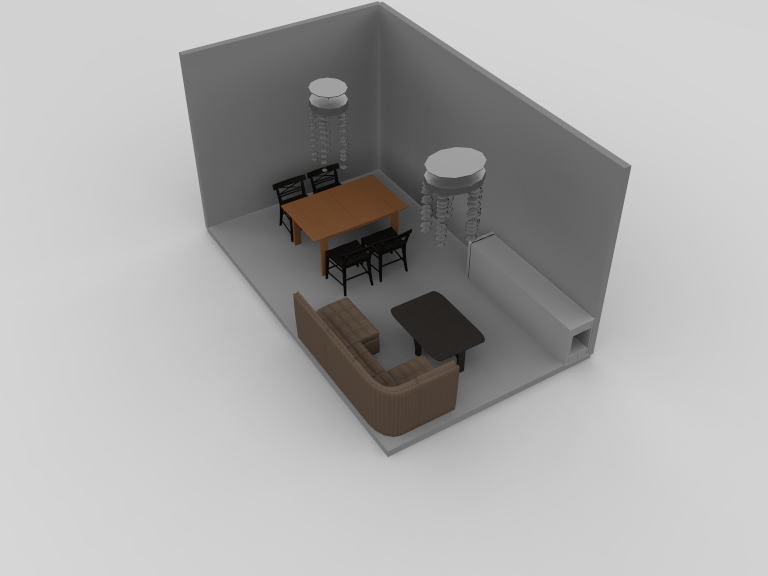}
        \includegraphics[width=0.32\textwidth, trim=80px 80px 80px 0px, clip]{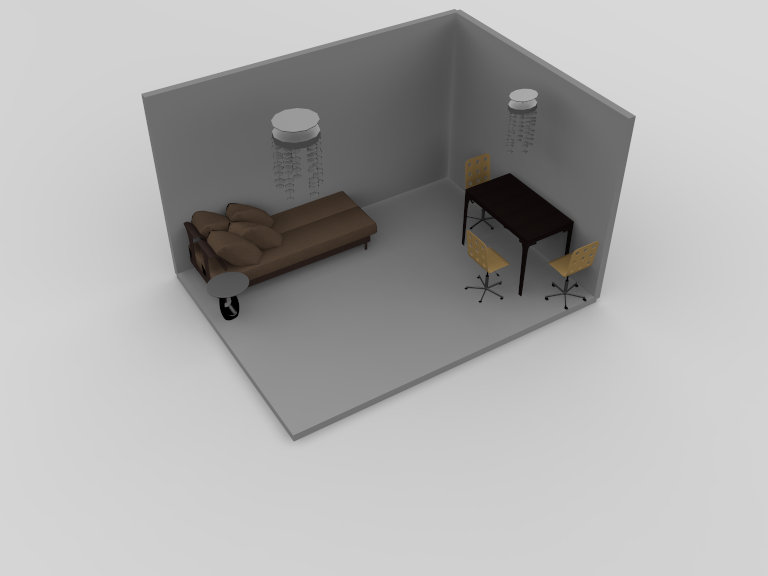}

        \includegraphics[width=0.32\textwidth, trim=80px 80px 80px 0px, clip]{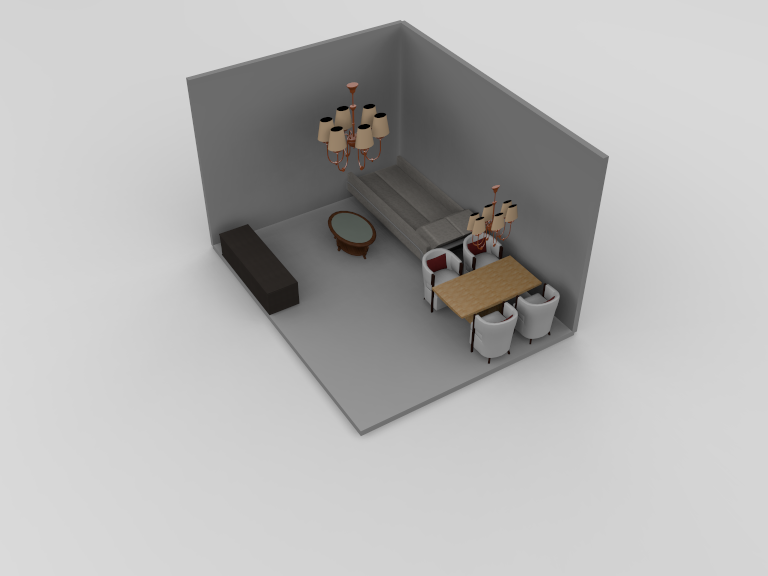}
        \includegraphics[width=0.32\textwidth, trim=80px 80px 80px 0px, clip]{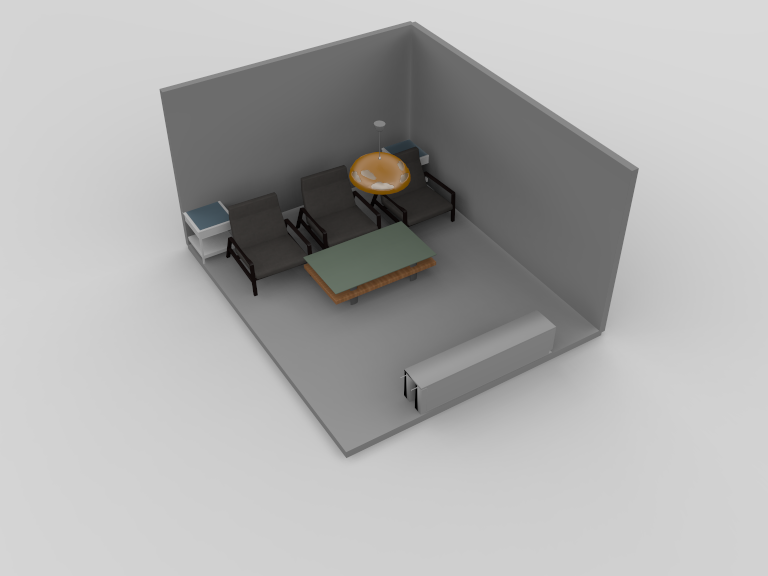}
        \includegraphics[width=0.32\textwidth, trim=80px 80px 80px 0px, clip]{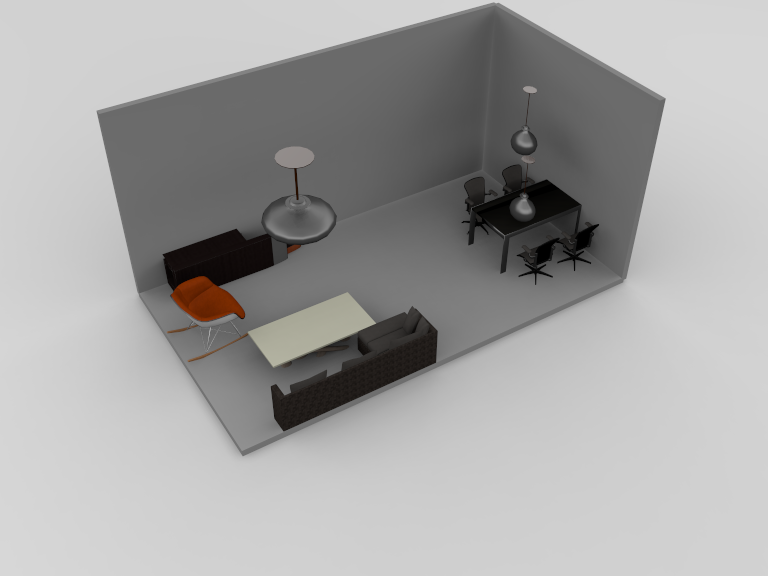}
        \caption[]{{\small 3D-FRONT Livingroom}}    
        \label{fig:gen:front:living}
    \end{subfigure}
\vspace{-7pt}
\caption{Randomly generated scenes using our method on 3D-FRONT~\protect\cite{Fu:2020:3dfront}.}
\label{Figure:Gen}
\vspace{-0.2in}
\end{figure*}


\subsection{Relative Attributes} 
\label{Subsec:Encoding:Relative:Parameterization}

Unlike object attributes, relative attributes are forced to capture patterns between pairs of objects, e.g., adjacent objects. Like object attributes, we encode the relative attributes $\abar_{e}$ of each edge $e = (v,v')$ as $\abar_{e} = (\bs{s}_e;\bs{r}_{e};\bs{t}_{e})$. Here $\bs{s}_{e}\in \R^9$ denotes the pairwise differences between the three scales of $\bs{s}_v$ and those of $\bs{s}_{v'}$. $\bs{r}_{e}\in R^3$ denotes the Euler angles of $v'$'s pose in the local coordinate system of $v$. $\bs{t}_{e}\in R^3$ denotes the center of $v'$ in the local coordinate system of $v$. The entire set of relative primitive parameters is encoded using a tensor $\overline{A}_{\set{E}} \in \R^{\sum_{c\in \set{C}}N_c\times \sum_{c\in \set{C}}N_c \times 15}$.

As shown in Figure~\ref{Fig:Shape:Geometry}, the network architecture of this module mimics the U-Net~\cite{UNet}. It is conceptually similar to an auto-encoder with two major differences. First, we do not sample the code space to synthesize relative attributes. Second, its input consists of relative attributes induced from predicted object attributes, which are expected to be noisy. The role of this U-Net is to produce rectified relative attributes. Please refer to the supplementary material for details.

\subsection{Network Training}
\label{Subsec:Training}

Training attribute synthesis modules extends the standard approach for training VAEs (c.f.~\cite{KingmaW13}). Let $h^{\phi}$ and $g_1^{\theta_1}$ be the encoder and decoder components of the VAE module for synthesizing object attributes. With $g_2^{\theta_2}$ we denote the U-Net module for synthesizing relative attributes. Here $\phi$ and $\theta = (\theta_1,\theta_2)$ denote the network parameters. Consider a training set $\set{T} = \{(\set{A}_{\set{V}},\set{A}_{\set{E}})\}$ where $\set{A}_{\set{V}}$ and $\set{A}_{\set{E}}$ denote encoded object attributes and relative attributes, respectively. We solve the following optimization problem to determine the optimized network weights $\phi$ and $\theta$:
\begin{align}
& \min\limits_{\phi,\theta}  \frac{1}{|\set{T}|}\sum\limits_{(\set{A}_{\set{V}},\set{A}_{\set{E}})\in \set{T}}\Big(\lambda_{\set{E}}\|g_2^{\theta_2}(g_1^{\theta_1}( h^{\phi}(\overline{A}_{\set{V}})))-\overline{A}_{\set{E}}\|^2 \nonumber \\
& \quad + \|g_1^{\theta_1}( h^{\phi}(\overline{A}_{\set{V}}))-\overline{A}_{\set{V}}\|^2\Big) + \lambda_{KL}KL\big(\{h^{\phi}(\overline{A}_{\set{V}})\}|\set{N}_d) \nonumber 
\end{align}
where $\set{N}_d$ is the normal distribution associated with the latent space of the object attribute VAE. The same as ~\cite{KingmaW13}, the last term forces the latent codes of the training instances to match $\set{N}_d$. This paper sets $\lambda_{\set{E}} = 1$ and $\lambda_{KL} = 0.01$. We use ADAM~\cite{KingmaB14} for network training.

\section{Results}
\label{sec:result}

This section presents an experimental evaluation of our approach. In Section~\ref{subsec:setup}, we describe the experimental setup. In Section~\ref{subsec:results}, we demonstrate the results of our approach and compare it with baseline methods. We analyze each component of our approach in Section~\ref{subsec:analysis}. Please refer to the supplementary material for more results and baseline comparisons. 

\subsection{Experimental Setup}
\label{subsec:setup}

\paragraph{Dataset.}
We perform experimental evaluation on the new large-scale 3D scene dataset 3D-FRONT~\cite{Fu:2020:3dfront}. We also include the results on SUNCG~\cite{Song_2017_CVPR_SUNCG}, on which most works have evaluated. Following~\cite{ZhangYMLHVH20}, we consider bedrooms and living rooms in both datasets and train scene synthesis models from each room type in each dataset. For all datasets, each room type contains 4000 training scenes and the maximum number of objects per class is four. Each room type contains 30 object classses for the SUNCG dataset and 20 object classes for the 3D-FRONT dataset. More details of the datasets are in the supplementary material.

\vspace{-5pt}
\paragraph{Baseline approaches.}
We consider five baseline approaches D-Prior~\cite{wang2018deep}, Fast~\cite{ritchie2019fast}, PlanIT~\cite{wang2019planit}, GRAINS~\cite{Li:2019:GRAINS}, and D-Gen~\cite{ZhangYMLHVH20}. They represent state-of-the-art in 3D scene synthesis.

\vspace{-5pt}
\paragraph{Evaluation metrics.}
We consider two ways to evaluate scene synthesis approaches. The first is a perceptual study, where we employed 10 non-experts to judge the visual quality of 100 synthesized 3D scenes. We compare our results with those of each method and count the percentage of our results that are more plausible than each baseline. The second metric assesses distributions of relative attributes of the generated scenes (c.f~\cite{ZhangYMLHVH20}).  

\subsection{Analysis of the Results}
\label{subsec:results}

Figure~\ref{Figure:Gen} shows randomly generated scenes
using our approach. We can see that the generated scenes contain rich sets of objects, meaning our approach can generate complex scenes. The scene layouts are highly plausible from multiple perspectives, including the shape of each object, the spatial relations among multiple objects, and object co-occurrence.   Moreover, the generated scenes are diverse. Figure~\ref{Figure:Teaser} shows that our generated scenes are different from the closest scenes in the training set. These results show that our approach captures diverse feature patterns of scenes and exhibits strong generalization ability. 

Table~\ref{tab:user_study} provides a perceptual study between our approach and baseline approaches. We can see that our approach outperforms all baseline approaches considerably (the top performing baseline~\cite{wang2019planit} utilizes additional inputs). The improvements are consistent across all the categories. Moreover, our approach is even competitive against the visual quality of the 3D scenes in the training set. These statistics demonstrate the superior performance of our approach. Please refer to the supplementary material for visual comparisons between our approach and baseline approaches. 

Currently, synthesizing one scene takes 2$\sim$5 seconds on a desktop with a 3.2G Hz CPU, 32G main memory, and a 2080 Ti GPU. The majority of the computation is spent on scene optimization, which runs on the CPU.

\begin{figure*}
    \begin{subfigure}[b]{0.49\textwidth}
        \centering
        \includegraphics[width=0.32\textwidth, trim=80px 80px 80px 0px, clip]{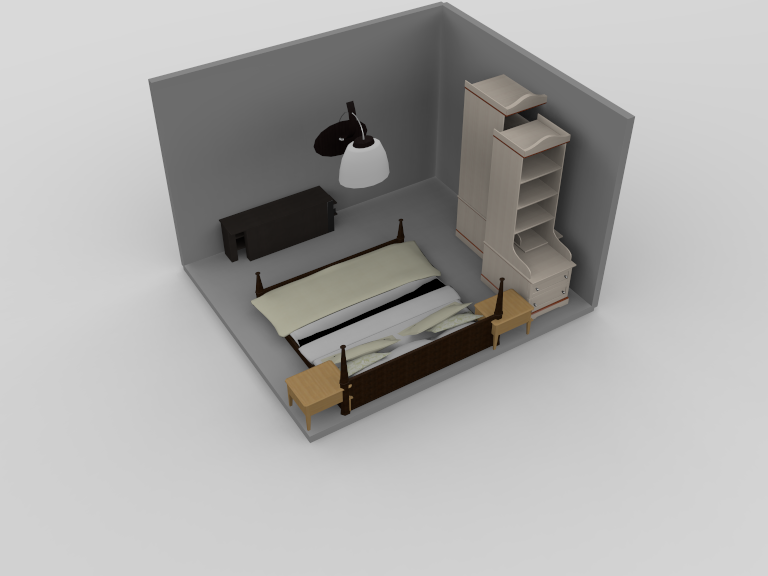}
        \includegraphics[width=0.32\textwidth, trim=80px 80px 80px 0px, clip]{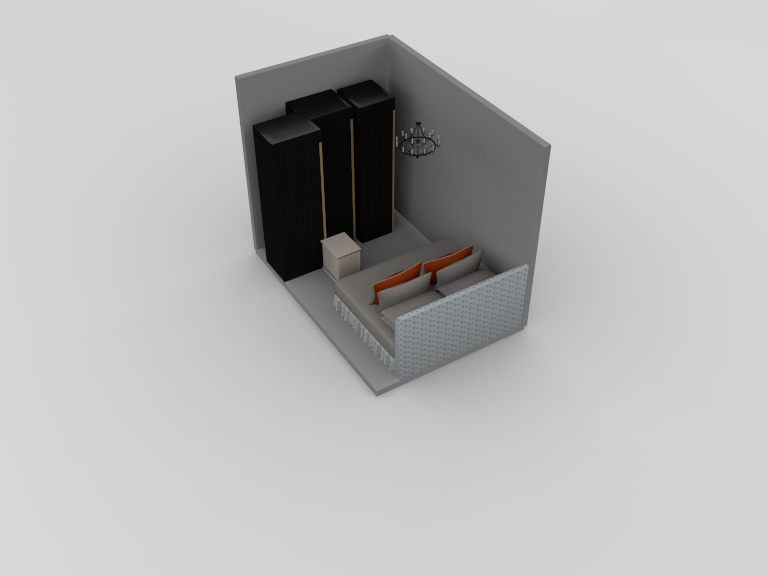}
        \includegraphics[width=0.32\textwidth, trim=80px 80px 80px 0px, clip]{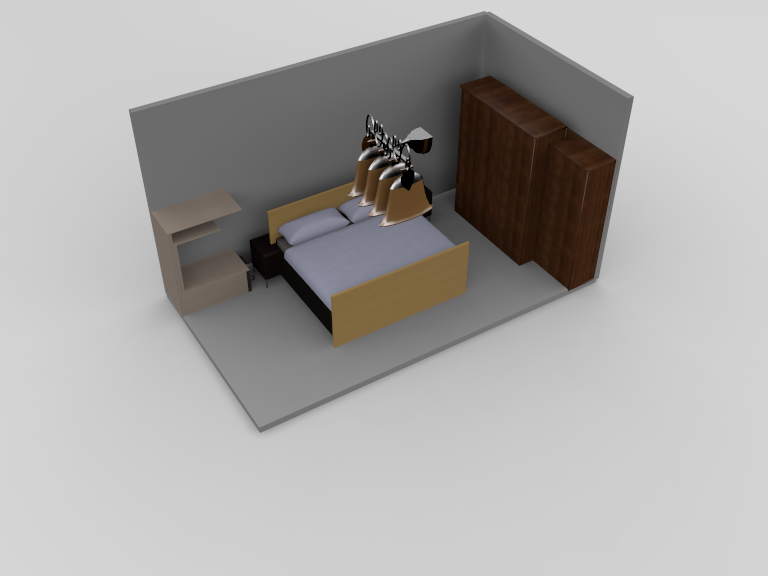}
        
        \includegraphics[width=0.32\textwidth, trim=80px 80px 80px 0px, clip]{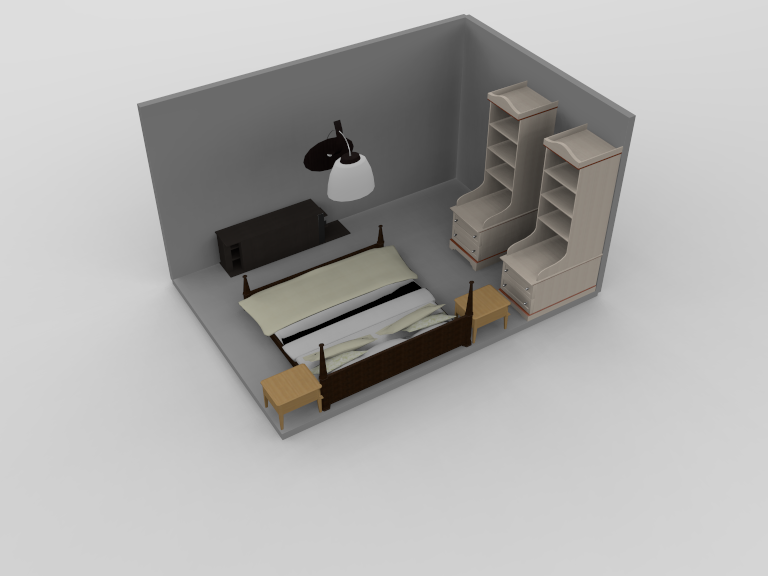}
        \includegraphics[width=0.32\textwidth, trim=80px 80px 80px 0px, clip]{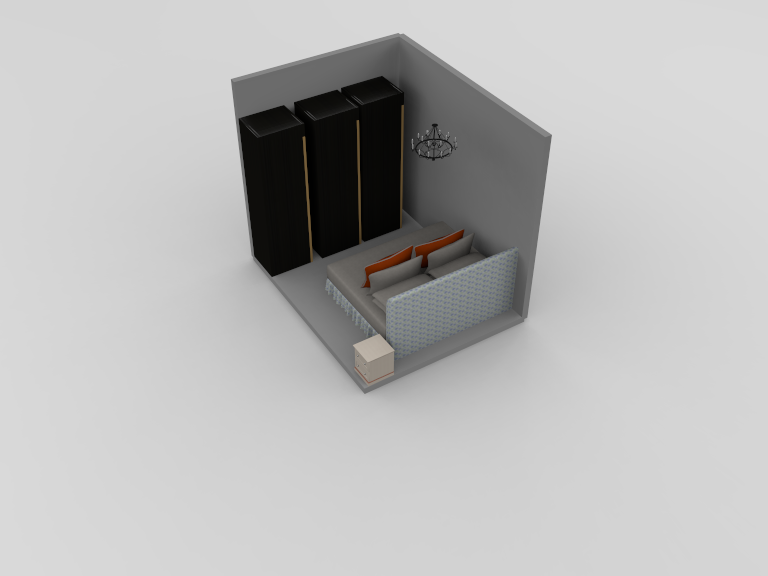}
        \includegraphics[width=0.32\textwidth, trim=80px 80px 80px 0px, clip]{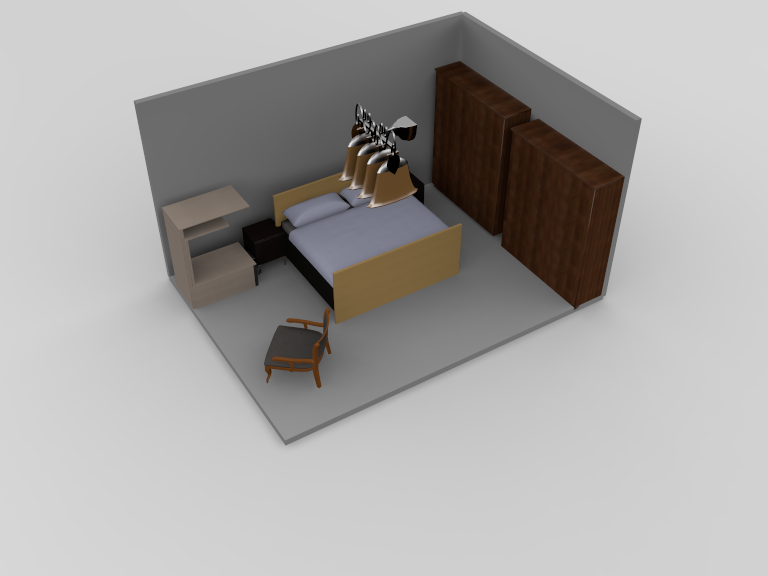}
        \caption[]{{\small 3D-FRONT Bedroom}}    
        \label{fig:sync:front:bed}
    \end{subfigure}
    \hfill
    \begin{subfigure}[b]{0.49\textwidth}
        \centering
        \includegraphics[width=0.32\textwidth, trim=80px 80px 80px 0px, clip]{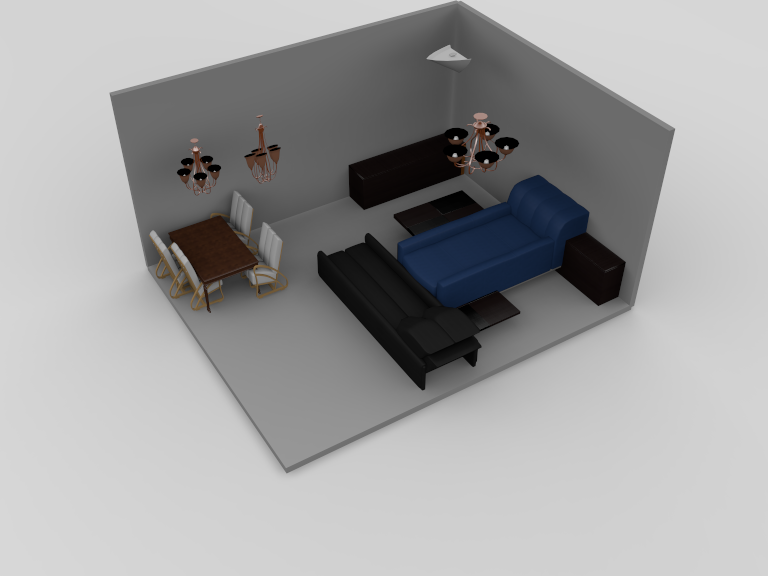}
        \includegraphics[width=0.32\textwidth, trim=80px 80px 80px 0px, clip]{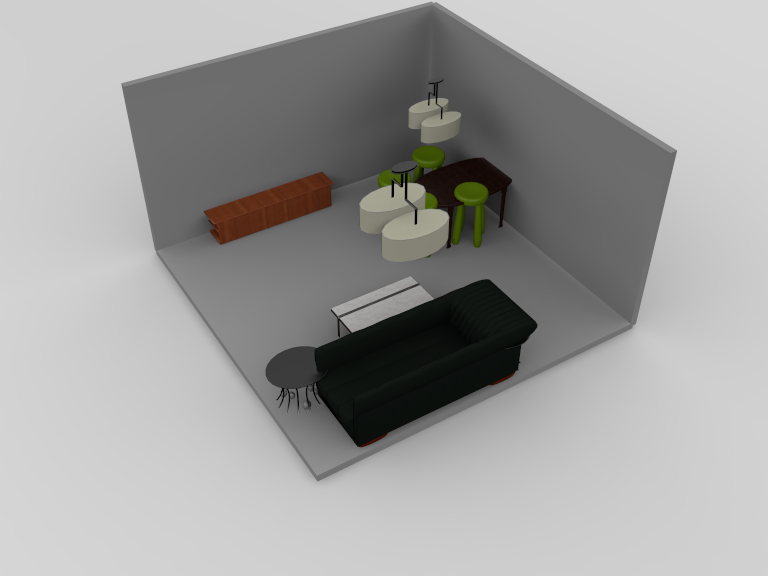}
        \includegraphics[width=0.32\textwidth, trim=80px 80px 80px 0px, clip]{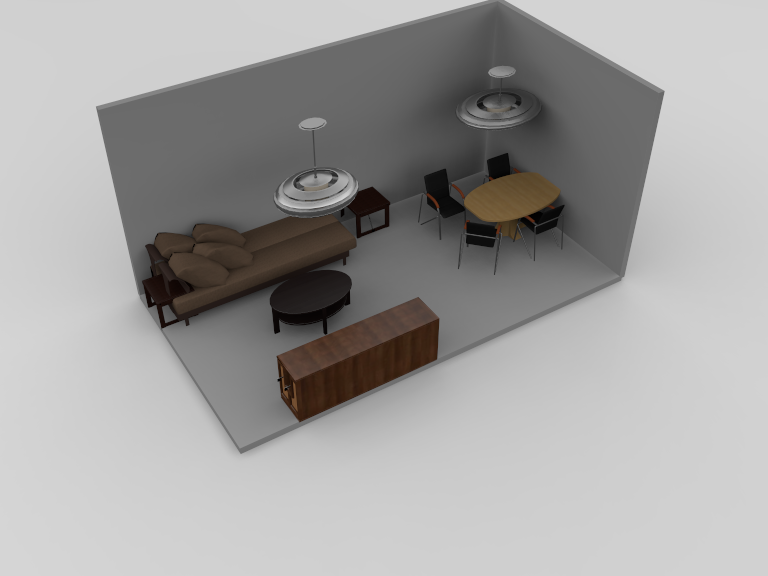}
        
        \includegraphics[width=0.32\textwidth, trim=80px 80px 80px 0px, clip]{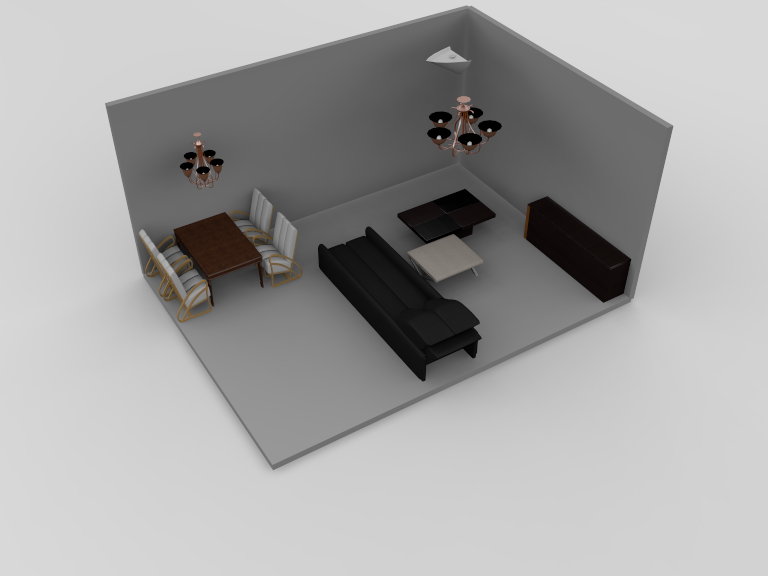}
        \includegraphics[width=0.32\textwidth, trim=80px 80px 80px 0px, clip]{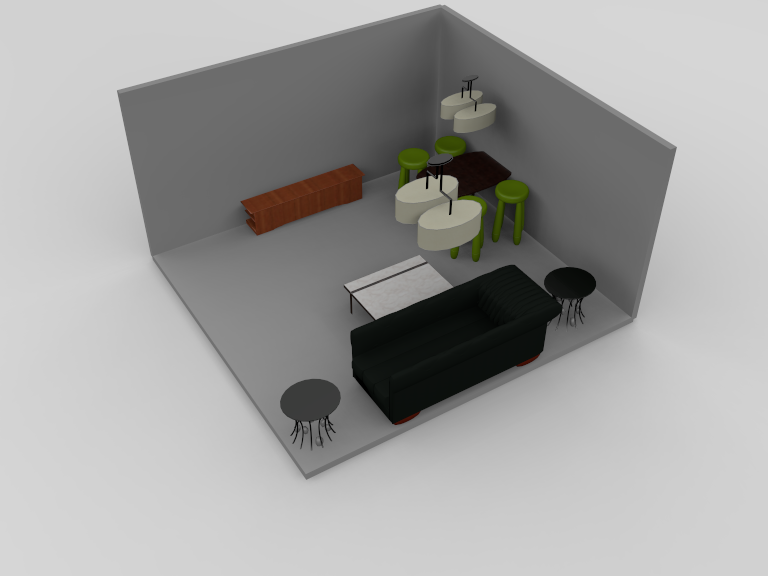}
        \includegraphics[width=0.32\textwidth, trim=80px 80px 80px 0px, clip]{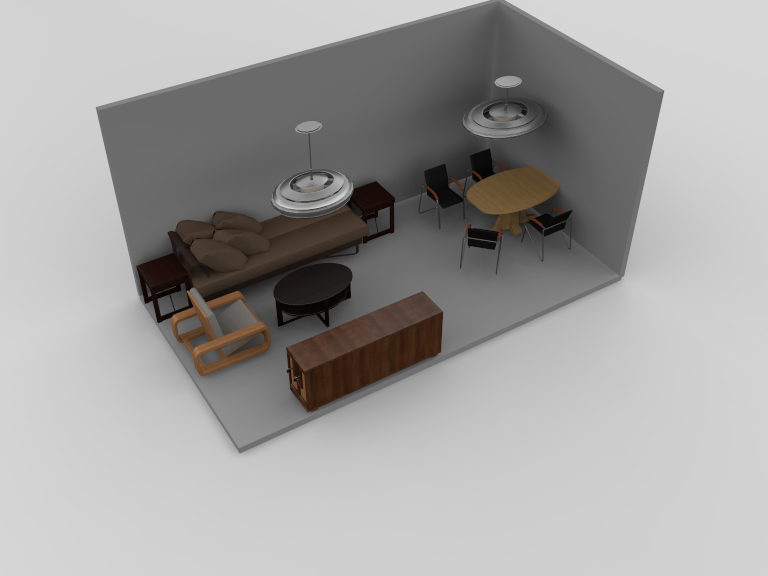}
        \caption[]{{\small 3D-FRONT Livingroom}}    
        \label{fig:sync:front:living}
    \end{subfigure}
\vspace{-0.1in}
\caption{Visual comparisons between output of the prediction module (top) and output of the synchronization module (bottom).}
\label{Figure:Sync:Ablation}
\vspace{-0.1in}
\end{figure*}

\begin{figure*}
\includegraphics[width=1.0\textwidth]{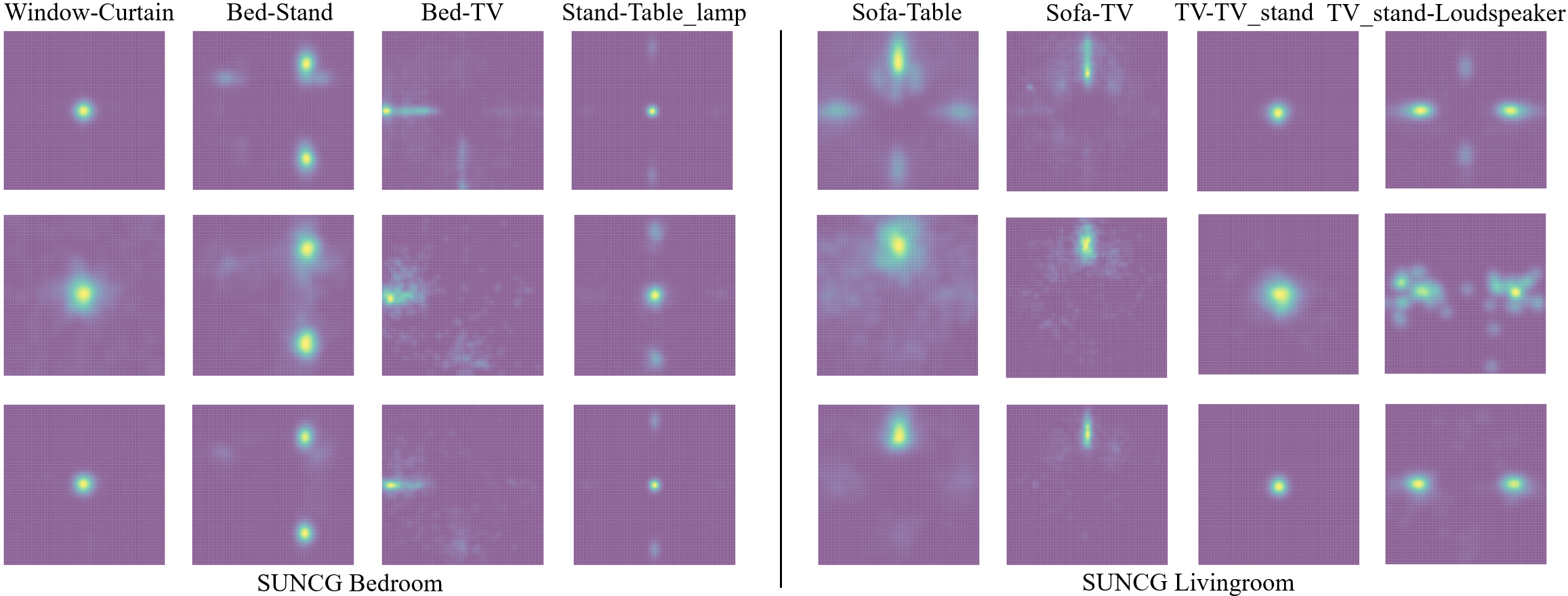}
\vspace{-20pt}
\caption{Distributions of relative translation. Top: distribution of the training data. Middle: distribution derived from predicted absolute parameters. Bottom: distribution derived from the optimized absolute parameters after synchronization.}
\label{Figure:Relative:Prediction}
\vspace{-0.1in}
\end{figure*}

\setlength\tabcolsep{0.5pt}
\begin{table}[]
    \footnotesize
    \centering
    \begin{tabular}{c|c|c|c|c|c|c}
     & \multicolumn{6}{c}{\textbf{Ours vs. other methods}}\\
    Room Type & D-Prior & Fast & GRAINS & PlanIT & D-Gen & GT\\
    \hline
    SUN-bed & 56.7$\pm$6.3 & 53.6$\pm$5.8 & 65.4$\pm$4.1 & 52.5$\pm$6.8 & 58.8$\pm$4.6 & 43.8$\pm$5.2\\
    SUN-living & 55.2$\pm$4.9 & 52.1$\pm$4.9 & 88.3$\pm$5.2 & 51.1$\pm$4.7 & 57.1$\pm$5.1 &44.2$\pm$4.8\\
    3DF-bed & 58.1$\pm$4.3 & 56.7$\pm$5.9 & 60.1$\pm$6.3 & 54.4$\pm$5.3 & 54.5$\pm$6.3 &49.7$\pm$5.3\\
    3DF-living & 72.8$\pm$6.4 & 73.1$\pm$4.4 & 89.2$\pm$5.1 & 68.9$\pm$6.6 & 64.3$\pm$5.9 &47.1$\pm$4.5\\
    \end{tabular}
    \vspace{-0.1in}
    \caption{Percentage ($\pm$ standard error) of forced-choice comparisons in which scenes generated by our method are judged as more plausible than scenes from another source. Higher is better. Our approach consistently outperforms baseline approaches. }
    \label{tab:user_study}
\vspace{-0.2in}
\end{table}

\subsection{Analysis of Our Approach}
\label{subsec:analysis}

We proceed to analyze the benefits of utilizing relative attributions and prior distributions. As shown in Figure~\ref{Figure:Scene:Opt:Effects}, using relative attributes can alleviate the issue of conflicting object attributes such as the relative poses between nightstands and beds. The optimized object locations are more plausible than those from the synthesized object attributes. However, it does not fully address issues such as penetrating objects and redundant objects. Imposing the prior distribution improves the object layout considerably, leading to penetration-free and less-crowded scene layouts. 

Figure~\ref{Figure:Sync:Ablation} shows additional visual comparisons between the synthesized attributes and the output of scene optimization. Again, we can see that the improvements are multi-faceted. Our approach improves the objects' locations and adds and deletes objects properly to exhibit more plausible object co-occurrences. Such improvements are justified in Figure~\ref{Figure:Relative:Prediction}, which compares the distributions of relative attributes induced from synthesized object attributions and scene optimization counterparts. We can see that those obtained from scene optimization match the underlying ground truth more closely than those induced from the synthesized object attributes. Such improvements come from relative attributes and modeling both the continuous prior distributions such as relative poses and discrete prior distributions such as joint distributions of object count pairs. 

In the supplementary material, we provide results to show that even with the scene optimization framework, synthesizing relative attributes is critical. This is because the prior distributions have rich local minimums. Utilizing the relative attributes enables us to obtain better initial solutions for scene optimization. 

Scene optimization can change the scene layout considerably, e.g., adding/removing objects. The supplementary material also shows that it is challenging to achieve similar results using off-the-shelf scene optimization techniques, e.g.,~\cite{Yu:2011:MIH}. We can understand this as relative attributes, which are unavailable in other techniques, serve as a critical information source for our approach. 

\section{Conclusions and Limitations}
\label{sec:conclusion}

We have shown that predicting relative attributes and learning prior distributions provide effective regularizations for synthesizing 3D scenes. The generated results preserve diverse feature patterns of 3D scenes and exhibit strong generalization behavior. Experimental results show that our approach outperforms prior state-of-the-art scene synthesis techniques. 

Our approach has a couple of limitations. First, unlike end-to-end synthesis, our approach employs an optimization module to generate the final output. Therefore, computing the derivatives between the final output and the latent parameter (e.g., using the implicit function theorem) becomes more complex than end-to-end networks. One way to address this issue is to realize scene optimization using a graph neural network. Another limitation of our approach is that it does not enforce symmetries among objects, which are available in indoor scenes. An interesting question is how to detect and enforce symmetries during scene optimization automatically. 

\vspace{-10pt}
\paragraph{Acknowledgements.}
 Chandrajit Bajaj would like to acknowledge the support from NIH-R01GM117594, by the Peter O’Donnell Foundation, and in part from a grant from the Army Research Office accomplished under Cooperative Agreement Number W911NF-19-2-0333. Qixing Huang would like to acknowledge the support from NSF Career IIS-2047677 and NSF HDR TRIPODS-1934932.


{\small
\bibliographystyle{ieee_fullname}
\bibliography{paper}
}
\newpage
\clearpage
\appendix


Our supplementary material provides details of the functions in prior modeling and joint hyperparameter optimization in Section \ref{sec:supp:function}, additional details on datasets and network architecture in Section \ref{sec:supp:network}, as well as more experiment results and analysis in Section \ref{sec:supp:results}.

\section{Details of Modeling the Objective Function for Synchronization}
\label{sec:supp:function}

\subsection{GGMM in Prior Modeling}
We mainly focus on translation when modeling the prior $P_c(\bs{a}_v)$ and $P_{(c, c')}(\phi(\bs{a}_v,\bs{a}_{v'}))$ because we notice that the rotation and size predicted by the network are accurate and can be directly utilized in synchronization. We model the prior for each dimension of the translation separately. A 6-component GMM is utilized to model $\GMM_{\mu_{c}}(\bs{a}_v)$ for each class. 

$P_{(c,c')}(\phi(\bs{a}_v,\bs{a}_{v'}))$ is modeled by an 8-component GMM multiplied by a mask which equals to one when objects of class $c$ and class $c'$ could penetrate (e.g., a chair and a desk) and $I_{(c, c')}(\bs{a}_v,\bs{a}_{v'})$ otherwise, where
\begin{align}
I_{(c, c')}(\bs{a}_v,\bs{a}_{v'}) = 
\exp(\text{max}\{0, \frac{\bs{s}_v + \bs{s}_{v'}}{2}-|\bs{t}_{v'} - \bs{t}_v|\})
\label{Eq:GMM:Mask}
\end{align}

We use a 2-component GMM to fit $P_c(\bs{z}_{\set{V}_c})$ and a 4-component GMM to fit $P_{(c, c')}(\bs{z}_{\set{V}_c}, \bs{z}_{\set{V}_{c'}})$.

\begin{figure}
\centering
    \begin{tabular}{cc}
        \includegraphics[width=0.22\textwidth, trim={20px 10px 30px 20px}, clip ]{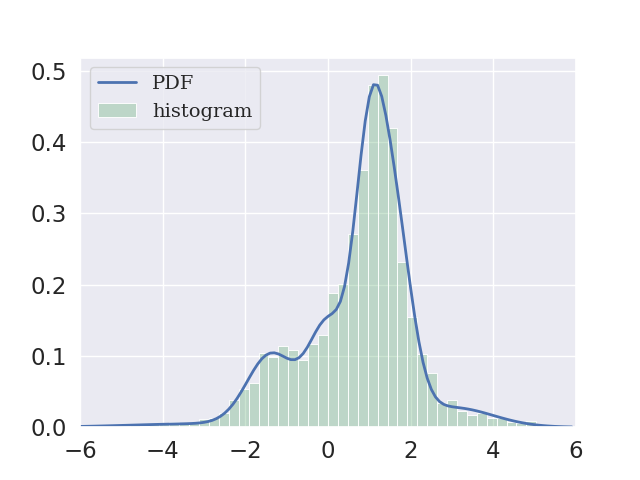} &
        \includegraphics[width=0.22\textwidth, trim={20px 10px 30px 20px}, clip ]{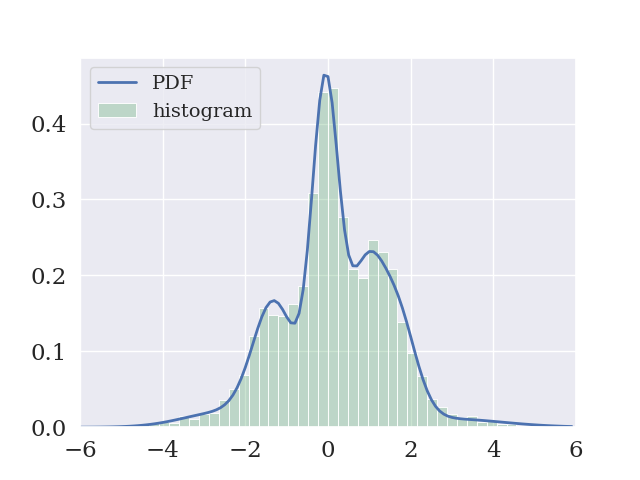} \\
        \includegraphics[width=0.22\textwidth, trim={20px 10px 30px 20px}, clip ]{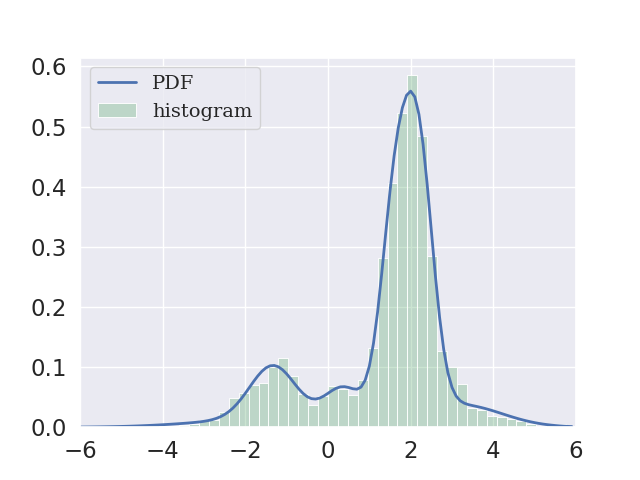} &
        \includegraphics[width=0.22\textwidth, trim={20px 10px 30px 20px}, clip ]{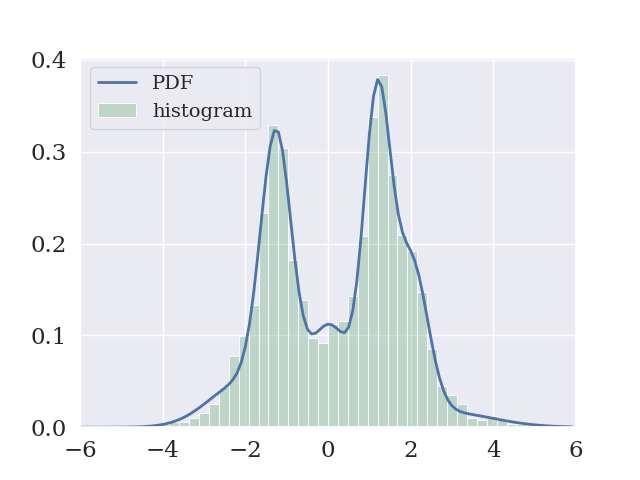} \\
        \includegraphics[width=0.22\textwidth, trim={20px 10px 30px 20px}, clip ]{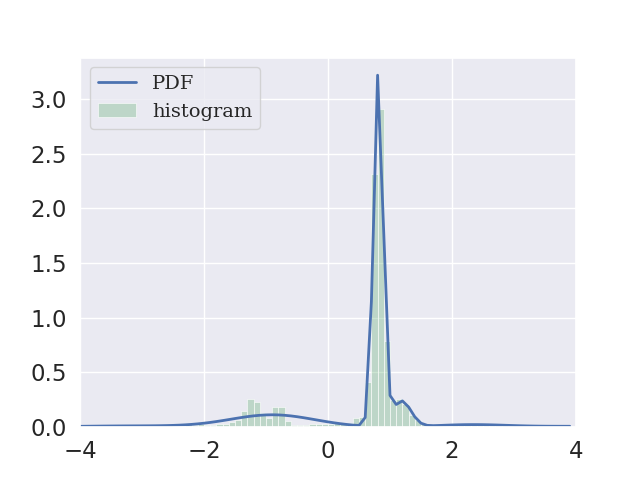} &
        \includegraphics[width=0.22\textwidth, trim={20px 10px 30px 20px}, clip ]{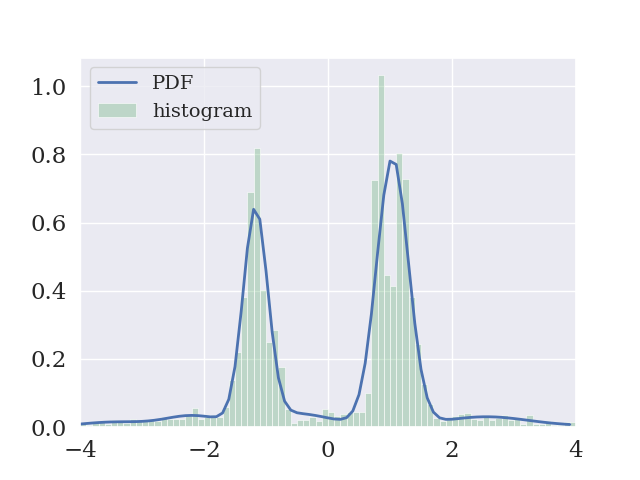} \\
    \end{tabular}
\caption{Distribution of translation in the SUNCG dataset. Top: absolute translation of the bed. Middle: absolute translation of the stand. Bottom: relative translation between the bed and the stand. Left: x coordinates. Right: y coordinates.}
\label{Figure:Supp:Prior:Translation}
\end{figure}

\begin{figure}
\centering
    \begin{tabular}{cc}
        \includegraphics[width=0.22\textwidth, scale=0.8, trim={20px 10px 30px 20px}, clip ]{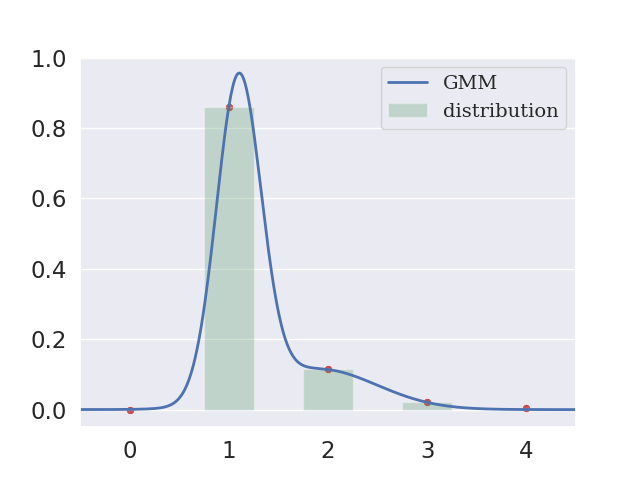} &
        \includegraphics[width=0.22\textwidth, scale=0.8, trim={20px 10px 30px 20px}, clip ]{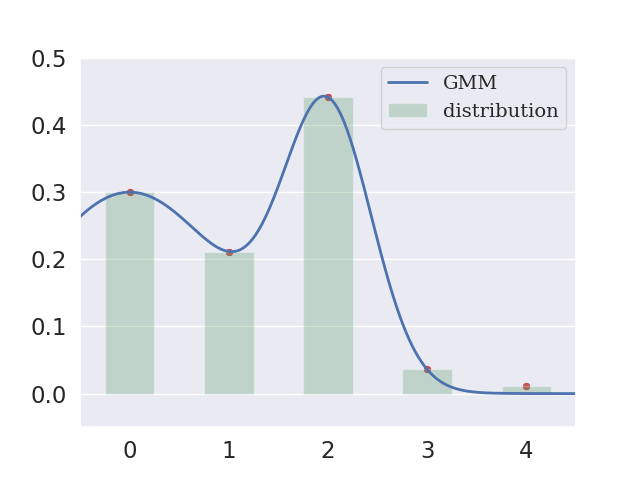} 
    \end{tabular}
\caption{Modeling $P_c(\bs{z}_{\set{V}_c})$ with GMM in the SUNCG bedroom dataset. Left: the bed. Right: the stand. From the statistical distribution, it is most likely that there exist 1 bed and 2 stands in each scene.}
\label{Figure:Supp:Prior:Pxv}
\end{figure}

\begin{figure}
\centering
    \begin{tabular}{cc}
        \includegraphics[width=0.23\textwidth, trim={10px 0px 30px 20px}, clip ]{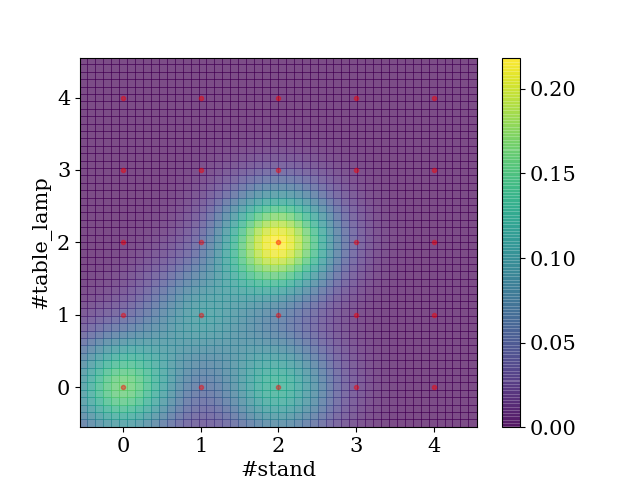} &
        \includegraphics[width=0.23\textwidth, trim={10px 0px 30px 20px}, clip ]{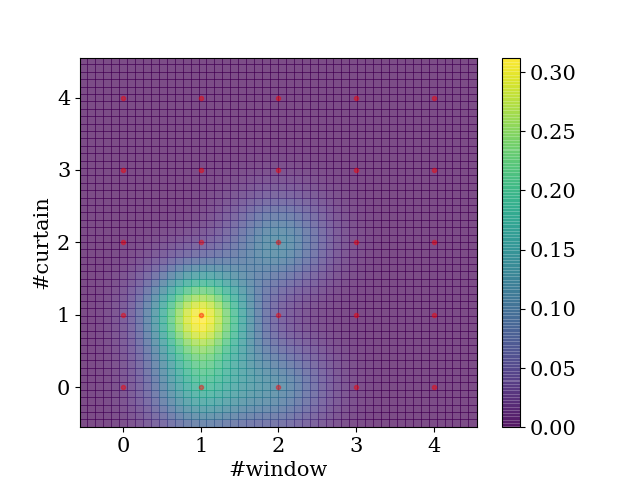} 
    \end{tabular}
\caption{Modeling $P_{(c, c')}(\bs{z}_{\set{V}_c}, \bs{z}_{\set{V}_{c'}})$ with GMM in the SUNCG bedroom dataset. Left: the stand and the table lamp. It is most likely that the number of stands and that of table lamps are equal, i.e. they co-exist. Right: the window and the curtain.}
\label{Figure:Supp:Prior:Pxe}
\end{figure}

Figure~\ref{Figure:Supp:Prior:Translation} shows an example of the distribution of translation. Figures~\ref{Figure:Supp:Prior:Pxv} and \ref{Figure:Supp:Prior:Pxe} show examples of the distribution of numbers of objects.

\begin{figure*}
\includegraphics[width=0.95\textwidth]{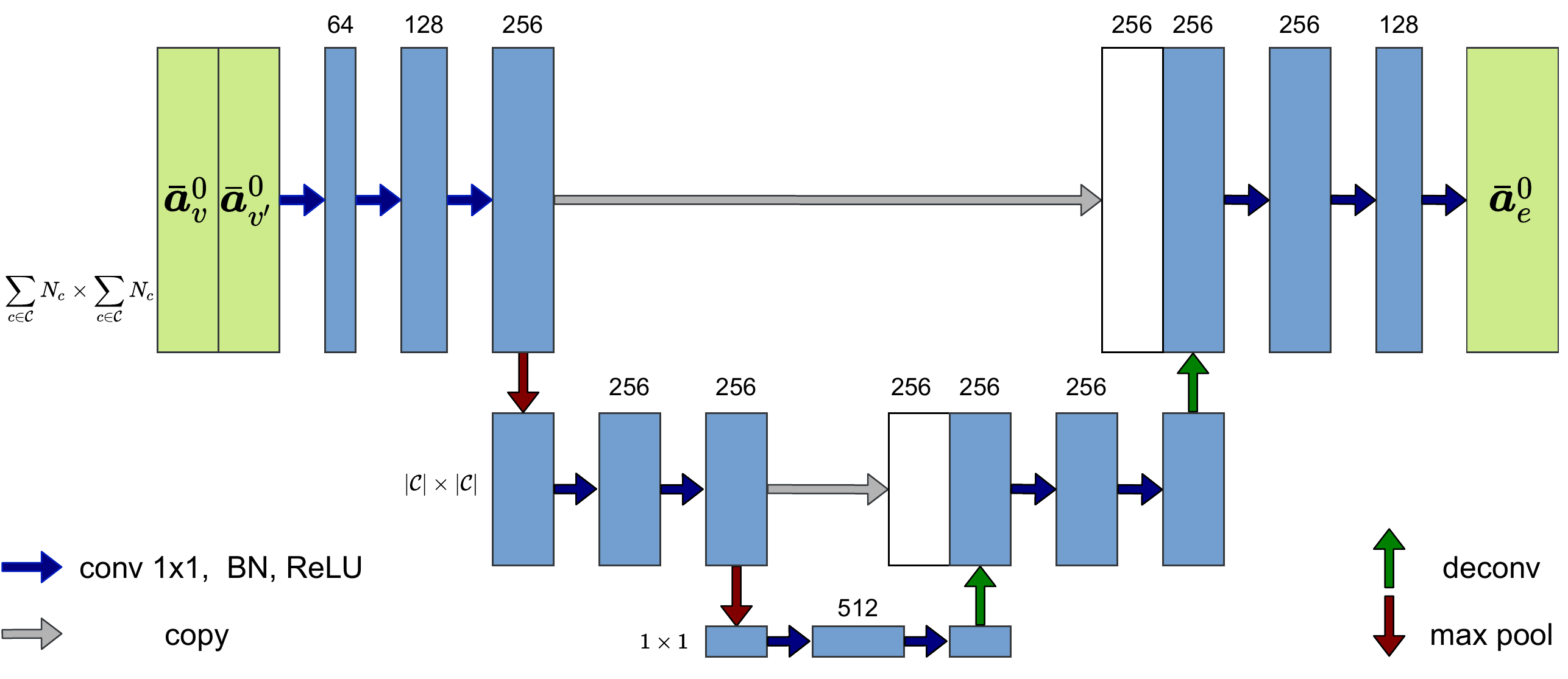}
\centering
\caption{Architecture of the relative attribute prediction network.}
\label{Figure:Network:Relnet}
\end{figure*}




\subsection{Joint Hyperparameter Optimization}
The regularization term $l(\Phi)$ is designed to learn hyper-parameters of the prior distribution from the training set $\set{T}_{train}:=\{\bs{t}_{\set{V}_c}, \bs{t}_{\set{E}_{(c, c')}}\}$, where $\bs{t}_{\set{V}_c}$ collects absolute translation of vertices belonging to the class $c$, $\bs{t}_{\set{E}_{(c, c')}}$ collects relative translation of vertex pairs belonging to class $c$ and $c'$. To learn the translation prior term, we utilize $l_1(\Phi)$ based on maximum likelihood estimation:
\begin{equation}
\begin{aligned}
l_1(\Phi) = &  -\sum_{c\in \set{C}}\sum_{\bs{t}_{v}\in \bs{t}_{\set{V}_c}} \log \GMM_{\mu_c} (\bs{t}_{v}) \nonumber \\
& - \sum_{c, c'\in \set{C}} \sum_{\bs{t}_{e} \in \bs{t}_{\set{E}_{(c, c')}}} \log \GMM_{\mu_{(c, c')}} (\bs{t}_{e})
\label{Eq:Supp:Prior:Translation}
\end{aligned}
\end{equation}

For object count prior, we first compute the discrete distribution $p_c$ and $p_{(c, c')}$ from the dataset.
Here
$$
p_c(i) = n_{c,i}/n_{scenes}, \quad i\in \{0,1,\cdots, N_c\}
$$
where $n_{scenes}$ is the total number of scenes, and $n_{c,i}$ is the total number of scenes with $i$ objects of class $c$.
Likewise,
$$
p_{c,c'}(i,j) = n_{c,c',i,j}/n_{scenes}, \quad i,j\in \{0,1,\cdots, N_c\}
$$
where $n_{c,c',i,j}$ is the number of scenes with $i$ objects of class $c$ and $j$ objects of class $c'$.
Then we fit the discrete distribution with continuous function using least square:
\begin{align}
& l_2(\Phi) = \sum_{c\in \set{C}} \sum_{\{(i, p_c(i))\}}\| \GMM_{\gamma_c} (i) - p_c(i) \|^2 + \nonumber \\
&  \sum_{c,c' \in \set{C}} \sum_{\{((i, j),  p_{(c, c')}(i, j))\}}\| \GMM_{\gamma_{(c, c')}} ((i, j)) - p_{(c, c')}(i, j)) \|^2 
\label{Eq:Supp:Prior:Indicator}
\end{align}

The total regularization term:
\begin{align}
l(\Phi) = l_1(\Phi) + l_2(\Phi)
\end{align}

\section{Dataset and Network Architecture}
\label{sec:supp:network}

\subsection{Preprocessing of Datasets}
For the 3D-FRONT dataset, we select two scene types: bedroom (Bedroom, MasterBedRoom and SecondBedRoom) and living room (LivingRoom and LivingDivingRoom). We fisrt filter out scenes with width and length larger than 8 meters. Then for bedroom and living room, we remove scenes whose number of objects is smaller than 6 and 4 respectively. Finally, for both scene types, we randomly sample 4000 scenes for training and about 100 scenes for validation. For the SUNCG dataset, we perform joint scene alignment following \cite{ZhangYMLHVH20} and use 1000 scenes for validation.

\subsection{Network Architecture Details}

The network architecture for the relative attribute prediction module is shown in Figure \ref{Figure:Network:Relnet}. Every pair of absolute attributes $\abar_v^0$ and $\abar_{v'}^0$ is concatenated as input to the network. The network outputs the refined relative attributes.

\begin{figure*}
\includegraphics[width=1.0\textwidth]{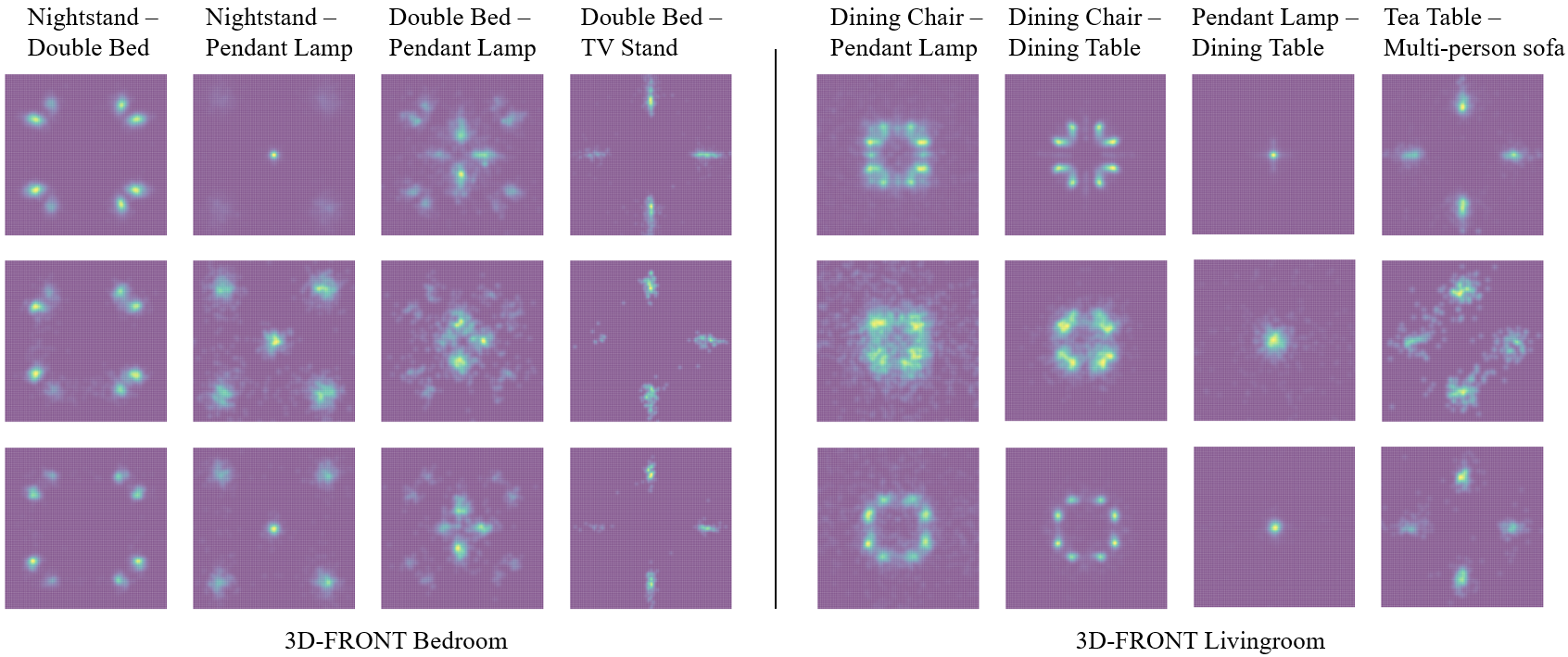}
\caption{Distributions of relative translation in the 3D-FRONT dataset. Top: distribution of the training data. Middle: distribution derived from predicted absolute parameters. Bottom: distribution derived from the optimized absolute parameters after synchronization.}
\label{Figure:Relative:Translation_3dfront}
\end{figure*}

\begin{figure}[H]
\centering
    \begin{tabular}{ccc}
        \includegraphics[width=0.15\textwidth, trim=80px 80px 80px 0px, clip]{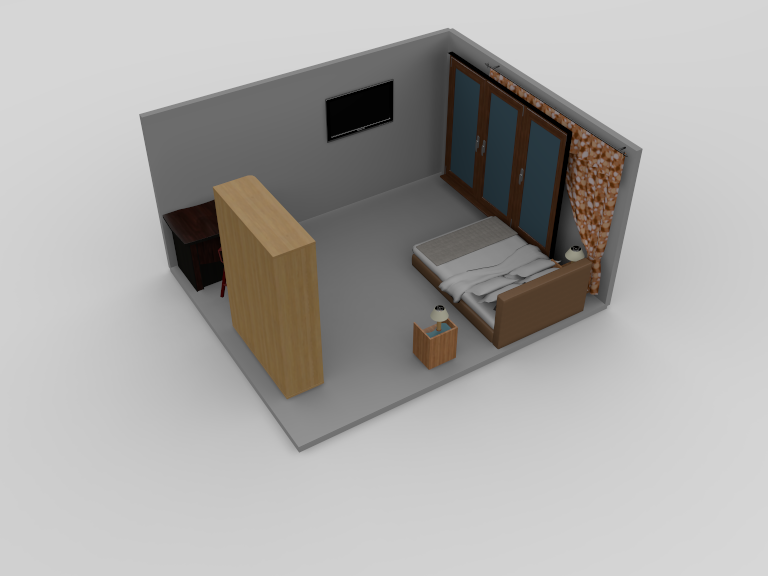} &
        \includegraphics[width=0.15\textwidth, trim=80px 80px 80px 0px, clip]{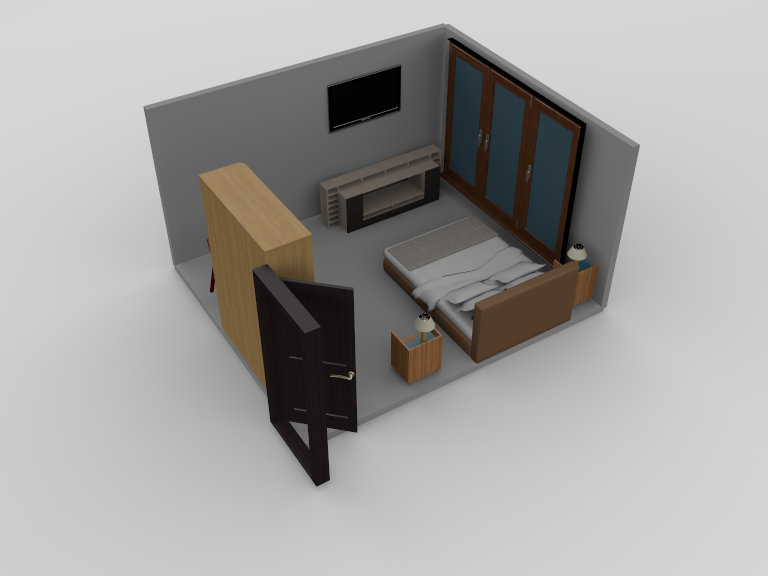} &
        \includegraphics[width=0.15\textwidth, trim=80px 80px 80px 0px, clip]{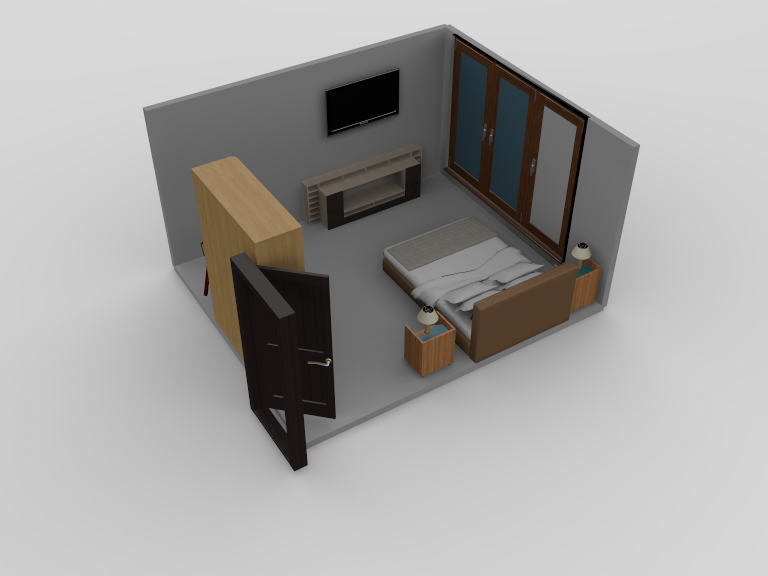}
    \end{tabular}
\caption{Left: output of the prediction module. Middle: scene optimization without using relative attribute predictions. Right: full pipeline. The relation between the bed and the nightstand, the TV and the TV stand are more realistic when utilizing the relative attributes in scene optimization.}
\label{Figure:Supp:Relative}
\end{figure}

\section{More Experimental Results and Analysis}
\label{sec:supp:results}

\subsection{Visual Comparisons between Our Method and Baseline Approaches}
\begin{figure*}
\centering
    \begin{tabular}{cccccc}
        \multirow{2}{4em}{SUNCG Bedroom} &
        \includegraphics[width=0.15\textwidth, trim=80px 80px 80px 0px, clip]{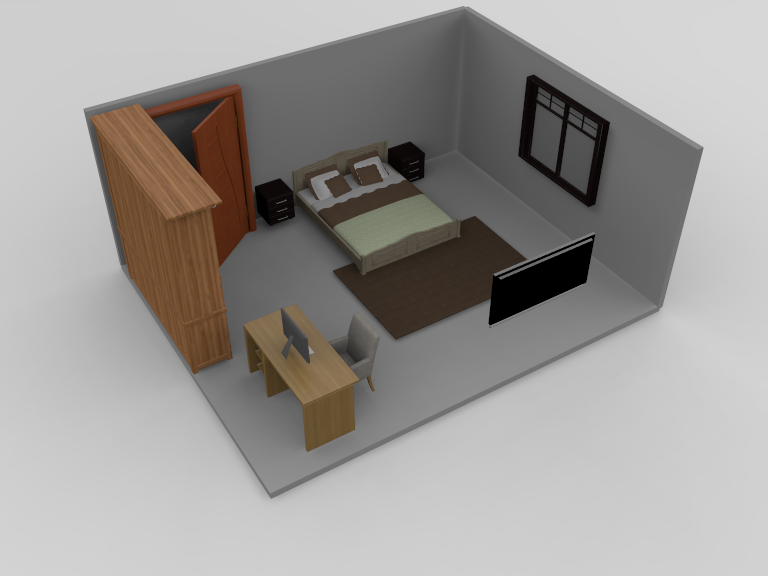} &
        \includegraphics[width=0.15\textwidth, trim=80px 80px 80px 0px, clip]{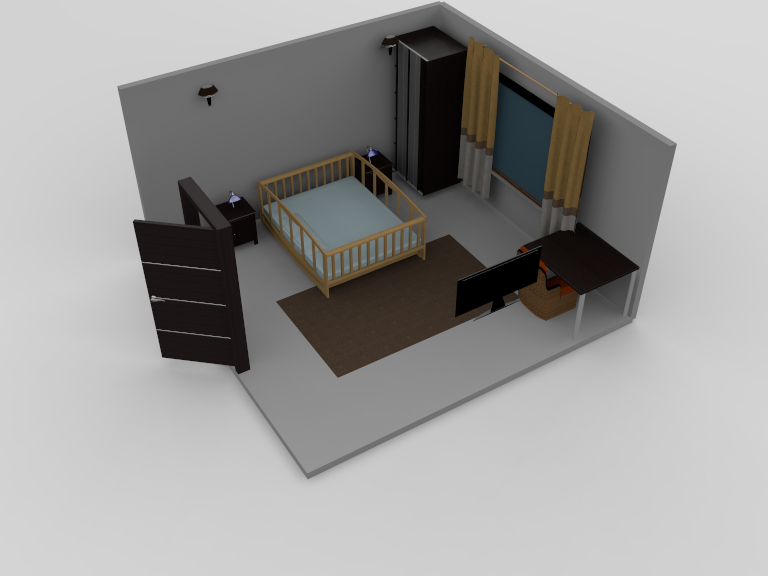} &
        \includegraphics[width=0.15\textwidth, trim=80px 80px 80px 0px, clip]{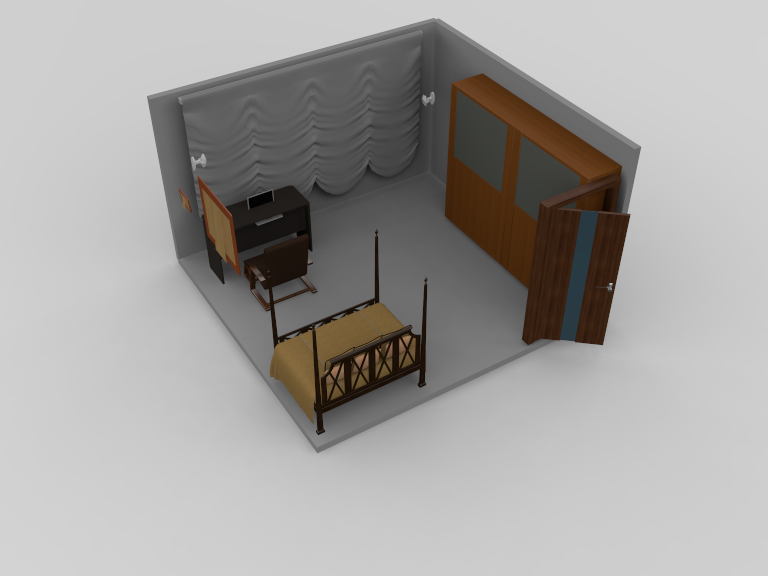} &
        \includegraphics[width=0.15\textwidth, trim=80px 80px 80px 0px, clip]{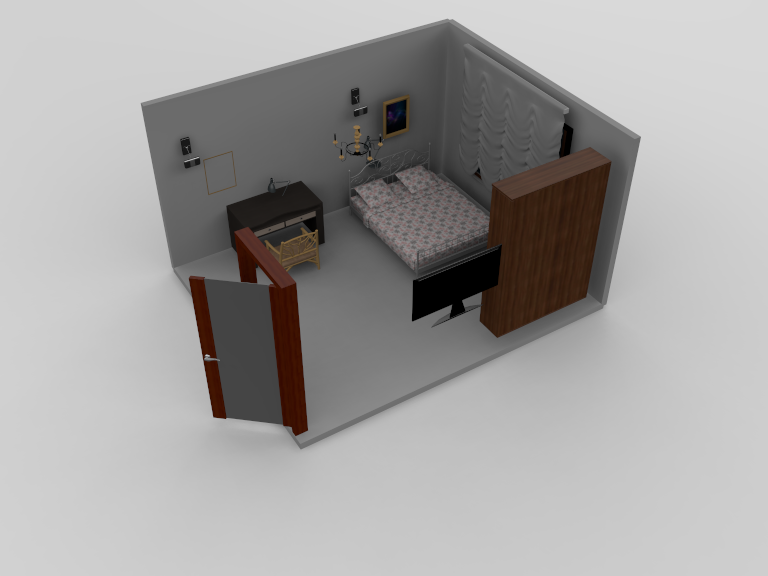} &
        \includegraphics[width=0.15\textwidth, trim=80px 80px 80px 0px, clip]{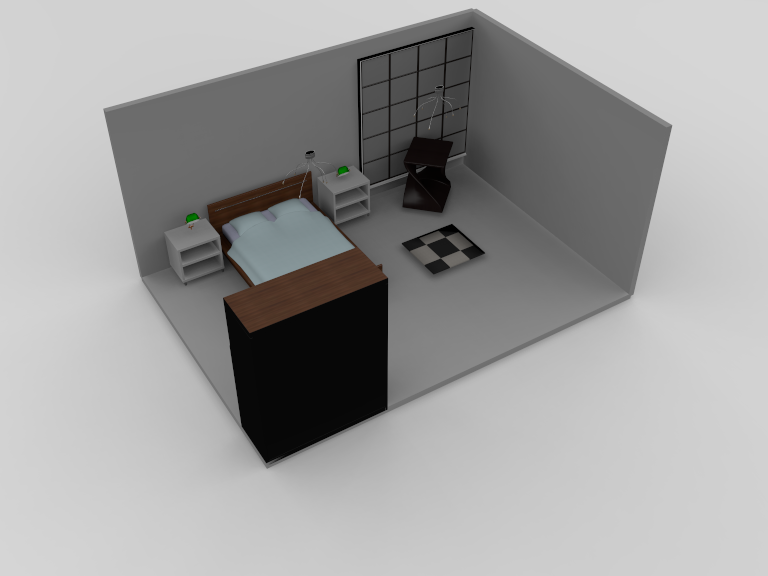} 
        \\ 
        & \includegraphics[width=0.15\textwidth, trim=80px 80px 80px 0px, clip]{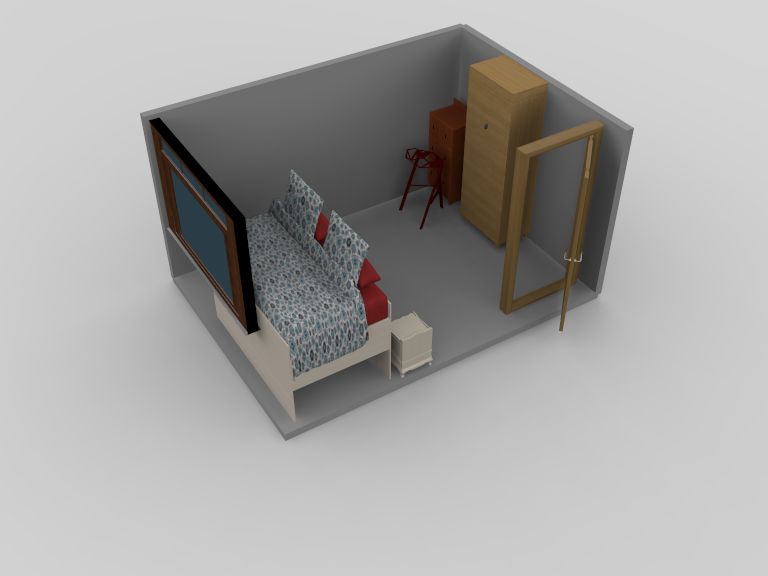} &
        \includegraphics[width=0.15\textwidth, trim=80px 80px 80px 0px, clip]{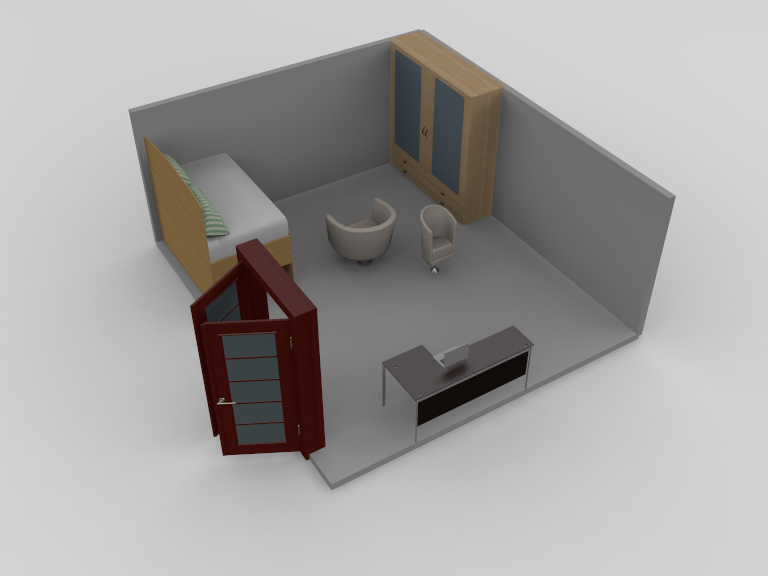} &
        \includegraphics[width=0.15\textwidth, trim=80px 80px 80px 0px, clip]{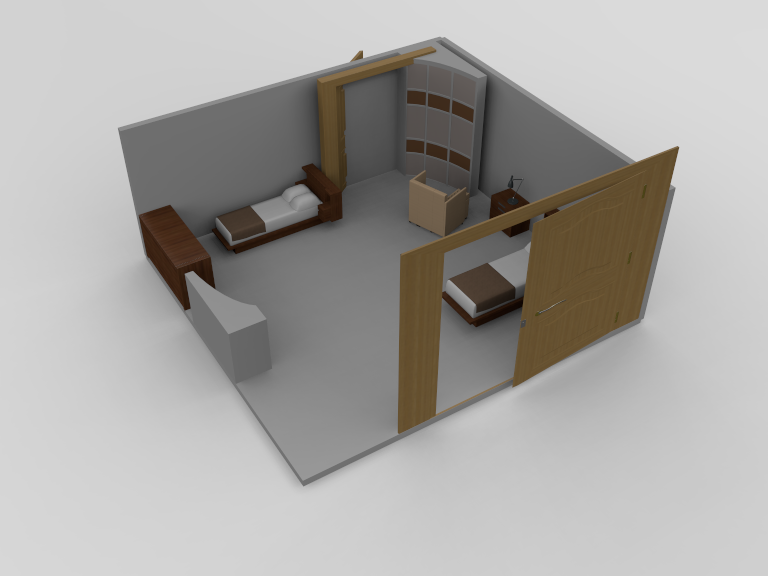} &
        \includegraphics[width=0.15\textwidth, trim=80px 80px 80px 0px, clip]{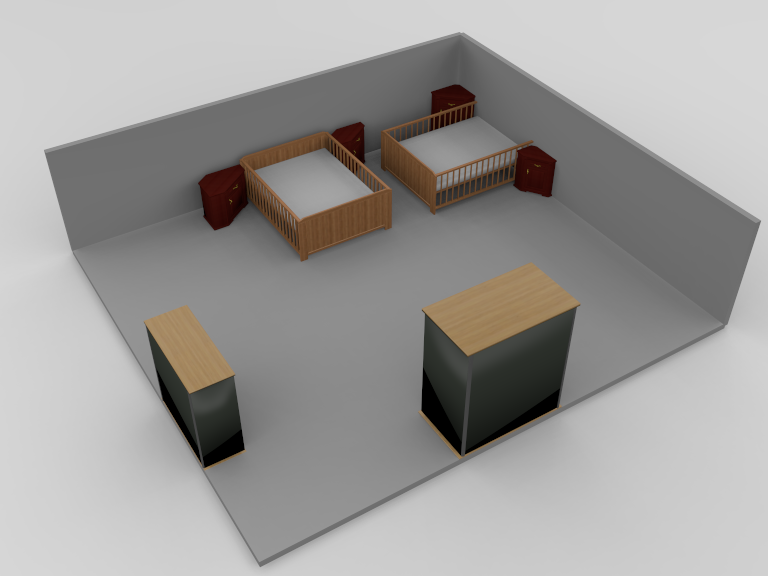} &
        \includegraphics[width=0.15\textwidth, trim=80px 80px 80px 0px, clip]{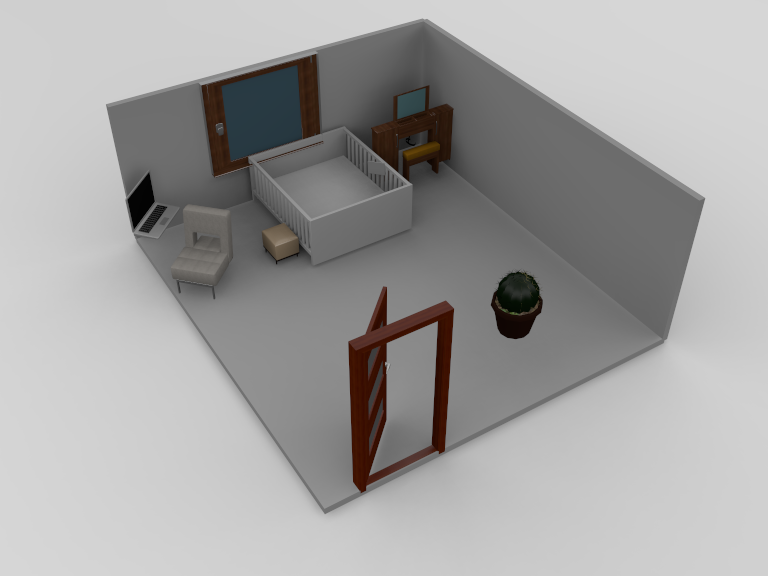}
        \\ \hline
        \multirow{2}{4em}{SUNCG Living} &
        \includegraphics[width=0.15\textwidth, trim=80px 80px 80px 0px, clip]{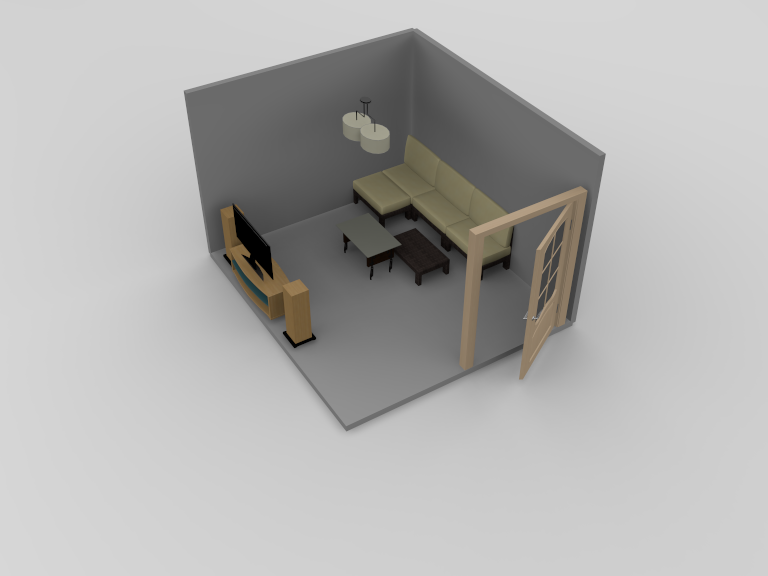} &
        \includegraphics[width=0.15\textwidth, trim=80px 80px 80px 0px, clip]{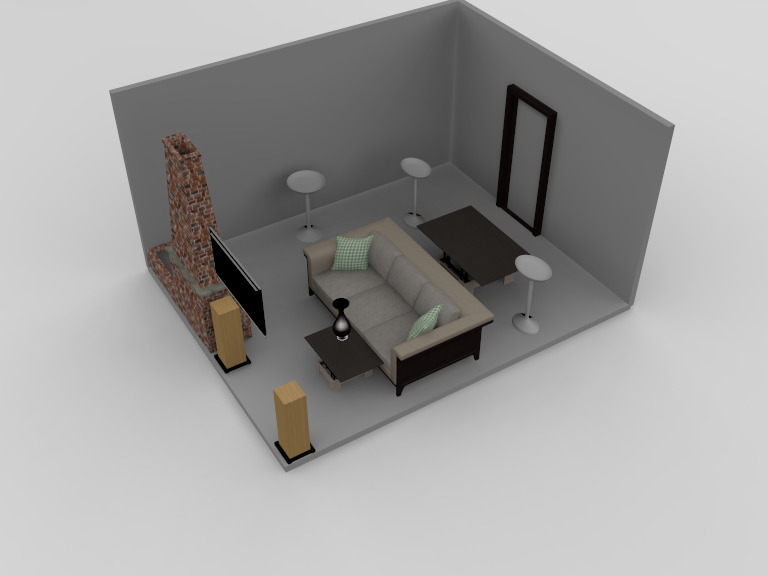} &
        \includegraphics[width=0.15\textwidth, trim=80px 80px 80px 0px, clip]{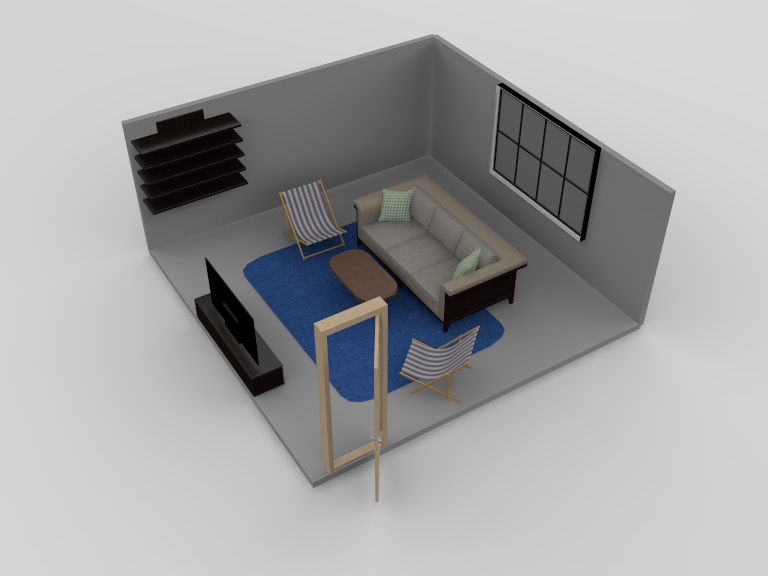} &
        \includegraphics[width=0.15\textwidth, trim=80px 80px 80px 0px, clip]{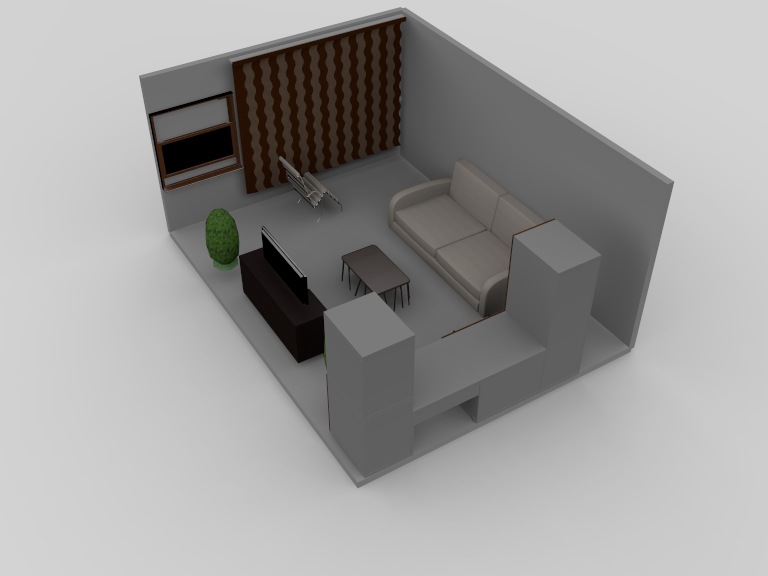} &
        \includegraphics[width=0.15\textwidth, trim=80px 80px 80px 0px, clip]{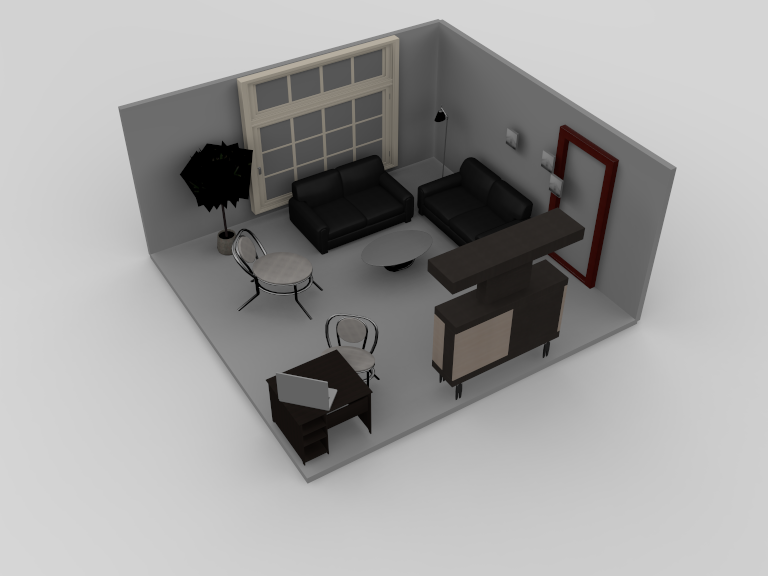}
        \\ 
        & \includegraphics[width=0.15\textwidth, trim=80px 80px 80px 0px, clip]{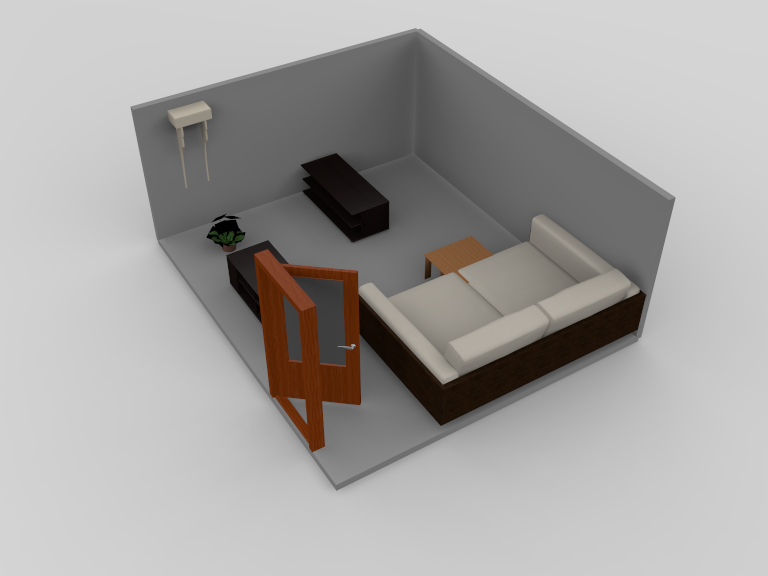}&
        \includegraphics[width=0.15\textwidth, trim=80px 80px 80px 0px, clip]{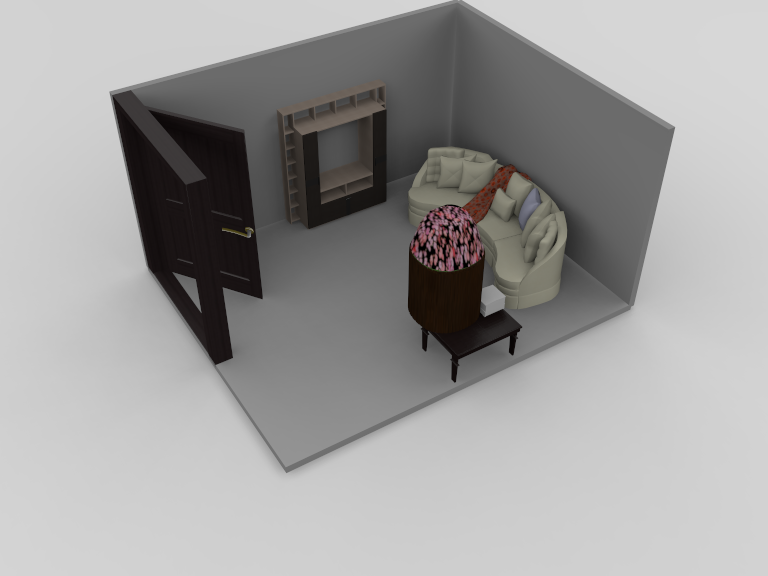} &
        \includegraphics[width=0.15\textwidth, trim=80px 80px 80px 0px, clip]{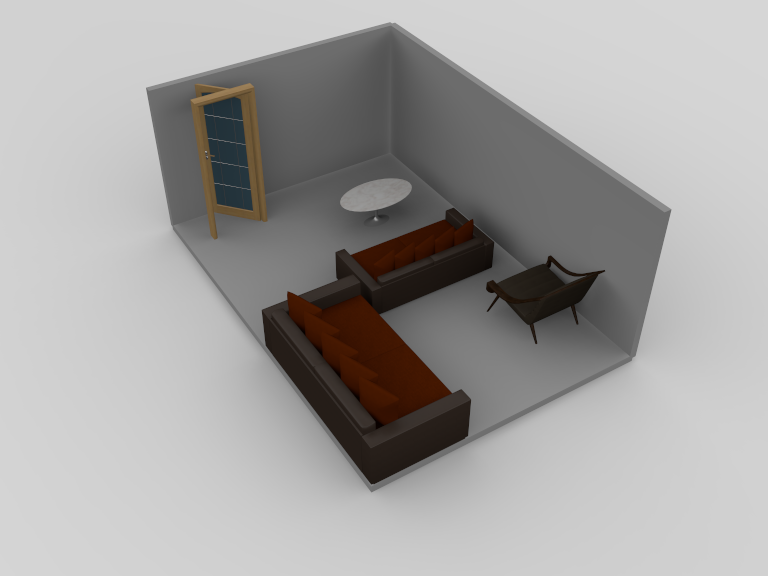} &
        \includegraphics[width=0.15\textwidth, trim=80px 80px 80px 0px, clip]{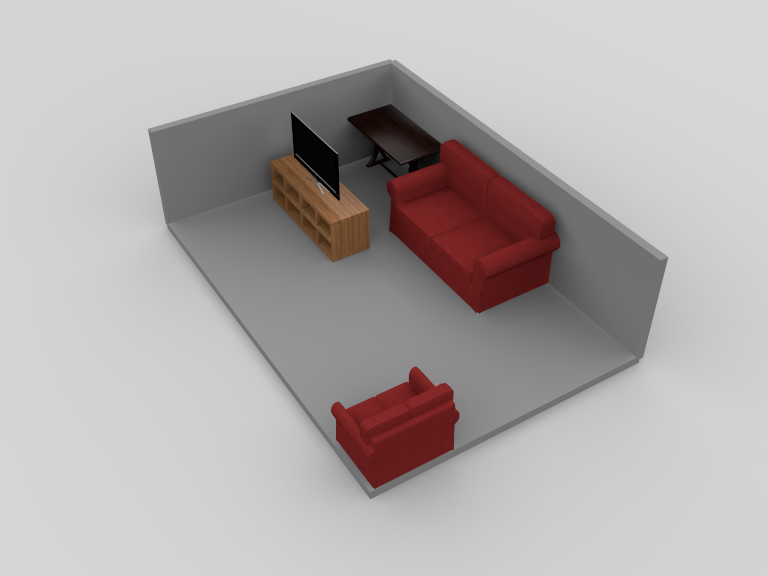} &
        \includegraphics[width=0.15\textwidth, trim=80px 80px 80px 0px, clip]{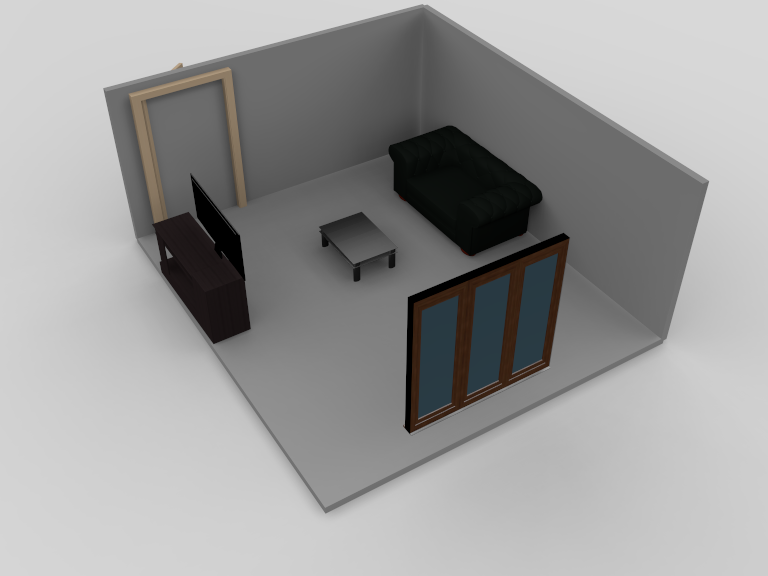}
        \\ \hline
        \multirow{2}{4em}{3D-FRONT Bedroom} &
        \includegraphics[width=0.15\textwidth, trim=80px 80px 80px 0px, clip]{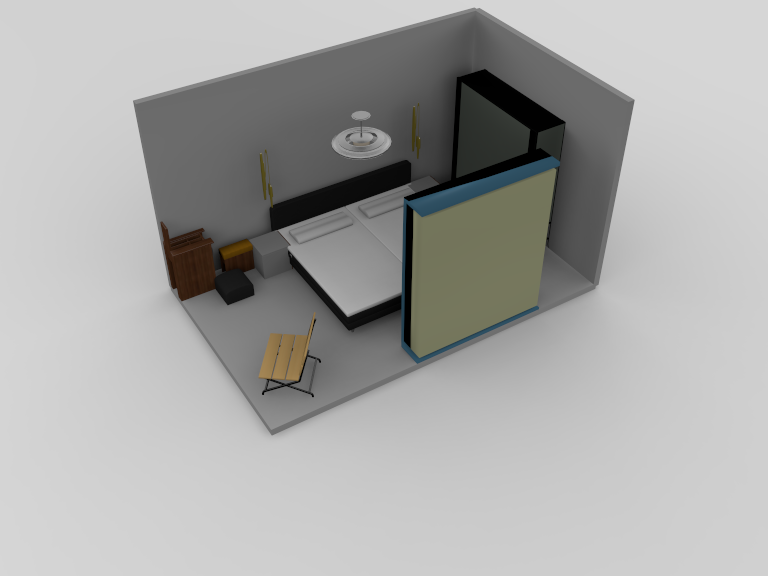} &
        \includegraphics[width=0.15\textwidth, trim=80px 80px 80px 0px, clip]{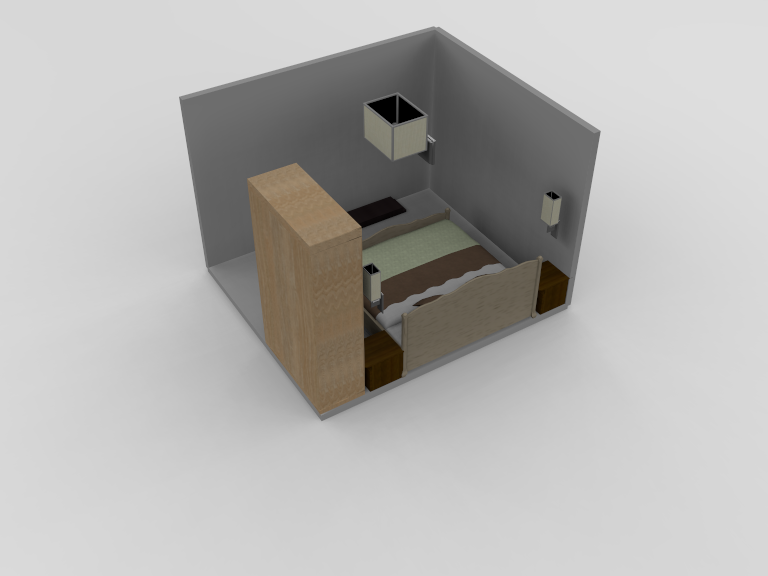} &
        \includegraphics[width=0.15\textwidth, trim=80px 80px 80px 0px, clip]{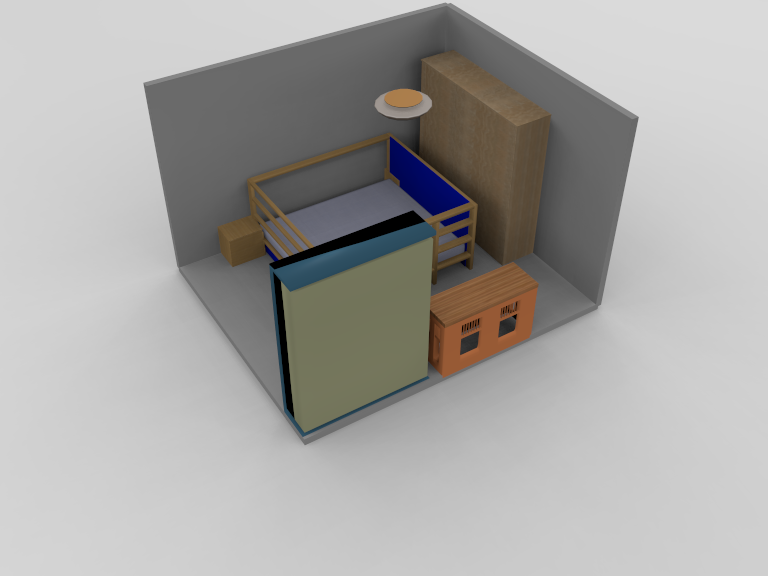} &
        \includegraphics[width=0.15\textwidth, trim=80px 80px 80px 0px, clip]{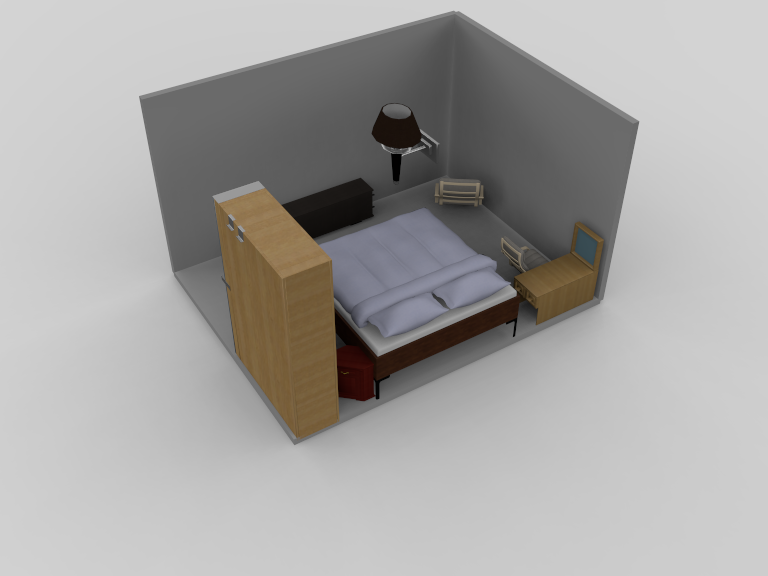} &
        \includegraphics[width=0.15\textwidth, trim=80px 80px 80px 0px, clip]{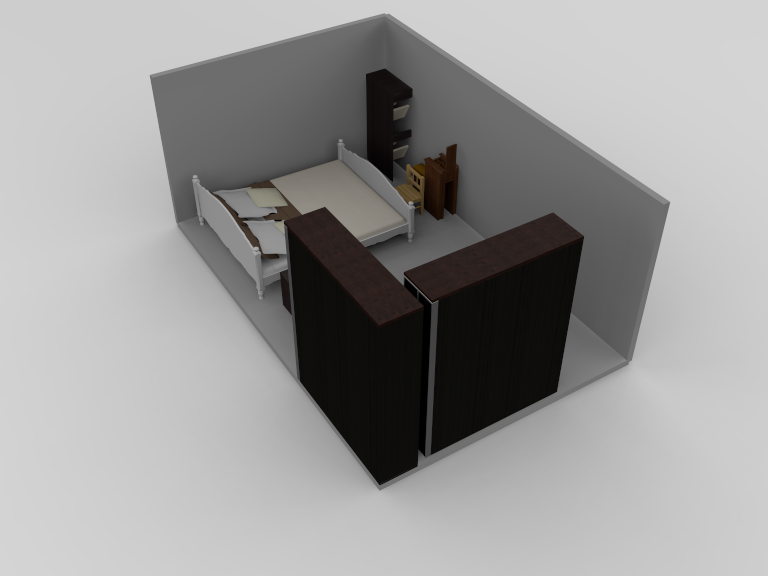}
        \\ 
        & \includegraphics[width=0.15\textwidth, trim=80px 80px 80px 0px, clip]{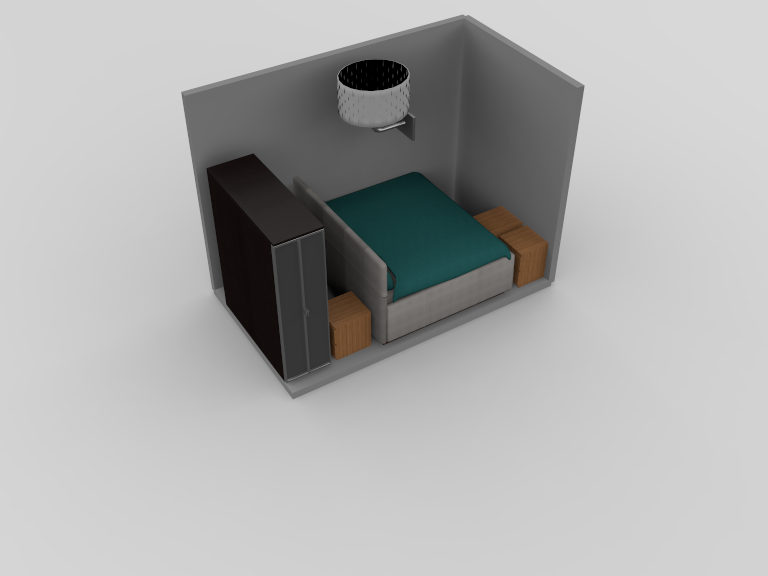} &
        \includegraphics[width=0.15\textwidth, trim=80px 80px 80px 0px, clip]{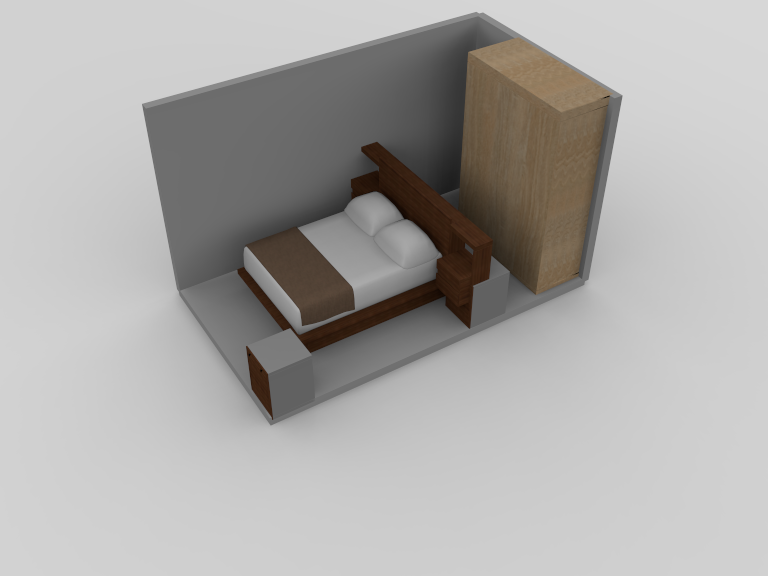} &
        \includegraphics[width=0.15\textwidth, trim=80px 80px 80px 0px, clip]{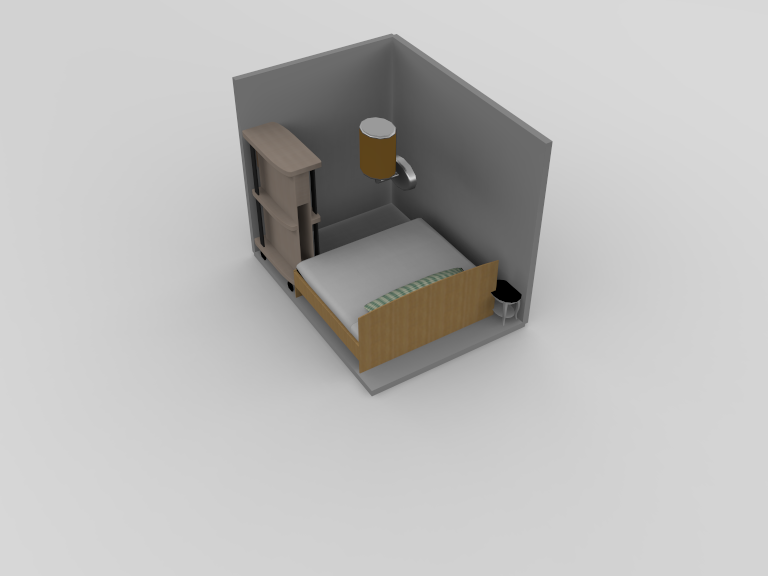}&
        \includegraphics[width=0.15\textwidth, trim=80px 80px 80px 0px, clip]{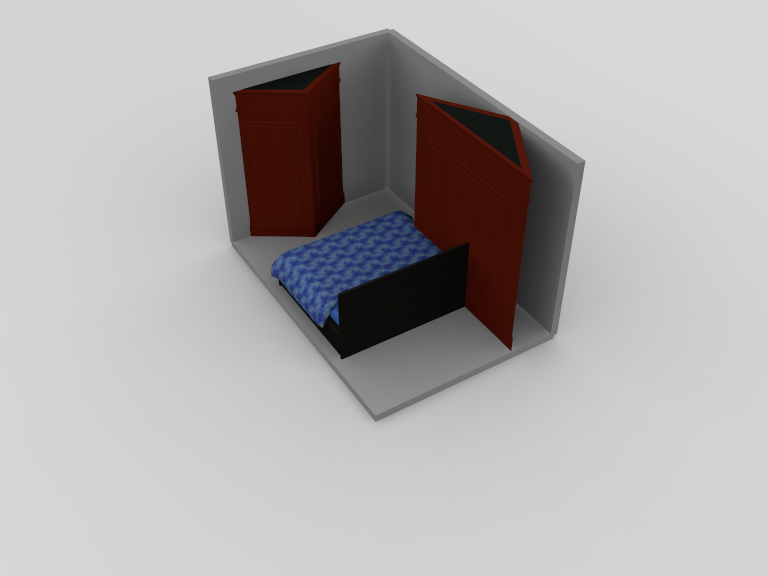} &
        \includegraphics[width=0.15\textwidth, trim=80px 80px 80px 0px, clip]{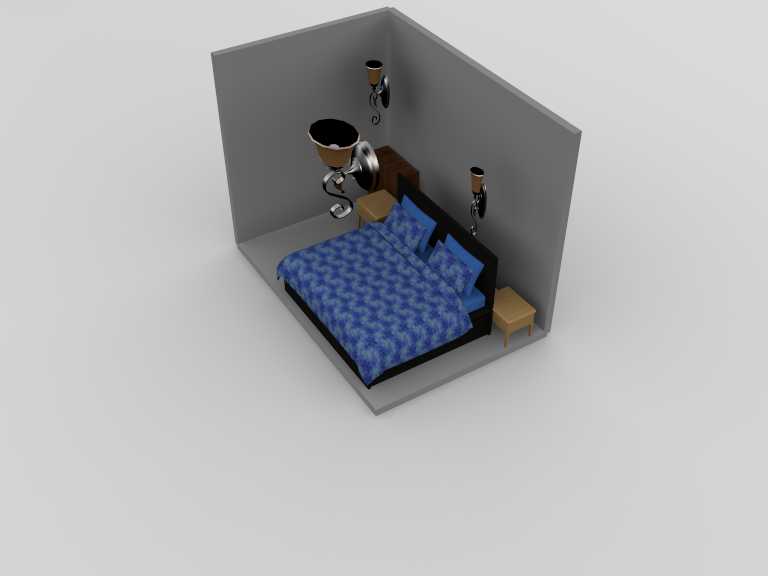}
        \\ \hline
        \multirow{2}{4em}{3D-FRONT Living} &
        \includegraphics[width=0.15\textwidth, trim=80px 80px 80px 0px, clip]{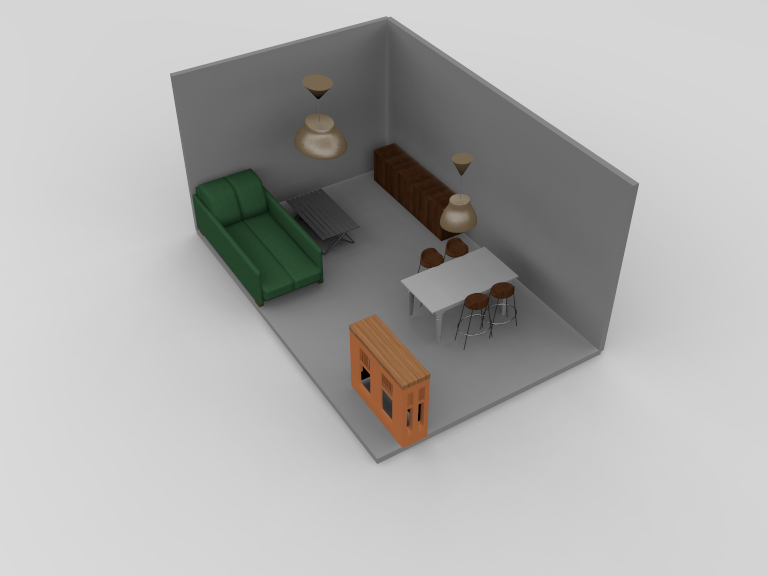} &
        \includegraphics[width=0.15\textwidth, trim=80px 80px 80px 0px, clip]{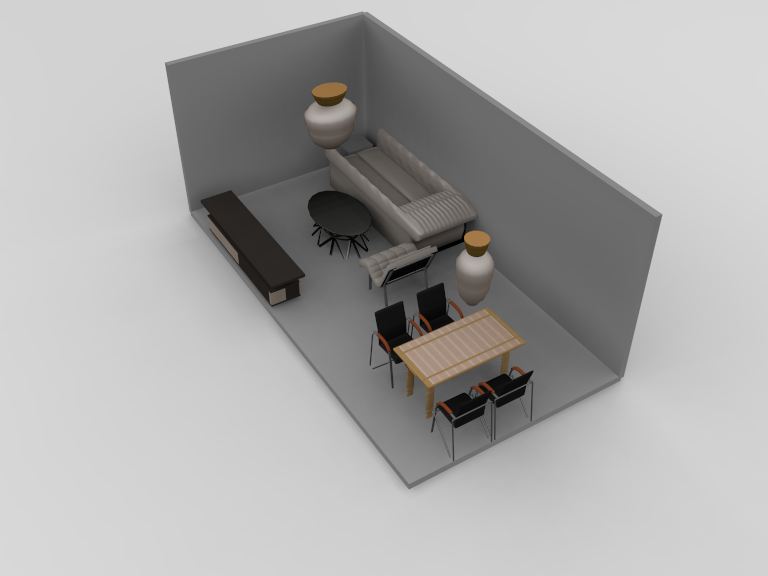} &
        \includegraphics[width=0.15\textwidth, trim=80px 80px 80px 0px, clip]{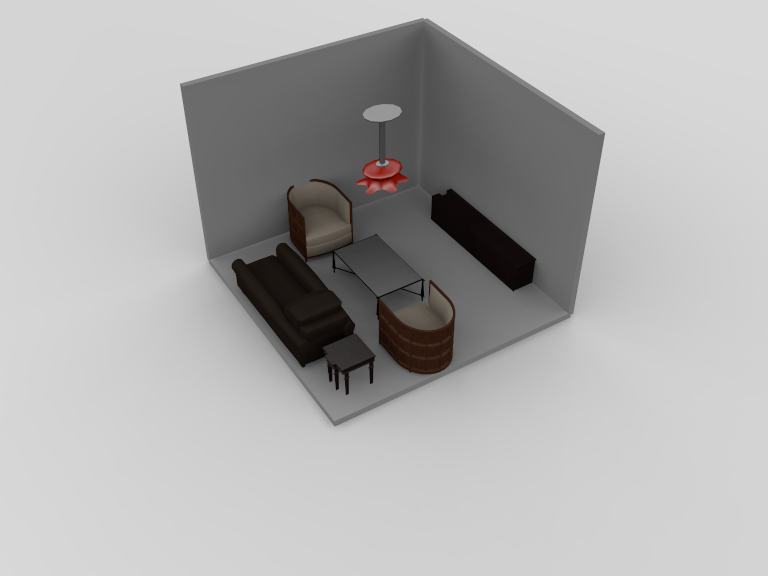} &
        \includegraphics[width=0.15\textwidth, trim=80px 80px 80px 0px, clip]{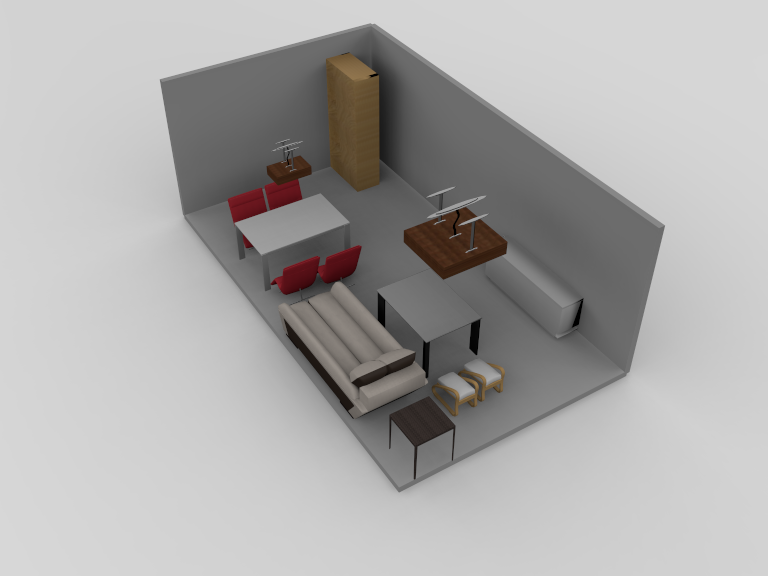} &
        \includegraphics[width=0.15\textwidth, trim=80px 80px 80px 0px, clip]{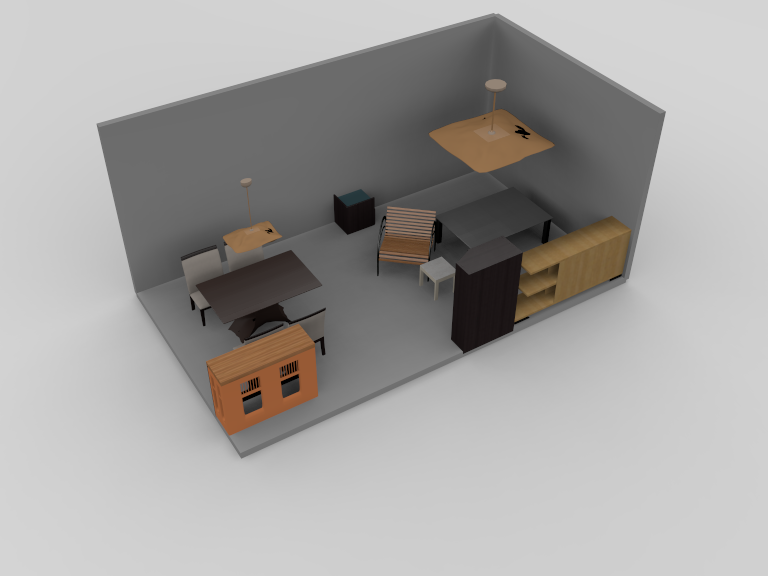}
        \\ 
        & \includegraphics[width=0.15\textwidth, trim=80px 80px 80px 0px, clip]{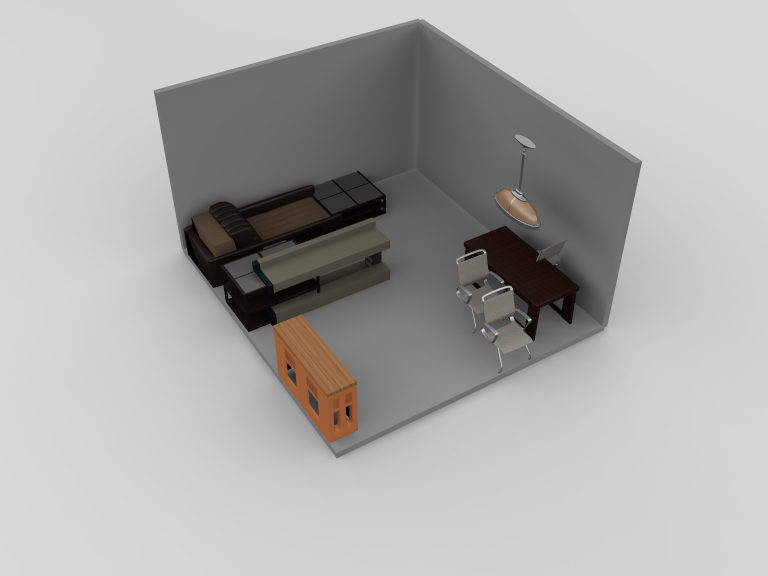} &
        \includegraphics[width=0.15\textwidth, trim=80px 80px 80px 0px, clip]{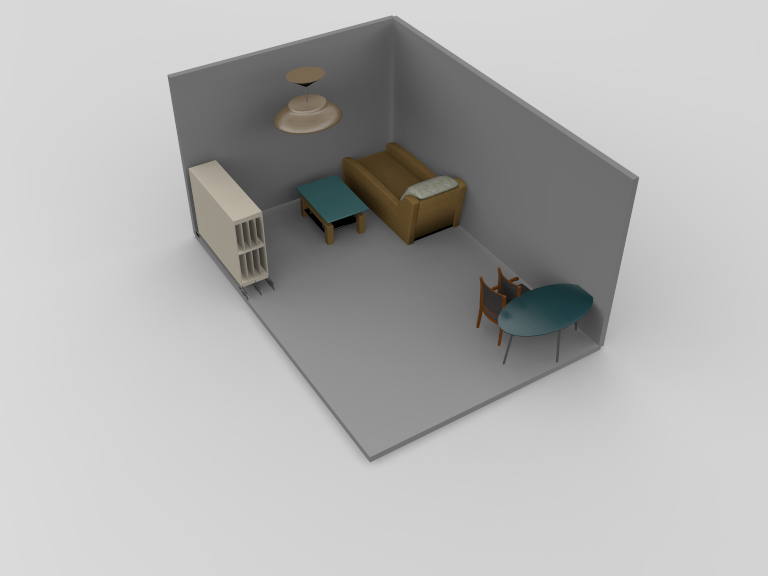} &
        \includegraphics[width=0.15\textwidth, trim=80px 80px 80px 0px, clip]{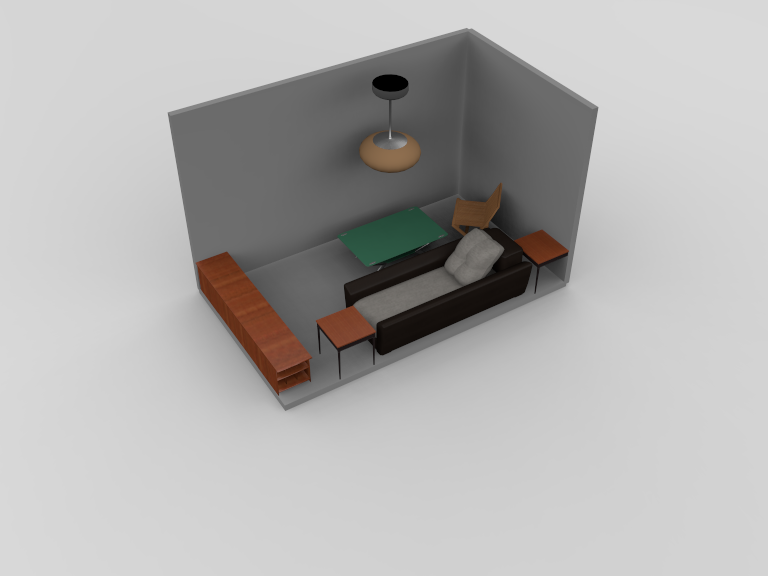} &
        \includegraphics[width=0.15\textwidth, trim=80px 80px 80px 0px, clip]{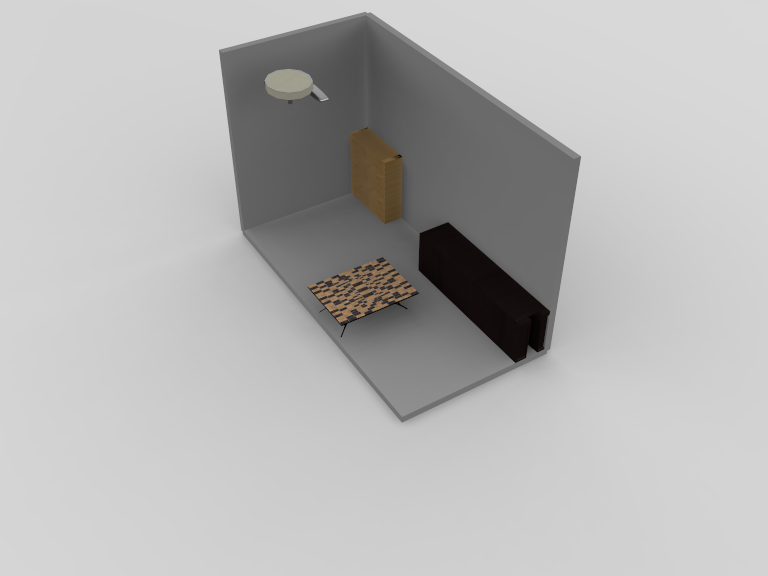} &
        \includegraphics[width=0.15\textwidth, trim=80px 80px 80px 0px, clip]{Figures/front_living/0004__init.png}
        \\\hline 
        & D-Prior & Fast & PlanIT & GRAINS & D-Gen
    \end{tabular}
\caption{Scene synthesis results between ours and baseline methods. For each dataset, the first row: our results, the second row: baseline methods. For baseline methods, from left to right: D-Prior, Fast, PlanIT, GRAINS, D-Gen.}
\label{Figure:Supp:Results}
\end{figure*}

\begin{figure*}
\centering
    \begin{tabular}{cccccc}
        \multirow{2}{4em}{SUNCG Bedroom} &
        \includegraphics[width=0.15\textwidth, trim=80px 80px 80px 0px, clip]{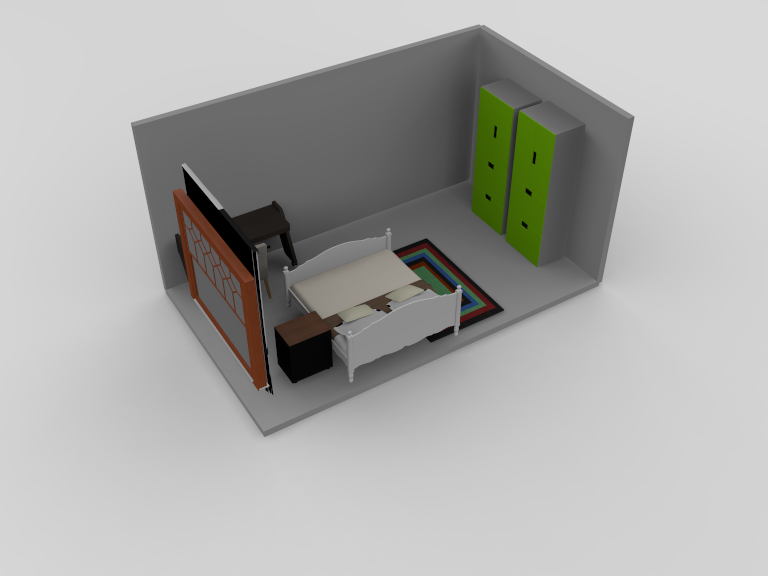} &
        \includegraphics[width=0.15\textwidth, trim=80px 80px 80px 0px, clip]{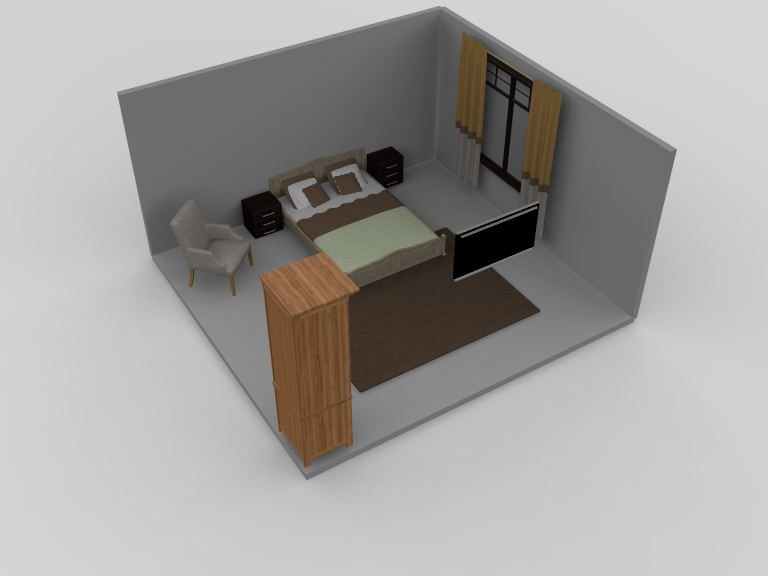} &
        \includegraphics[width=0.15\textwidth, trim=80px 80px 80px 0px, clip]{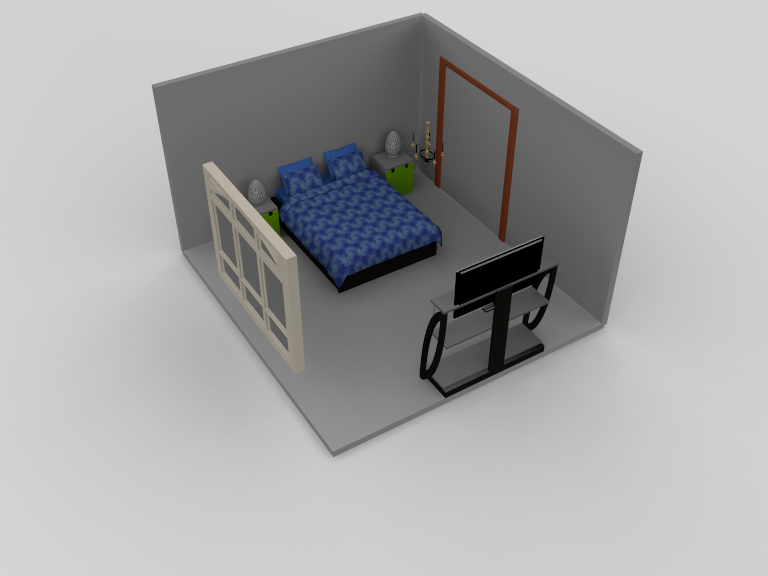}  &
        \includegraphics[width=0.15\textwidth, trim=80px 80px 80px 0px, clip]{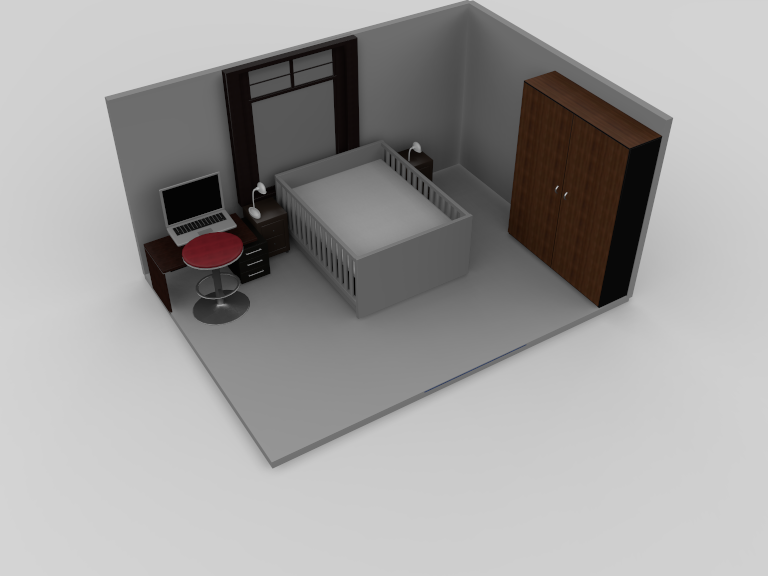}  &
        \includegraphics[width=0.15\textwidth, trim=80px 80px 80px 0px, clip]{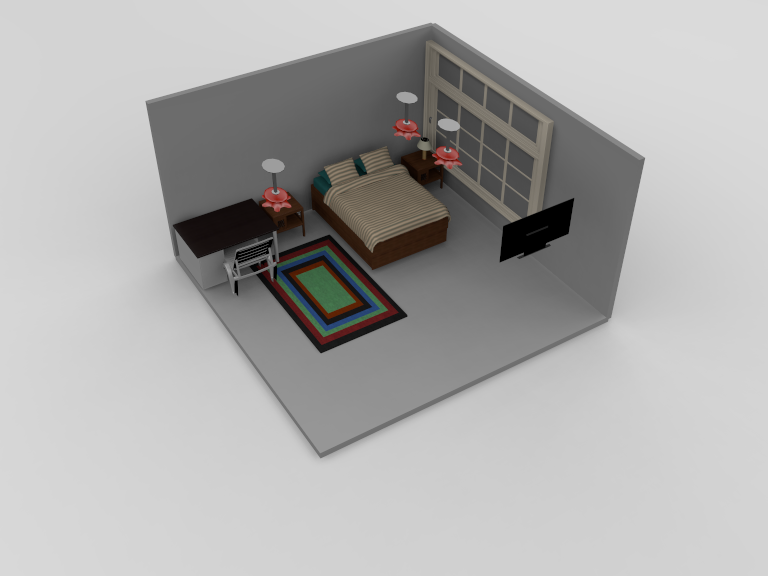}
        \\ 
        & \includegraphics[width=0.15\textwidth, trim=80px 80px 80px 0px, clip]{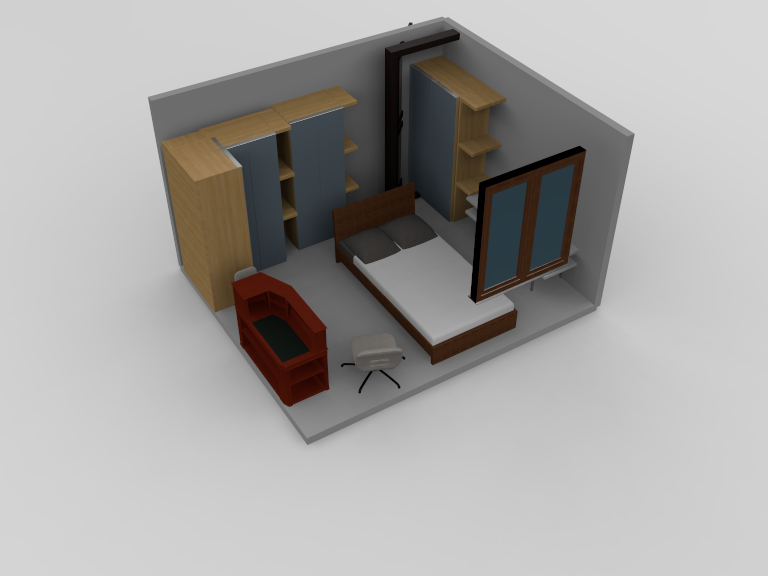} &
        \includegraphics[width=0.15\textwidth, trim=80px 80px 80px 0px, clip]{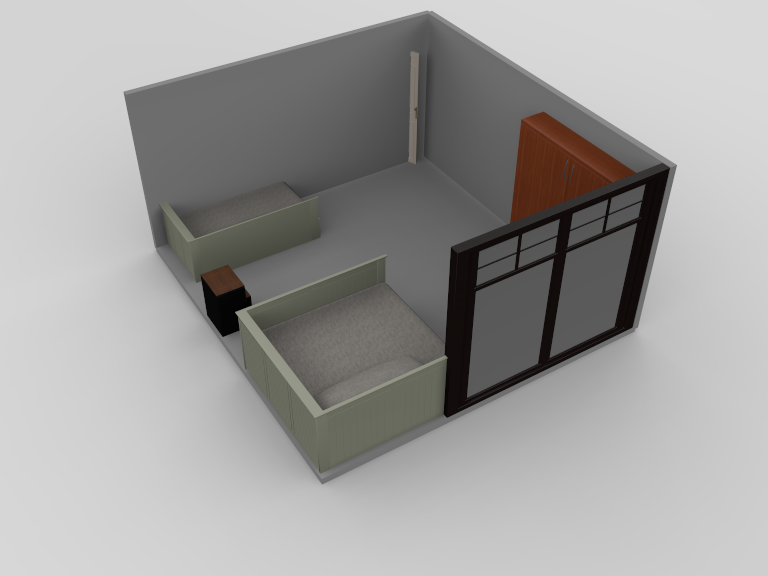} &
        \includegraphics[width=0.15\textwidth, trim=80px 80px 80px 0px, clip]{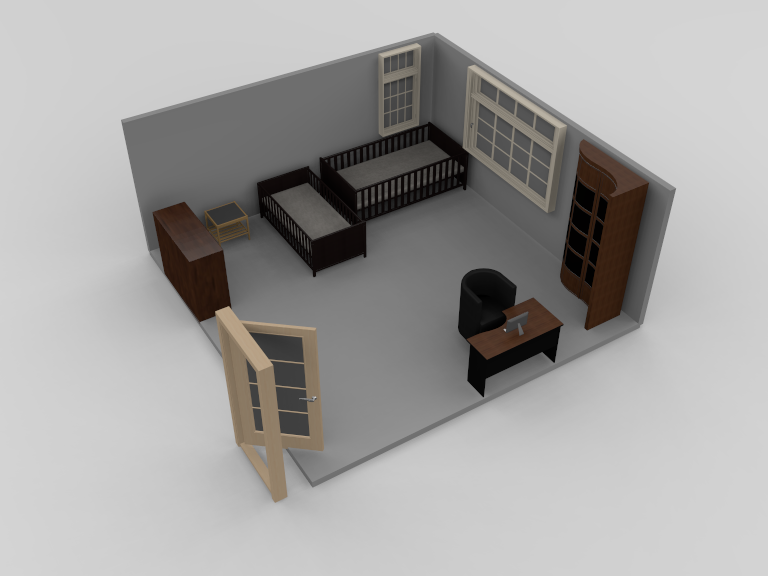} &
        \includegraphics[width=0.15\textwidth, trim=80px 80px 80px 0px, clip]{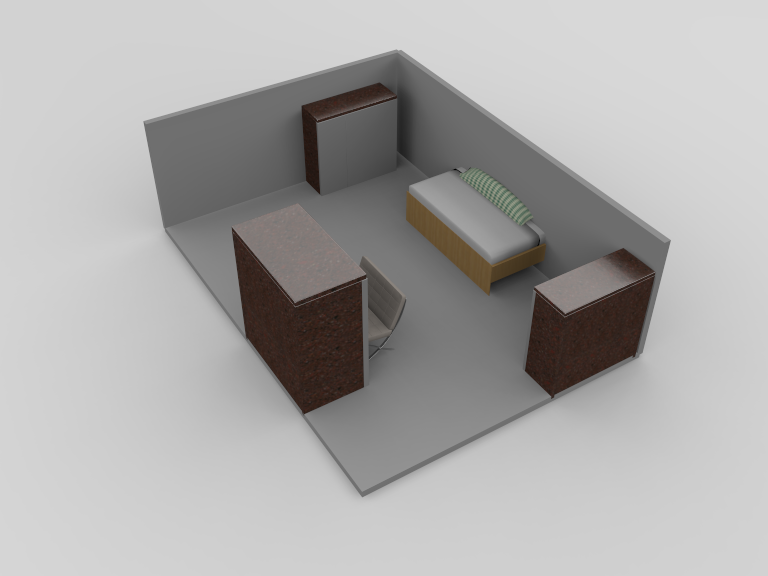} &
        \includegraphics[width=0.15\textwidth, trim=80px 80px 80px 0px, clip]{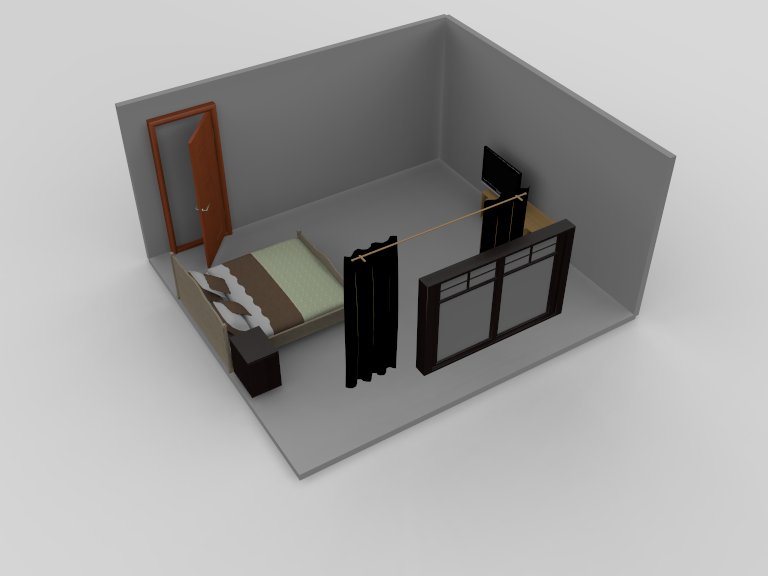}
        \\ \hline
        \multirow{2}{4em}{SUNCG Living} &
        \includegraphics[width=0.15\textwidth, trim=80px 80px 80px 0px, clip]{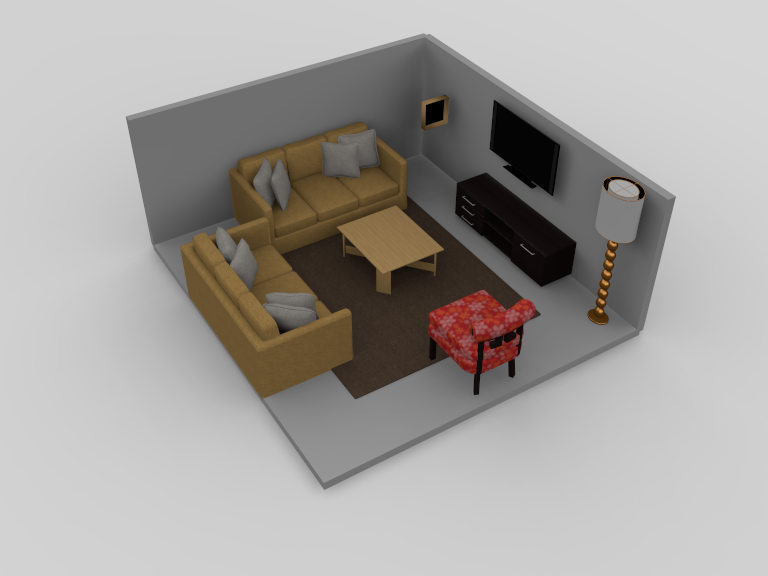} &
        \includegraphics[width=0.15\textwidth, trim=80px 80px 80px 0px, clip]{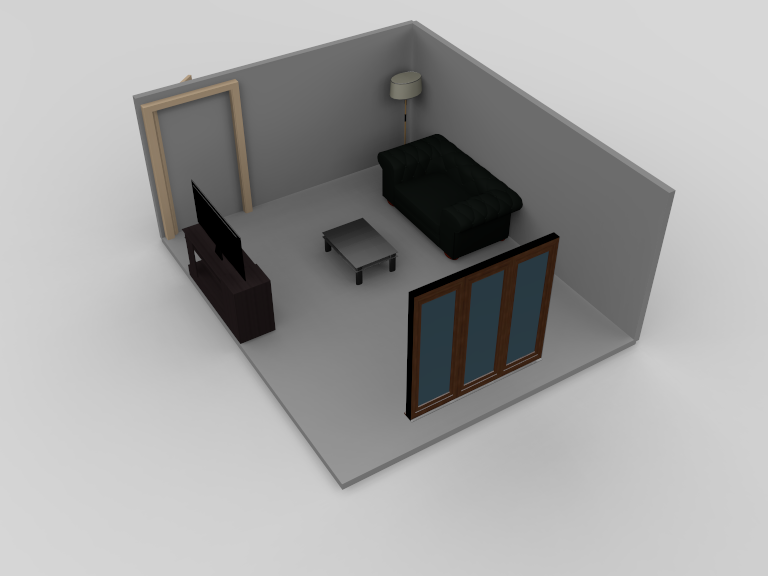} &
        \includegraphics[width=0.15\textwidth, trim=80px 80px 80px 0px, clip]{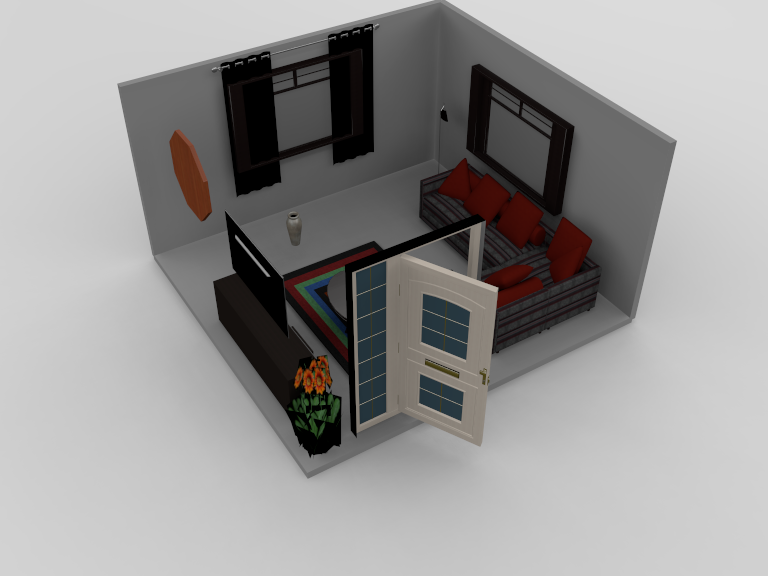} &
        \includegraphics[width=0.15\textwidth, trim=80px 80px 80px 0px, clip]{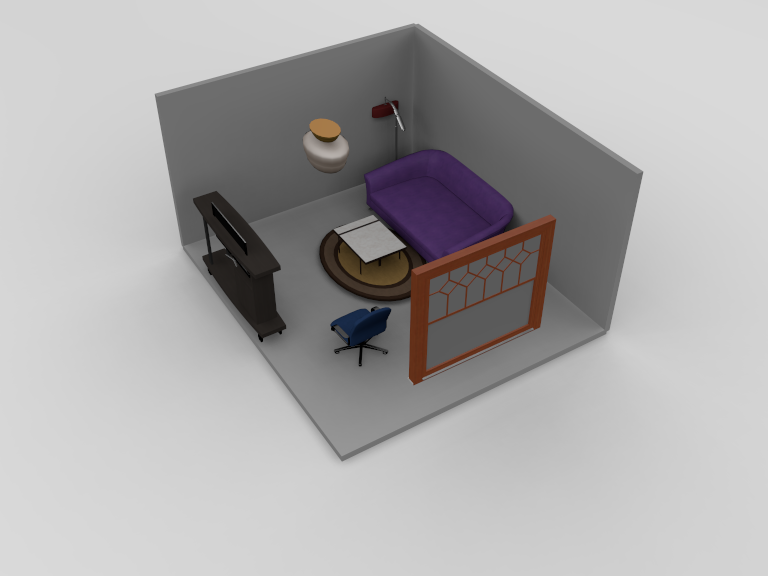} &
        \includegraphics[width=0.15\textwidth, trim=80px 80px 80px 0px, clip]{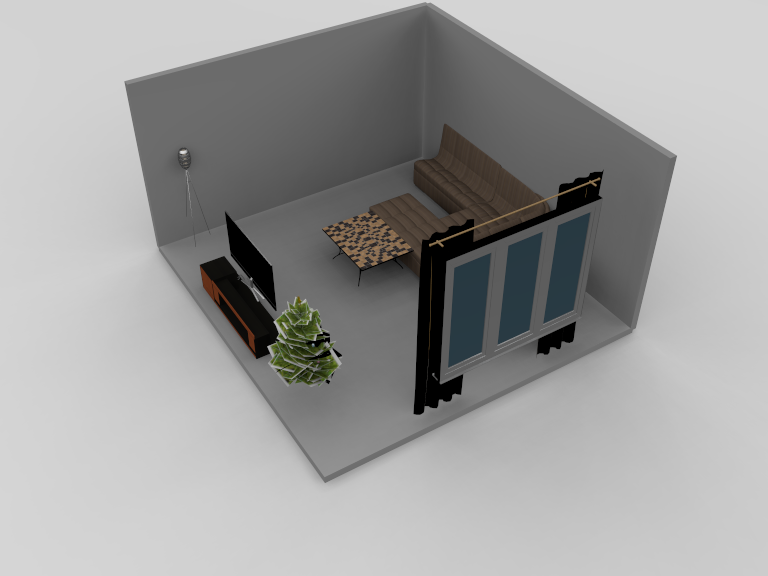}
        \\ 
        & \includegraphics[width=0.15\textwidth, trim=80px 80px 80px 0px, clip]{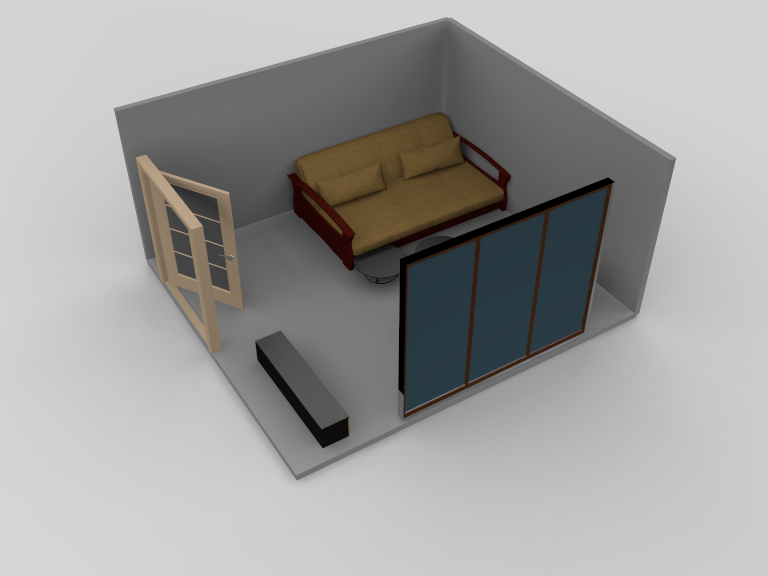}&
        \includegraphics[width=0.15\textwidth, trim=80px 80px 80px 0px, clip]{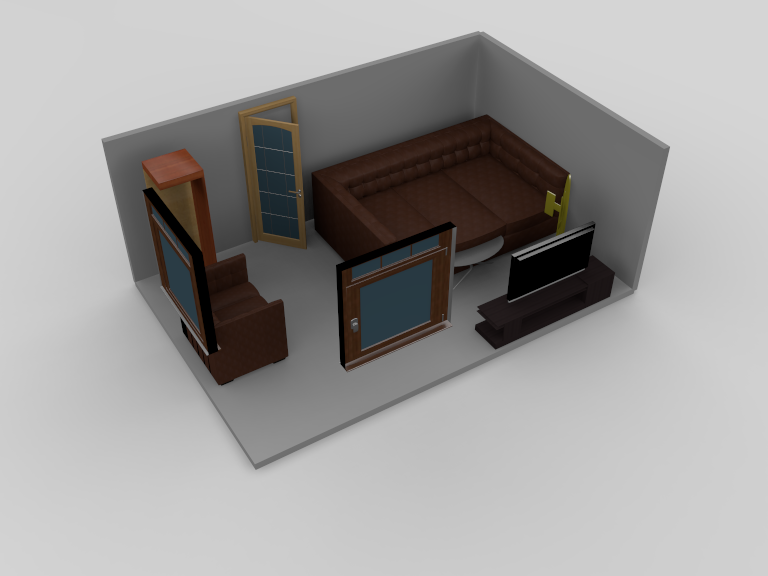} &
        \includegraphics[width=0.15\textwidth, trim=80px 80px 80px 0px, clip]{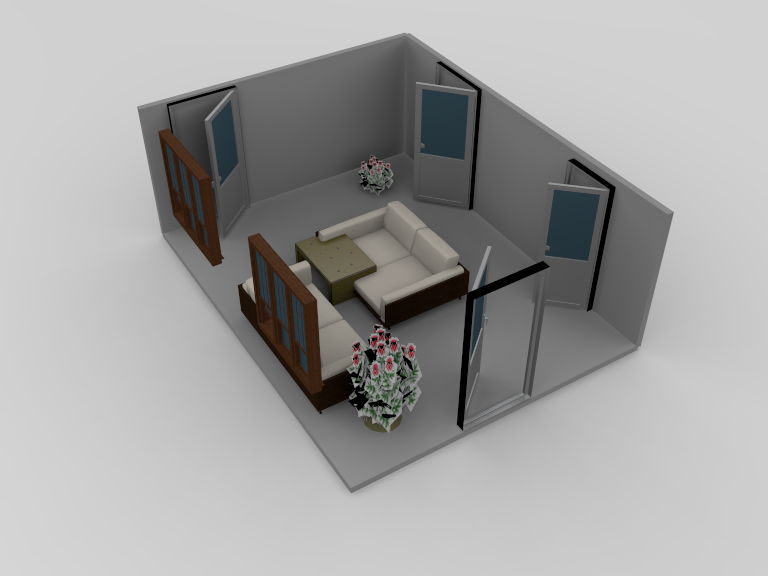} &
        \includegraphics[width=0.15\textwidth, trim=80px 80px 80px 0px, clip]{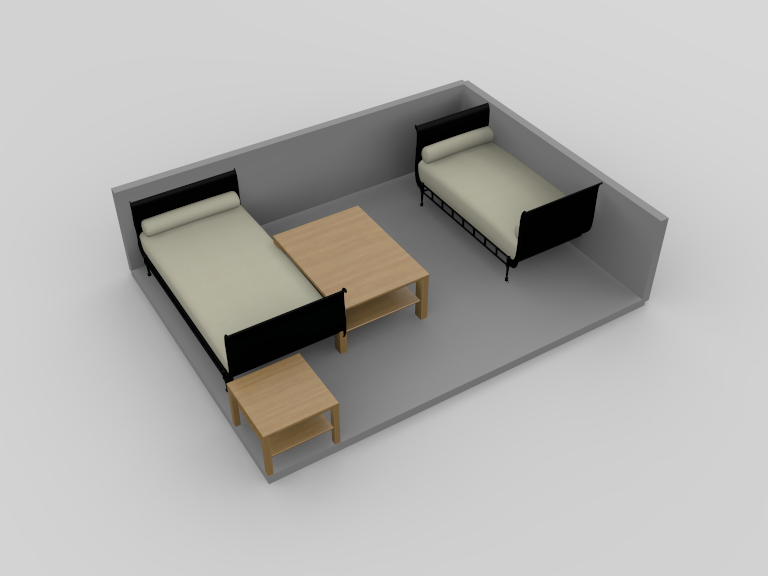} &
        \includegraphics[width=0.15\textwidth, trim=80px 80px 80px 0px, clip]{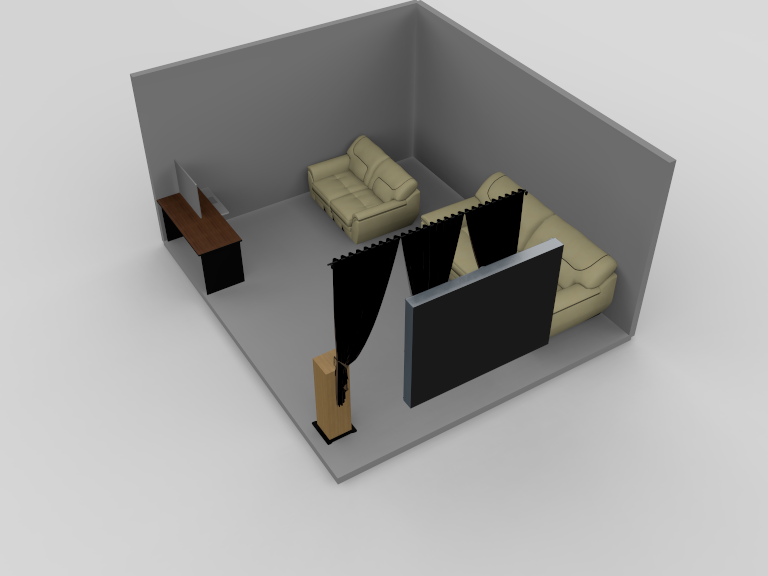}
        \\ \hline
        \multirow{2}{4em}{3D-FRONT Bedroom} &
        \includegraphics[width=0.15\textwidth, trim=80px 80px 80px 0px, clip]{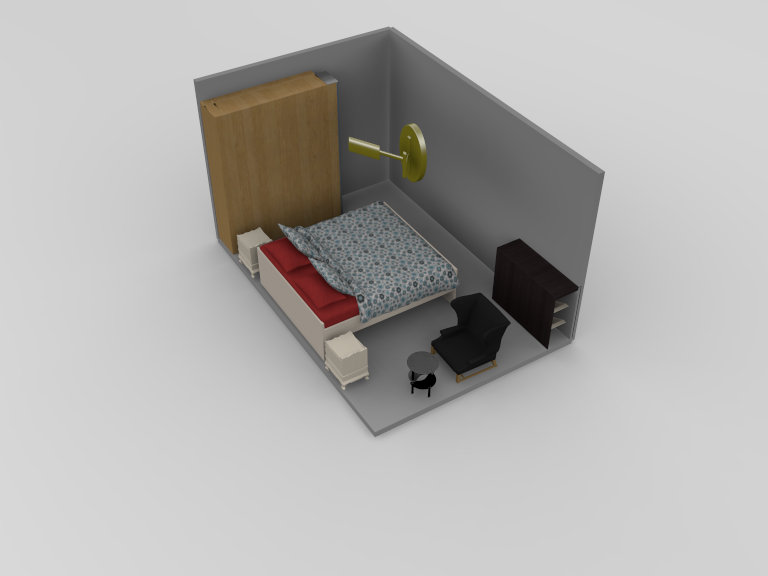} &
        \includegraphics[width=0.15\textwidth, trim=80px 80px 80px 0px, clip]{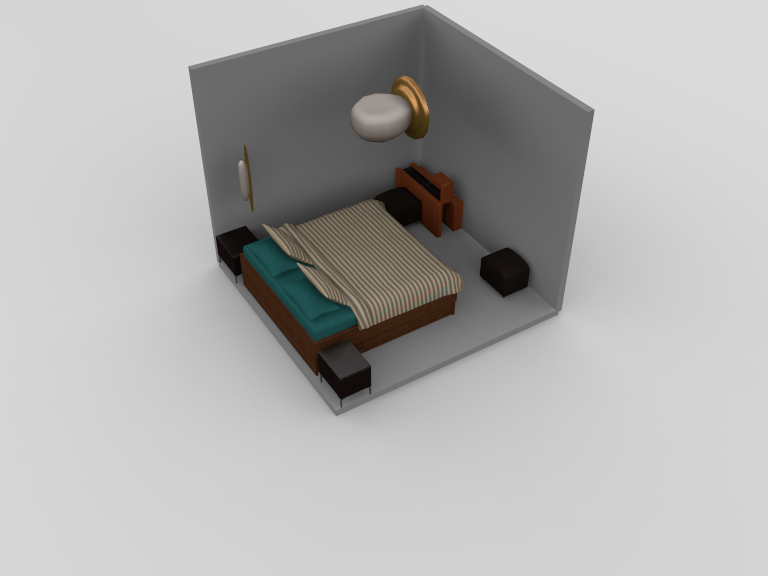} &
        \includegraphics[width=0.15\textwidth, trim=80px 80px 80px 0px, clip]{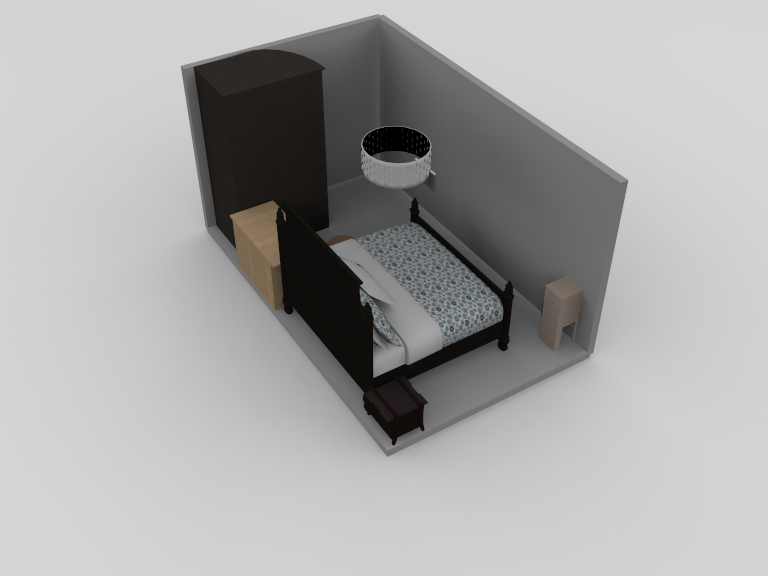} &
        \includegraphics[width=0.15\textwidth, trim=80px 80px 80px 0px, clip]{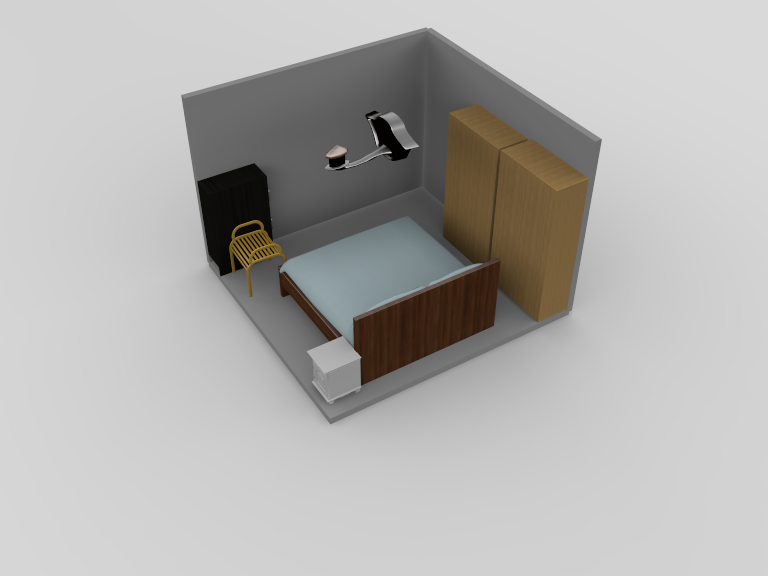} &
        \includegraphics[width=0.15\textwidth, trim=80px 80px 80px 0px, clip]{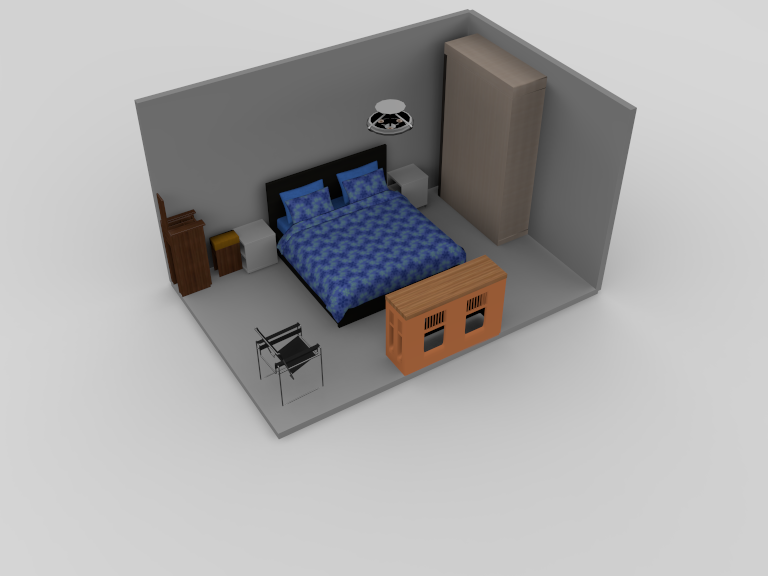}
        \\ 
        & \includegraphics[width=0.15\textwidth, trim=80px 80px 80px 0px, clip]{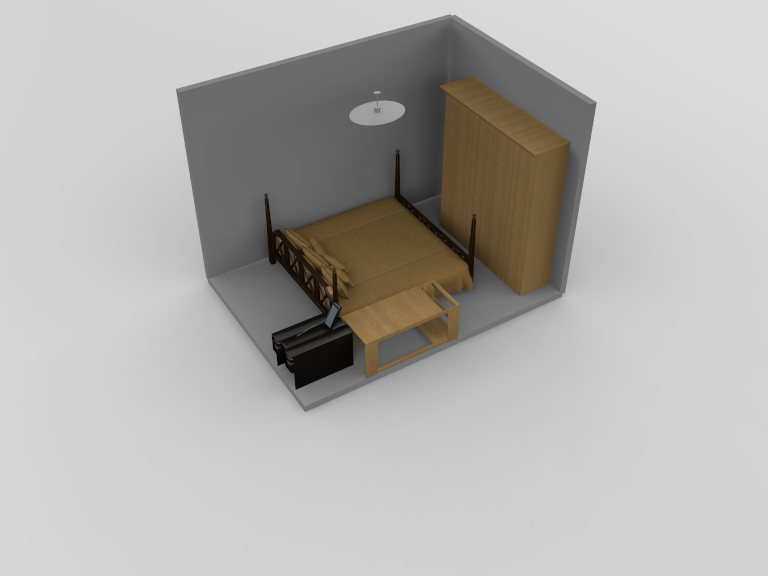} &
        \includegraphics[width=0.15\textwidth, trim=80px 80px 80px 0px, clip]{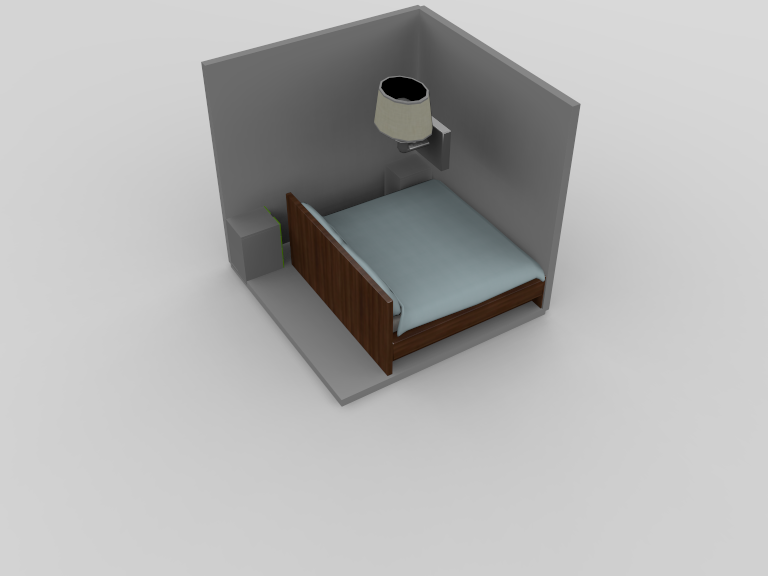} &
        \includegraphics[width=0.15\textwidth, trim=80px 80px 80px 0px, clip]{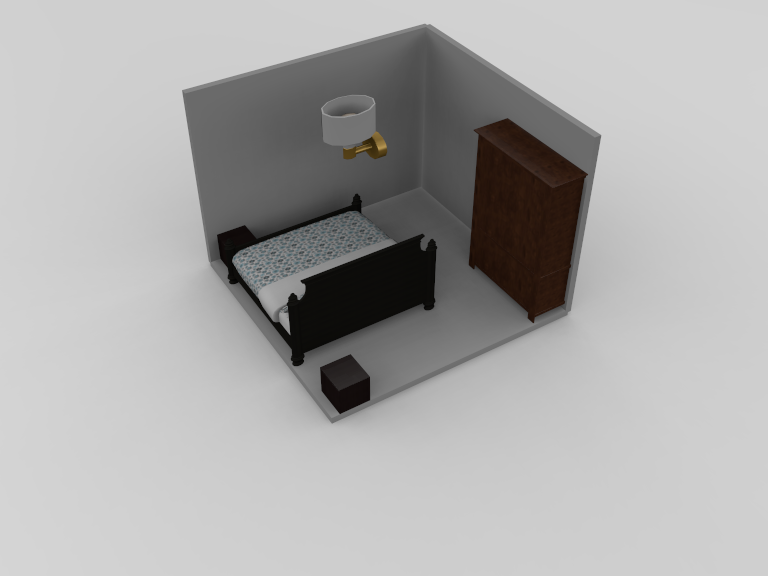}&
        \includegraphics[width=0.15\textwidth, trim=80px 80px 80px 0px, clip]{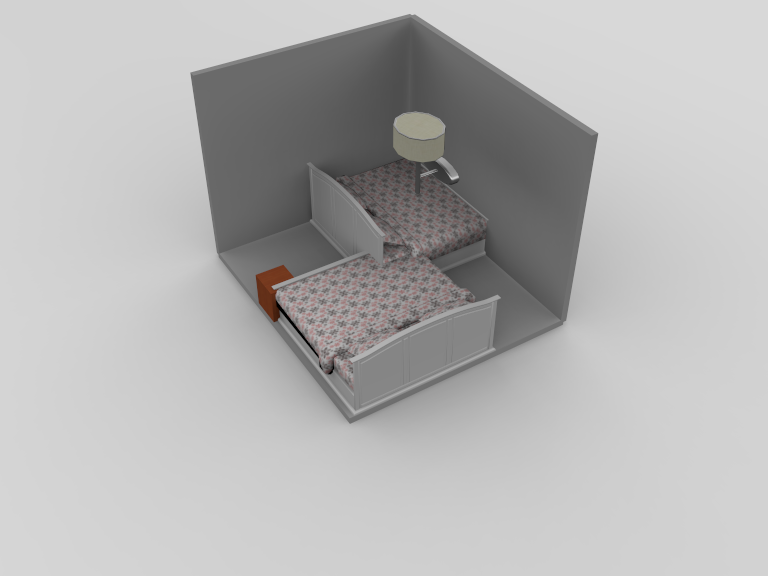} &
        \includegraphics[width=0.15\textwidth, trim=80px 80px 80px 0px, clip]{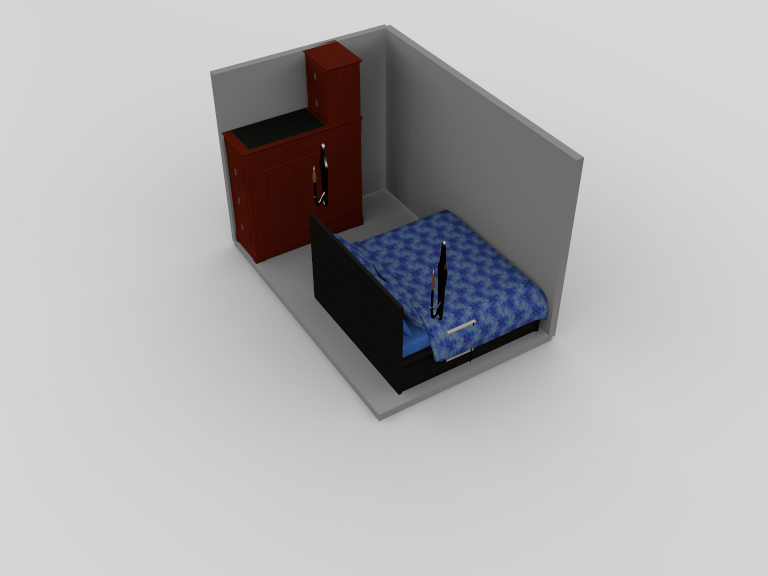}
        \\ \hline
        \multirow{2}{4em}{3D-FRONT Living} &
        \includegraphics[width=0.15\textwidth, trim=80px 80px 80px 0px, clip]{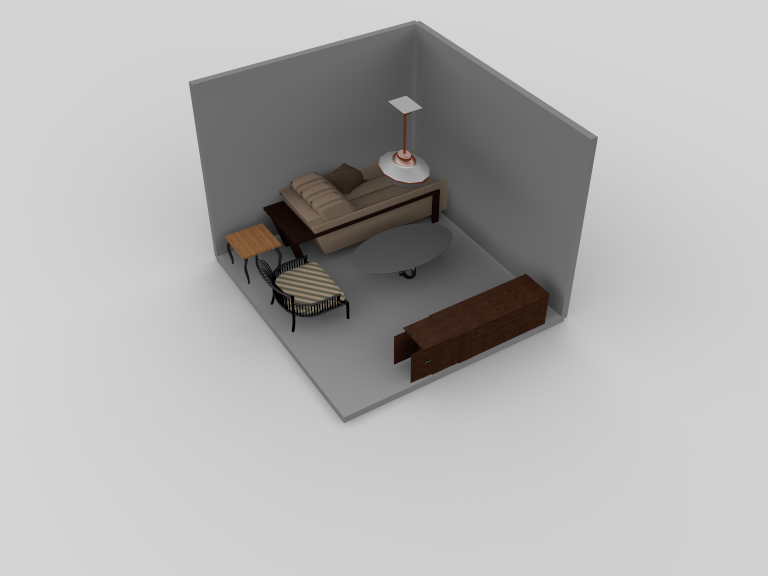} &
        \includegraphics[width=0.15\textwidth, trim=80px 80px 80px 0px, clip]{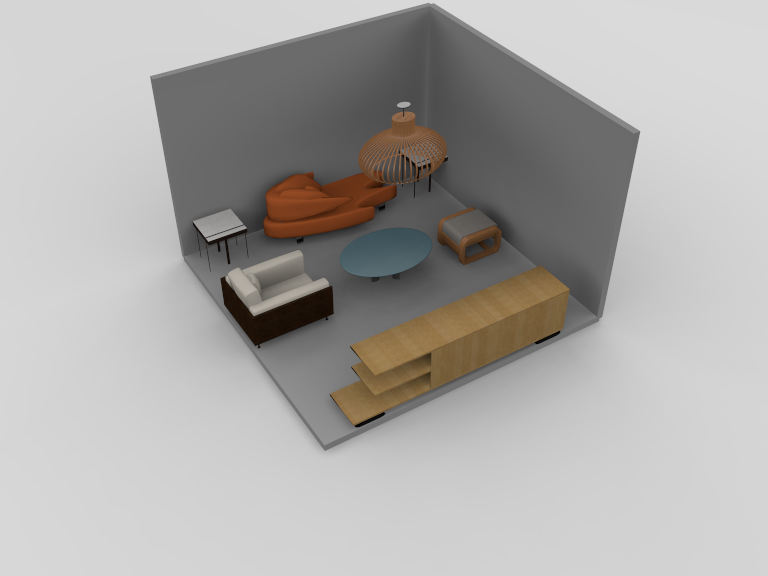} &
        \includegraphics[width=0.15\textwidth, trim=80px 80px 80px 0px, clip]{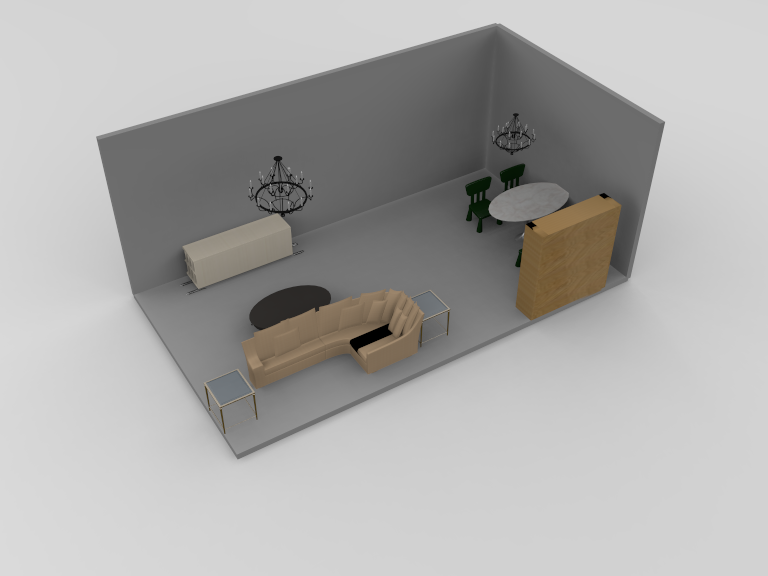} &
        \includegraphics[width=0.15\textwidth, trim=80px 80px 80px 0px, clip]{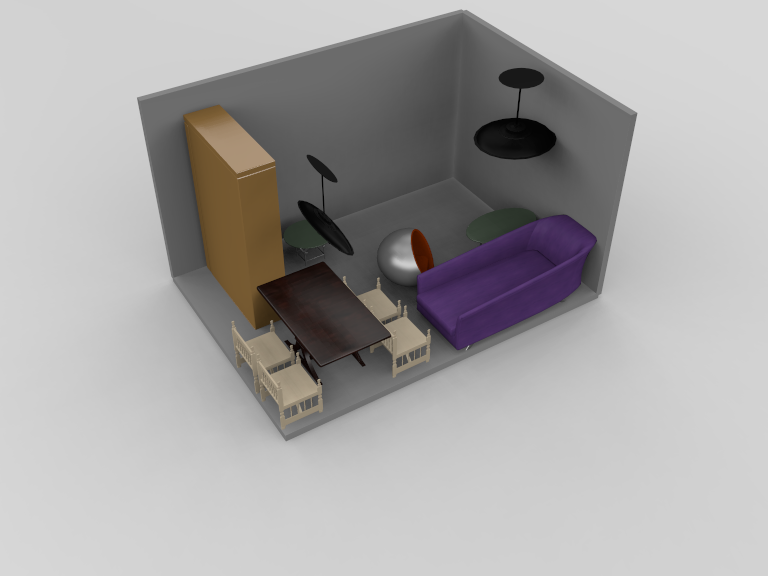} &
        \includegraphics[width=0.15\textwidth, trim=80px 80px 80px 0px, clip]{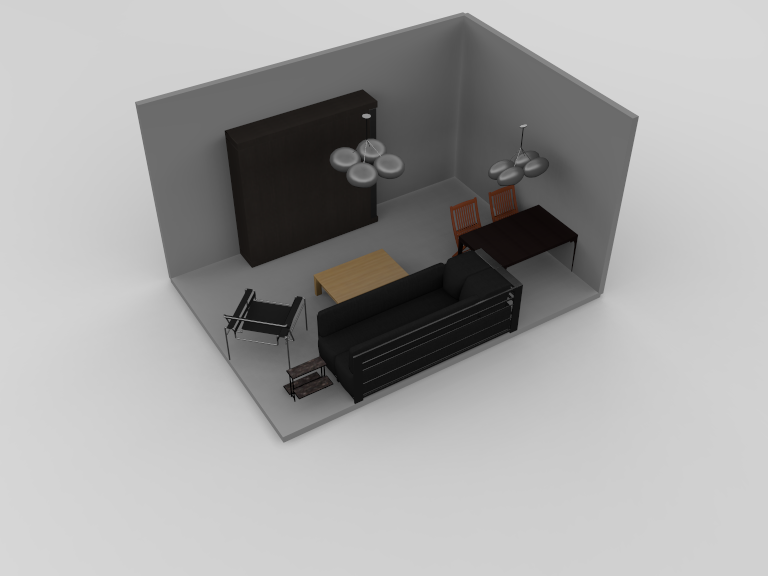}
        \\ 
        & \includegraphics[width=0.15\textwidth, trim=80px 80px 80px 0px, clip]{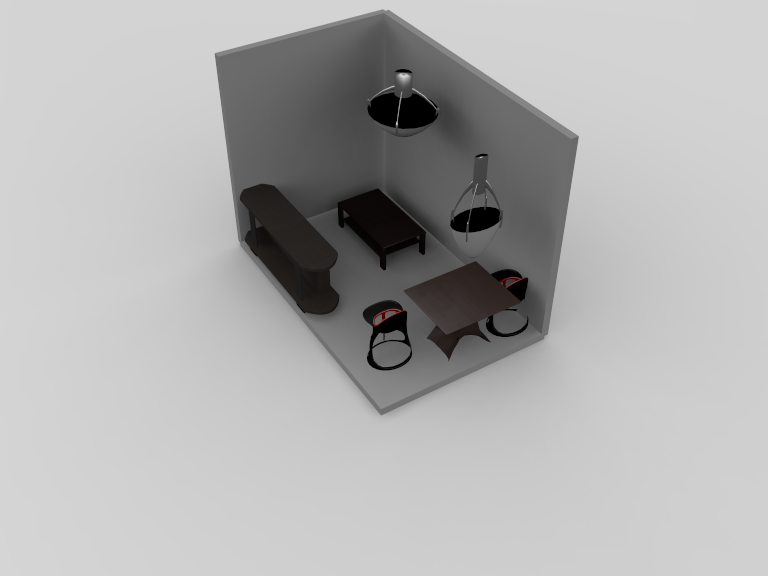} &
        \includegraphics[width=0.15\textwidth, trim=80px 80px 80px 0px, clip]{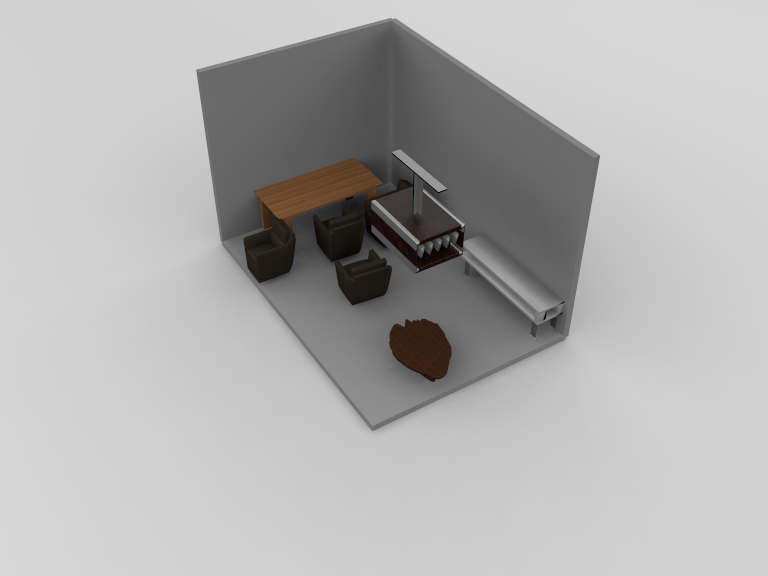} &
        \includegraphics[width=0.15\textwidth, trim=80px 80px 80px 0px, clip]{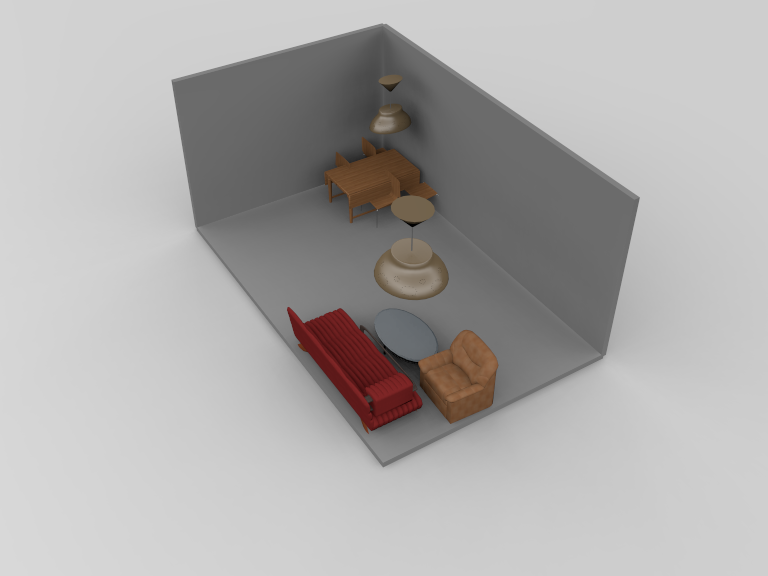} &
        \includegraphics[width=0.15\textwidth, trim=80px 80px 80px 0px, clip]{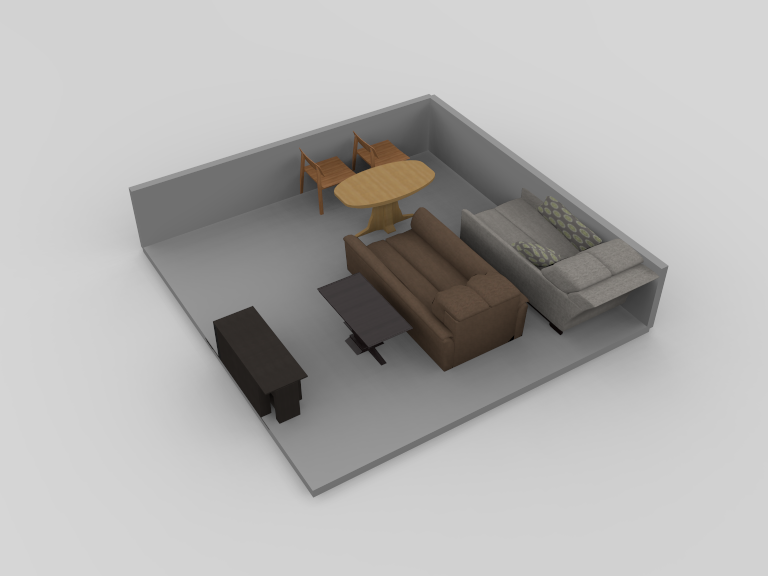} &
        \includegraphics[width=0.15\textwidth, trim=80px 80px 80px 0px, clip]{Figures/front_living/0031__init.png}
        \\\hline 
        & D-Prior & Fast & PlanIT & GRAINS & D-Gen
    \end{tabular}
\caption{Scene synthesis results between ours and baseline methods. For each dataset, the first row: our results, the second row: baseline methods. For baseline methods, from left to right: D-Prior, Fast, PlanIT, GRAINS, D-Gen.}
\label{Figure:Supp:Results_more}
\end{figure*}

Figure~\ref{Figure:Supp:Results} and Figure~\ref{Figure:Supp:Results_more} show more visual comparisons between our approach and baseline approaches. We can see that our approach generates more reasonable scenes than the baselines.

\subsection{More Analysis on Distribution of Relative Attributes}

Figure~\ref{Figure:Relative:Translation_3dfront} shows the distributions of relative translation in the 3D-FRONT dataset. Again we can see the improvements of relative translation, which benefit from incorporating predicted relative attributes and prior modeling.

\subsection{Importance of Utilizing Relative Attributes}
Figure~\ref{Figure:Supp:Relative} shows the comparison between scene optimization with/without utilizing relative attributes. We can see from the results that merely using prior distribution for optimization easily gets stuck in local minimums and can not improve the relative position well.

\begin{figure}
\centering
    \begin{tabular}{ccc}
        \includegraphics[width=0.15\textwidth, trim=80px 80px 80px 0px, clip]{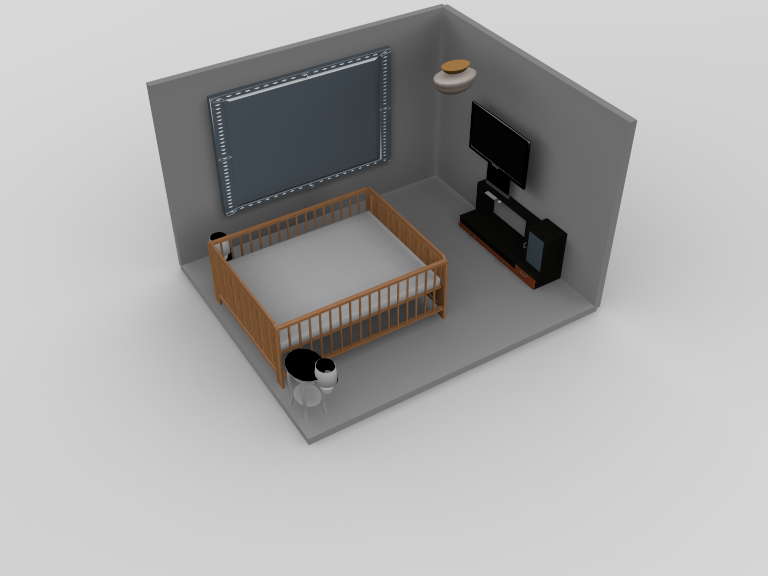} &
        \includegraphics[width=0.15\textwidth, trim=80px 80px 80px 0px, clip]{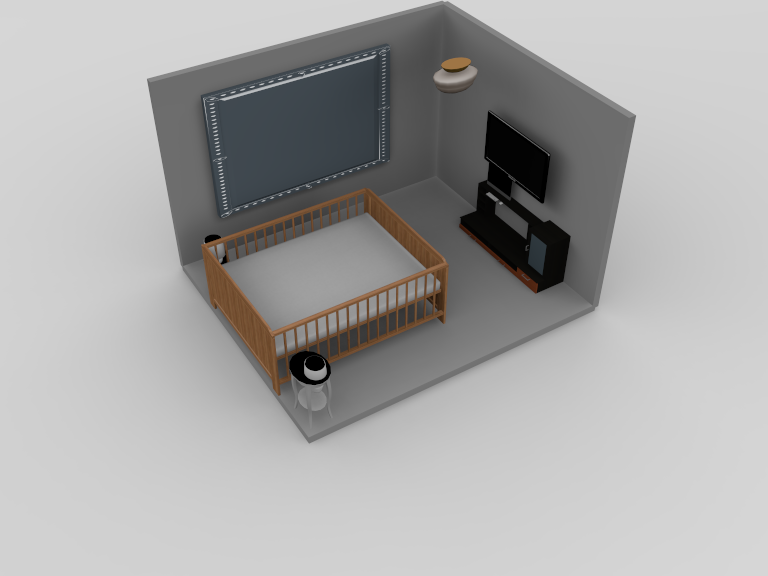} &
        \includegraphics[width=0.15\textwidth, trim=80px 80px 80px 0px, clip]{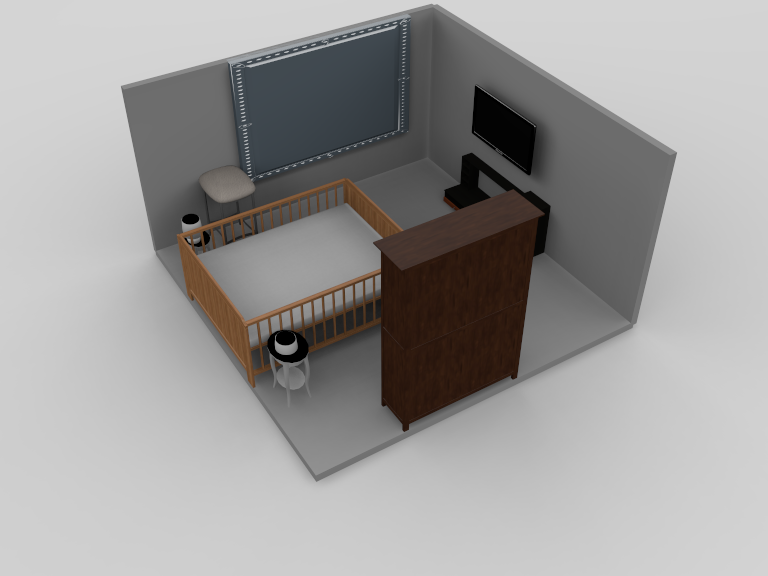}
    \end{tabular}
\caption{Comparison between off-the-shelf scene optimization techniques. Left: output of the prediction module. Middle: scene optimization using off-the-shelf techniques. Right: our method.}
\label{Figure:Supp:Home}
\end{figure}

\subsection{Comparison between Off-the-shelf Scene Optimization Techniques}
Figure~\ref{Figure:Supp:Home} shows the comparison between our approach and off-the-shelf techniques~\cite{Yu:2011:MIH}. Given the same initial scene, both approaches can achieve reasonable arrangements. However, by incorporating the prior distribution and relative attributes into the optimization pipeline, our approach can remove redundant objects (e.g. the laptop on the TV stand) and add diverse objects (e.g. the chair and the cabinet).


\end{document}